\documentclass{article}
\usepackage{microtype}
\usepackage{graphicx}
\usepackage{subcaption}
\usepackage{booktabs}
\usepackage{hyperref}

\newif\ifanon
\anonfalse
\ifanon
  \usepackage{icml2026}
\else
  \usepackage[preprint]{icml2026}
\fi
\usepackage{amsmath,amssymb,mathtools,xcolor}
\usepackage[capitalize,noabbrev]{cleveref}
\newcommand{\oracle}{\textsc{Oracle}}
\newcommand{\valnormal}{\textsc{Val-Normal}}
\newcommand{\last}{\textsc{Last}}
\newcommand{\hpfree}{\textsc{HP-Free}}
\newcommand{\eselect}{\textsc{E-Select}}
\newcommand{\audY}{$\bullet$}
\newcommand{\audN}{$\circ$}
\makeatletter
\providecommand{\appRaggedCells}{\let\app@apbr\@arrayparboxrestore
  \def\@arrayparboxrestore{\app@apbr\raggedright\let\\\tabularnewline}}
\makeatother
\icmltitlerunning{Exposing Evaluation Flaws in Graph-Level Anomaly Detection}
\hypersetup{pdfsubject={},pdftitle={What Remains Without Leaking Test Labels? Exposing Evaluation Flaws in Graph-Level Anomaly Detection}}
\begin{document}
\flushbottom
\twocolumn[
  \icmltitle{What Remains Without Leaking Test Labels? \\
    Exposing Evaluation Flaws in Graph-Level Anomaly Detection}
  \ifanon
\else
    \begin{icmlauthorlist}
      \icmlauthor{Yujie Lu}{iu}
      \icmlauthor{Da Yan}{iu}
    \end{icmlauthorlist}
    \icmlaffiliation{iu}{Luddy School of Informatics, Computing, and Engineering,
      Indiana University Bloomington, Bloomington, Indiana, USA}
    \icmlcorrespondingauthor{Yujie Lu}{yujlu@iu.edu}
    \icmlcorrespondingauthor{Da Yan}{yanda@iu.edu}
  \fi
  \icmlkeywords{graph-level anomaly detection, benchmarking, evaluation protocol, model selection, reproducibility}
  \vskip 0.3in
]
\printAffiliationsAndNotice{}

\begin{abstract}
Graph-level anomaly detection (GLAD) identifies unusual graphs in applications such as molecular screening, often using only normal graphs for training.
Yet the evaluations of some detectors select models with test labels, and high benchmark scores alone do not show whether detectors recognize structural changes.
We audit released code and compare the area under the receiver operating characteristic curve (AUROC) for model states (checkpoints) chosen with and without test labels from the same training run.
We separately compare test-tuned hyperparameters with repository defaults and test sensitivity to structural changes while controlling graph size.
We evaluate 21 detectors on the UB-GOLD benchmark and on synthetic and molecular graphs with controlled structural changes.
About half of the checkpoint-selection gain is linked to choosing a checkpoint and reporting its performance on the same labeled test set.
The mean gap persists across training seeds and when normal and anomalous graphs come from different datasets.
Most detectors miss many tested structural changes that simple counts detect.
With tuned hyperparameters, the two leaders remain the top two by AUROC under every checkpoint rule.
Across nine main datasets, test-label checkpoint selection raises the mean AUROC of UB-GOLD's 12 detectors by 5.12 points relative to the final checkpoint.
\end{abstract}

\section{Introduction}
\label{sec:intro}

Graph-level anomaly detection (GLAD) aims to identify unusual graph-structured objects and has attracted growing interest \citep{qiao2025gadsurvey} for applications such as molecular screening \citep{glocalkd}, financial network monitoring \citep{ocgtl}, and system security \citep{streamspot}.
In this task, each data object is a whole graph, and anomalies may differ in graph structure, node attributes, or both \citep{glocalkd,gladc}.
Because labeled anomalies are scarce and their forms may be unknown in advance, detectors are commonly trained on normal graphs alone \citep{ma2023gadsurvey,ocgtl}.
When labeled anomalies are unavailable, this constraint also applies to selecting hyperparameters and the checkpoint (the model state at a training epoch) used for testing \citep{ma2023uoms}.

Recent methods learn graph representations through one-class objectives, knowledge distillation, and contrastive learning \citep{zhao2023ocgin,ocgtl,glocalkd,gladc,cvtgad}.
Studies typically convert graph classification datasets into anomaly detection tasks and compare detectors by the area under the receiver operating characteristic curve (AUROC) on test graphs \citep{zhao2023ocgin,ubgold}.
UB-GOLD evaluates detectors through shared code and supplies baseline results reused in later method papers \citep{ubgold,lgkde,glass}.
These comparisons guide detector choice and judgments of progress.

However, we observe that some released evaluation pipelines use anomaly labels from the test set to select checkpoints even though the detectors train on normal graphs alone.
UB-GOLD explicitly discloses hyperparameter tuning by test AUROC and describes its results as upper bounds \citep{ubgold}, but its shared training loop also retains the checkpoint with the highest test AUROC, a rule the paper does not describe.
Comparing these scores with results selected without anomaly labels makes it harder to separate detector performance from the benefit of using labels for selection \citep{domainbed,ma2023uoms}.
Selecting among noisy test estimates can raise the reported score even when checkpoints have the same true performance \citep{varma2006bias,cawley2010overfitting}.

A second question is whether high benchmark scores show that detectors recognize anomalous graph structure.
When normal and anomalous graphs differ in simple statistics such as size, high AUROC does not show whether a detector recognizes structural changes that preserve graph size.
We therefore examine both whether checkpoint selection uses test labels unavailable in the target setting and which specific structural changes detectors respond to.

We address these questions through a systematic re-evaluation of UB-GOLD and related GLAD methods, with three contributions.
\begin{enumerate}
  \setlength{\itemsep}{2pt}\setlength{\parskip}{0pt}
  \item \textbf{Audit of released code.}
  We trace how test labels and data handling affect reported scores in released implementations and compare the code's behavior with the procedures described in the papers.
  \item \textbf{Effects of model selection.}
  We measure the AUROC gain from selecting checkpoints using test labels relative to using the final checkpoint or selecting on held-out normal graphs.
  These comparisons keep the training run and hyperparameters fixed, while test-tuned hyperparameters and repository defaults are compared separately.
  To study the sources of these increases, we permute test labels and evaluate another seed's checkpoint at the same epoch chosen in the original seed.
  We also select checkpoints on one labeled subset and report their performance on another, the \emph{disjoint labeled split} (Section~\ref{sec:f2-mech}).

  \item \textbf{Controlled tests of structural sensitivity.}
  We construct synthetic anomalies that change motif composition while preserving the normal graph-size distribution, and molecular anomalies that change ring structure while preserving each molecule's node and edge counts.
  We compare detectors with size and count references that require no gradient training and form benchmark test subsets by pairing anomalous and normal graphs with similar node and edge counts.
  These comparisons help identify which graph features contribute to the measured performance.
\end{enumerate}
We release data splits, evaluation code, synthetic and molecular graph datasets with explicitly specified structural anomalies, and per-run outputs for verification without retraining.

We conduct extensive experiments on UB-GOLD's original folds, covering its 12 detectors and nine recent methods.
The main comparison evaluates all 21 detectors on nine datasets, denoted UB9.
Some detectors are also evaluated on additional datasets (Appendix~\ref{app:extra-perdataset}).
Follow-up analyses of UB-GOLD's detectors cover additional training seeds and Type III, where normal and anomalous graphs come from different datasets.
Synthetic and molecular graph datasets provide known structural changes for testing detector sensitivity.
We could not reproduce or verify the published results for every recent method, so these comparisons remain descriptive (Appendix~\ref{app:ports-fidelity}).

\begin{figure}[t]
  \centering
  \includegraphics[width=\linewidth]{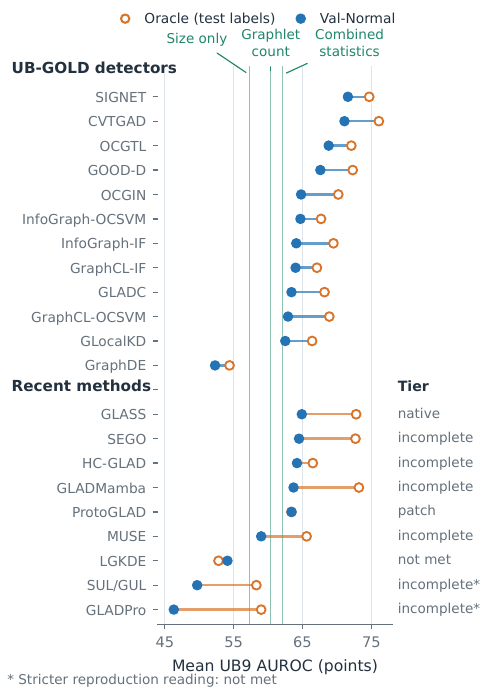}
  \caption{UB9 AUROC with and without test-label checkpoint selection.}
  \label{fig:teaser}
  \vspace{-0.5cm}
\end{figure}

Figure~\ref{fig:teaser} shows mean UB9 AUROC when checkpoints are selected using test labels (\oracle{}) or held-out normal graphs (\valnormal{}), with tuned hyperparameters held fixed.
The upper group contains UB-GOLD detectors, and the lower group gives descriptive recent-method results.
Orange open circles show \oracle{}, blue filled circles show \valnormal{}, and green vertical lines show the fixed references.
ProtoGLAD has no checkpoints, so its two markers coincide.
Recent methods use their own test-label rule where one exists.
Labels on the right summarize implementation changes, missing settings, and reproduction checks.
An asterisk marks checks not met under the stricter interpretation in Appendix~\ref{app:ports-fidelity}.

The audit finds test-label checkpoint or hyperparameter selection in eight of the 14 code bases examined beyond UB-GOLD.
Under \valnormal{}, UB-GOLD's test-tuned hyperparameters add 1.53 AUROC points over repository defaults on UB9 and show no clear average benefit in Type III.
With tuned hyperparameters, SIGNET and CVTGAD retain their AUROC lead under every checkpoint rule, while the ranking below them depends on the rule, hyperparameters, and seed.
In tests that control graph size, most of the 12 UB-GOLD detectors miss many structural changes that simple counts detect.
For UB-GOLD, the mean checkpoint-selection gap persists across additional training seeds and when normal and anomalous graphs come from different datasets.
The \emph{disjoint labeled split} analysis indicates that about half of the gain over the final checkpoint is associated with selecting and reporting on the same graphs.
Most notably, across UB9, test-label checkpoint selection raises mean AUROC over the 12 UB-GOLD detectors by 5.12 points relative to the final checkpoint of the full training budget (\last{}), with a 95\% confidence interval of [3.97, 6.24].

\section{Related Work}
\label{sec:background}

\paragraph{Graph-level detection and benchmarks.}
UB-GOLD provides a shared framework that unifies data loading, model training, and evaluation for graph-level anomaly and out-of-distribution detection \citep{ubgold}.
We study three of its scenarios, in which anomalies are active compounds in toxicity assays (Type I), graphs from a designated class (Type II), or graphs from another dataset (Type III).
Its detectors include end-to-end models \citep{zhao2023ocgin,ocgtl,glocalkd,signet,gladc,cvtgad,goodd,graphde} and two-stage methods that first learn graph embeddings \citep{graphcl,infograph} and then score them using an isolation forest or a one-class SVM \citep{iforest,ocsvm}.
Our evaluation also includes nine recent detectors \citep{muse,lgkde,gladpro,gladmamba,sego,glass,protoglad,hcglad,sulgul} and graph-kernel baselines \citep{wlkernel,pkernel}.
BOND separately reports expected and best performance over a hyperparameter grid \citep{bond}, ADBench evaluates detectors with default hyperparameters \citep{adbench}, and GADBench reports results both with and without hyperparameter tuning \citep{gadbench}.
UB-GOLD's released code uses test AUROC to choose both hyperparameters and the checkpoint used for final evaluation.

\paragraph{Model selection without anomaly labels.}
In an evaluation of label-free model selection, stand-alone internal criteria did not significantly outperform random selection, and consensus strategies did not outperform a default isolation forest \citep{ma2023uoms}.
Other strategies include meta-learning, hyper-ensembles, self-supervised validation losses, and synthetic anomalies \citep{metaod,elect,robod,autouad,dsv,swsa}.
For GLAD, GLAM selects models without labels \citep{glam}, while OCGTL uses validation loss on normal graphs \citep{ocgtl}.
Our \valnormal{} follows the latter idea without claiming optimality.
Comparisons between methods can change when hyperparameters are tuned on test data, self-supervised tasks are chosen using known anomalies, or tuned settings are compared with defaults \citep{robod,yoo2023augment,autogad,soenen2021effect}.

\paragraph{Selection as part of evaluation.}
EfficientAD reports the final model's score rather than the best test score observed during training \citep{efficientad}.
ADer fixes checkpoint rules \citep{ader}, MVTec AD 2 keeps test labels on an evaluation server \citep{mvtecad2}, and OpenOOD v1.5 adds validation data for selection \citep{openood15}.
DomainBed treats selection as part of the method and reports test-domain selection separately as an oracle \citep{domainbed}.
Audits in unsupervised domain adaptation document target-label checkpoint selection and performance losses under realistic validation \citep{musgrave2021uda,ericsson2023better}, with related concerns in semi-supervised and few-shot learning \citep{oliver2018realistic,perez2021true}.
Maximizing noisy test estimates biases reported scores upward even when candidates have equal true performance \citep{varma2006bias,cawley2010overfitting}.
Prior studies find that repeated community use of large public test sets can cause little overfitting \citep{recht2019imagenet,roelofs2019meta}.
We study a different practice, using test labels to choose checkpoints within each training run.

\paragraph{Evaluation audits.}
Graph-classification re-evaluations show that structure-agnostic baselines can outperform graph neural networks on some datasets when selection excludes test data \citep{errica2020}.
Time-series anomaly detection audits likewise identify protocols and datasets that favor simple detectors \citep{kim2022tsad,wu2021tsad}.
In GLAD, reversing normal and anomalous classes can yield below-chance AUROC \citep{zhao2023ocgin}.
MUSE tunes all compared methods on a small labeled validation set kept separate from the test set \citep{muse}.
These validation labels are available for selection, while test labels are reserved for final evaluation.
A concurrent audit studies open-set node-level anomaly detection with a few labeled anomalies \citep{hossain2026selection}.
We audit a one-class GLAD benchmark, compare checkpoint rules on identical training runs, and assess hyperparameter effects separately.
We examine how selecting checkpoints on the reporting graphs raises scores and whether detectors recognize known structural changes.

\section{Task and Evaluation Design}
\label{sec:design}

\paragraph{Problem setup.}
Each example is a whole graph, such as a molecule with atoms as nodes and bonds as edges.
We use UB-GOLD's one-class setting, in which the original classification labels define the normal and anomalous classes \citep{zhao2023ocgin,ubgold}.
A detector trains on normal graphs alone and assigns anomaly scores to test graphs, with higher scores indicating greater abnormality.
AUROC measures the probability that a random anomaly scores above a random normal graph, with ties receiving half credit and 50 points denoting chance.
Model parameters are fitted only on normal training graphs.
The hyperparameters, checkpoints, and score direction may be selected using only normal training data and validation graphs, without access to labeled anomalies or test labels.
Test labels are reserved for final evaluation.

\begin{figure*}[t]
  \centering
  \includegraphics[width=0.95\textwidth]{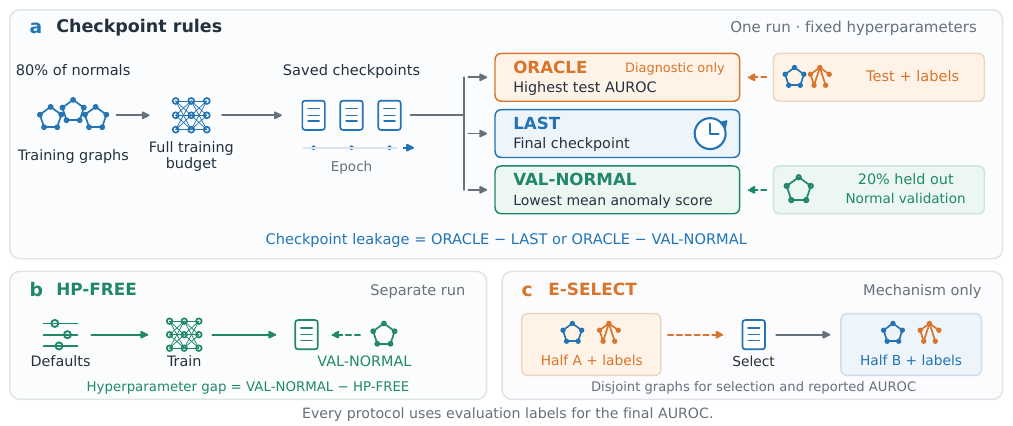}
  \caption{Evaluation protocols for separating checkpoint and hyperparameter effects.}
  \label{fig:overview}
\end{figure*}

\paragraph{Research questions.}
RQ1 examines \emph{implementation validity}, asking whether the released code follows the evaluation procedures described in the papers.
RQ2 examines \emph{selection validity}, asking how using test labels for model selection changes reported performance and what explains these changes.
RQ3 examines \emph{construct validity}, asking whether high benchmark scores show that detectors recognize anomalous graph structure.
We use code audits for RQ1, compare checkpoint rules on the same training run for RQ2, and test anomalies with known structural changes for RQ3.

\paragraph{Model selection protocols.}
Figure~\ref{fig:overview}a compares checkpoint rules on one training run, panel b shows the separate run with default hyperparameters, and panel c separates labeled selection from final evaluation.
Solid branches connect the shared candidate checkpoints, and dashed arrows indicate information used for selection.
Each run in the main comparison holds out 20\% of its normal training graphs for validation, completes the full training budget on the remainder, and saves checkpoints.
\oracle{}, \last{}, and \valnormal{} select checkpoints from the same run with hyperparameters held fixed.
For UB-GOLD detectors, these hyperparameters are the benchmark's per-dataset settings selected by AUROC on test graphs.
For recent methods, we use released configurations.

We use \emph{test-label selection} to mean choosing a checkpoint using AUROC computed from test labels.
\oracle{} applies this selection rule to give a diagnostic upper bound \citep{domainbed}.
UB-GOLD retains the checkpoint with the highest test AUROC.
Among its 12 detectors, the eight end-to-end methods also stop after 30 evaluations without improvement, whereas the four two-stage methods do not stop early.
\oracle{} applies this stopping condition to all 12 detectors.
Training still completes the full budget, and \last{} uses its final checkpoint.
Without this stopping condition for the two-stage detectors, the mean UB9 gaps to \last{} and \valnormal{} each increase by about 0.01 AUROC points (Appendix~\ref{app:ports}).
For checkpoint gaps averaged across datasets and detectors, we apply the same \oracle{} protocol to all methods with checkpoints.
In the main leaderboard, \oracle{} uses each recent method's own test-label selection rule when available.

\last{} uses the final checkpoint without a checkpoint search.
\valnormal{} chooses the checkpoint with the lowest mean anomaly score on held-out normal graphs, following selection on normal data \citep{ocgtl}.
It uses only normal graphs for selection, and we do not claim it is the best such rule.
\hpfree{} applies \valnormal{} to a separate run with repository defaults and makes no selection with test labels in our replay.
Test-label selection with default hyperparameters is used only for robustness analysis.
We use a \emph{disjoint labeled split} only for mechanism analysis.
\eselect{} selects a checkpoint by UB-GOLD's rule using one half of the test graphs and their labels, then reports AUROC on the other half.
The graphs and labels used to select the checkpoint are therefore separate from those used to report AUROC.
All protocols use labels to compute the final metric, but differ in whether and which labels guide model selection.

\paragraph{Performance gaps.}
\emph{Checkpoint leakage} is \oracle{} minus either \last{} or \valnormal{} within the same run, measured in AUROC points.
With the training run and hyperparameters held fixed, this difference measures the increase in reported score from using test labels to select a checkpoint.
The \emph{hyperparameter gap} is \valnormal{} minus \hpfree{}.
This difference combines the effects of tuning and test-label information, so it does not isolate the effect of labels alone.
Both gaps are relative to the specified rules and defaults and do not measure the performance attainable by an optimal label-free procedure.
\hpfree{} may itself be optimistic if the defaults reflect earlier tuning on these datasets.
These runs retain the original test evaluations during training, including random operations such as shuffling test graphs.
These operations share the random-number generator used for training and may affect subsequent parameter updates, an effect examined in Appendix~\ref{app:followup-iso}.

\paragraph{Controlled structural anomalies.}
Benchmark anomalies may differ from normal graphs in several properties at once, so high AUROC alone cannot identify which changes a detector responds to.
We therefore test structural changes while controlling graph size.
Simple statistical controls check whether the changes are detectable and whether size statistics or adjacent node-label pair counts can distinguish normal from anomalous graphs.
Section~\ref{sec:setup} describes the dataset construction and reference methods.

\paragraph{Study stages.}
Our study includes pilot experiments, formal experiments, and post hoc analyses.
Pilot experiments on four UB-GOLD detectors suggested checkpoint leakage and missed structural changes, informing the hypotheses and design of the formal experiments.
The formal structural tests use the remaining eight UB-GOLD detectors, called the formal-test detectors.
Their protocols and criteria were set before these detectors produced structural results.
Checkpoint-gap comparisons include all 12 UB-GOLD detectors.
Post hoc analyses follow up on the observed results to investigate possible explanations and assess robustness.
Appendix~\ref{app:prereg} provides the detailed criteria and study record.

\section{Experimental Setup}
\label{sec:setup}

\paragraph{Data and detectors.}
The main comparison evaluates 21 detectors, including the 12 UB-GOLD detectors and nine recent methods, on nine datasets.
The code audit covers UB-GOLD and 14 other code bases.
We evaluate nine of these additional methods in our experiments and inspect the other five without rerunning them.
These datasets, denoted UB9, are AIDS, BZR, COX2, DHFR, ENZYMES, NCI1, PROTEINS\_full, IMDB-BINARY, and REDDIT-BINARY.
We use the five folds from UB-GOLD commit c940457 and retain its anomaly classes and node features.
In Tox21 HSE and PPAR-gamma, active molecules are the anomaly class at test time.
However, in each training file, UB-GOLD maps the first label it encounters to 0 and the other label to 1, then trains on graphs mapped to 1.
Because these files start with an inactive molecule, the loader incorrectly selects active molecules as normal training graphs.
We exclude both tasks because training and testing use inconsistent definitions of normality.
A separate Type III experiment evaluates the 12 UB-GOLD detectors with normal graphs from one dataset and anomalies from another (Appendix~\ref{app:t3}).

\paragraph{Implementation and selection.}
We retain the released training loops, complete the full training budget, and record evaluation results at scheduled evaluation points.
We then apply the selection rules in Section~\ref{sec:design} to these records.
We take per-dataset settings and epoch counts from the released scripts and default hyperparameters from the repositories.
We perform no additional hyperparameter search.
We negate GraphDE's normal-probability score according to its paper and correct score--label pairing for the two-stage detectors.
Experiments run on a shared cluster with AMD EPYC CPUs and NVIDIA A100 GPUs with 40\,GB of memory per GPU.
Appendices~\ref{app:ports} and~\ref{app:compute} document per-method configurations, reproduction checks, and resource use.

\paragraph{Controlled structural anomalies.}
We test sensitivity to structural changes using the synthetic generator GX and the molecular generator MS-v2.
GX combines tree backbones with small motifs and creates anomalies by introducing unseen motifs or changing their proportions while preserving the graph-size distribution.
MS-v2 edits normal molecules from NCI1 by replacing six-membered rings with smaller rings and short chains or by relocating rings.
The main variants preserve each graph's node count, edge count, and atom labels.
Size-changing molecular variants test whether the same edits are easier to detect when graph size also changes.
These experiments use five seeds, fixed per-detector configurations, and the \valnormal{} checkpoint rule.

\paragraph{Simple statistical references.}
We report three fixed reference methods in the main comparison.
\emph{Size only} uses log-transformed node and edge counts together with graph density.
\emph{Graphlet count} uses log-transformed counts of small connected induced subgraphs, called graphlets, together with log-transformed node and edge counts.
Both references use Mahalanobis distance with Ledoit--Wolf covariance estimation \citep{ledoitwolf}.
\emph{Combined statistics} combines calibrated scores for graph size, size-adjusted graphlet counts, graphlet patterns around individual nodes, and node-label statistics when available.
All three are fitted only on normal training graphs without per-dataset hyperparameter tuning.
In the structural tests, graphlet counts check whether the structural changes are detectable.
Size statistics and adjacent node-label pair counts check whether graph size or label pairs on adjacent nodes alone identify anomalies.
Degree statistics and ring counts check which edits can be detected from node degrees or ring structure.
Appendices~\ref{app:typed} and~\ref{app:kernels} give full definitions of the generators, references, and graph-kernel baselines.

\paragraph{Metrics and uncertainty.}
We report AUROC on a 0--100 scale, with anomalies as the positive class.
For UB9, we average the five folds per dataset, weight datasets equally, and average over detectors for overall gaps.
Checkpoint-gap comparisons exclude methods without checkpoints, while hyperparameter-gap comparisons exclude GLASS and SUL/GUL, which release only one configuration.
For intervals, we resample the nine UB9 datasets with replacement 4000 times, using the same draw across compared protocols and detectors, and report the 2.5th and 97.5th percentiles in brackets.
For Type III, we first average the five trials per normal/anomaly dataset pair, then similarly resample the five pairs.
These intervals capture dataset-resampling uncertainty with the detector set fixed, while seed variation is analyzed separately.
Secondary metrics are area under the precision--recall curve (AUPRC) and the false positive rate at 95\% anomaly recall (FPR95).
Appendices~\ref{app:extra} and~\ref{app:followup} report these metrics and additional checks with more seeds, alternative checkpoint selectors, and the \emph{disjoint labeled split}.

\section{Findings}
\label{sec:results}
Unless stated otherwise, scores are mean UB9 AUROC, and brackets give the 95\% intervals defined in Section~\ref{sec:setup}.

\subsection{Released code uses test labels for selection}
\label{sec:f1}
\label{sec:defects}
For RQ1, the audit finds test-AUROC checkpoint selection in all 12 UB-GOLD detectors.
The eight end-to-end detectors also stop early using the same metric.
Although training uses only normal graphs, choosing the final model still uses anomaly labels.
UB-GOLD's paper discloses test-set hyperparameter search and its upper-bound status but does not describe checkpoint selection.
Of the 14 other audited code bases, eight use test labels to select checkpoints or hyperparameters and five show no such use.
For the remaining code base, we could not determine which results the paper reports.
These findings concern the audited versions.
Data processing can also change which class counts as anomalous.
UB-GOLD treats the majority class as anomalous on five datasets, while later studies reusing its baseline results may instead treat the minority class as anomalous.
Appendix~\ref{app:defects} gives the code evidence, and Appendix~\ref{app:additional-audit} compares released code with paper descriptions and explains the audit-table symbols.
The audit makes no claim about author intent or whether published numbers came from these code paths.

\begin{table*}[t]
  \centering\small
  \caption{Mean AUROC gains from test-label checkpoint selection.}
  \label{tab:main-selection}
  \setlength{\tabcolsep}{7pt}
  \begin{tabular}{@{}llcc@{}}
\toprule
Data & Detectors & \oracle{} $-$ \last{} & \oracle{} $-$ \valnormal{} \\
\midrule
UB9 & UB-GOLD (12) & 5.12 [3.97, 6.24] & 4.13 [3.32, 4.79] \\
UB9 & Recent, with checkpoints (8) & 6.69 [5.75, 7.61] & 7.47 [6.15, 8.83] \\
Type III & UB-GOLD (12) & 9.84 [6.73, 13.07] & 8.11 [5.36, 11.00] \\
\bottomrule
\end{tabular}

\end{table*}

\subsection{Checkpoint selection increases reported scores}
\label{sec:f2}
The main comparison for RQ2 fixes the training run and hyperparameters and changes only checkpoint selection.
Table~\ref{tab:main-selection} summarizes the mean gains of \oracle{} over \last{} and \valnormal{}, with detector-level results in Appendix~\ref{app:extra-inflation}.
All rows, including recent methods, use the common \oracle{} protocol defined in Section~\ref{sec:design}.
Intervals resample datasets for UB9 and normal/anomaly dataset pairs for Type III.
Across the 12 UB-GOLD detectors, \oracle{} exceeds \last{} by 5.12 points [3.97, 6.24] and \valnormal{} by 4.13 points [3.32, 4.79].
For every UB-GOLD detector, the mean \oracle{}--\last{} and \oracle{}--\valnormal{} gaps are each at least 2 points.
These are dataset averages and do not imply gains in every run.
SIGNET's interval for the gap between \oracle{} and \valnormal{} includes zero.
For the eight recent methods with checkpoints, \oracle{} also exceeds \last{} and \valnormal{} on average under UB-GOLD's selection rule.

\paragraph{Hyperparameter gains are smaller and depend on the setting.}
In a comparison using separate training runs, test-tuned settings under \valnormal{} improve UB-GOLD by 1.53 points [0.64, 2.42] over defaults.
This comparison combines the effects of tuning and test-label information and does not isolate the effect of labels alone.
For UB-GOLD, \oracle{} exceeds \hpfree{} by 5.66 points [4.27, 6.86] on average, with method-level differences in Appendix~\ref{app:extra-hpleak}.

\begin{table}[t]
  \centering\footnotesize
  \caption{Leading detectors and simple references on UB9.}
  \label{tab:main-leaders}
  \setlength{\tabcolsep}{2pt}
  \begin{tabular}{@{}lcccc@{}}
\toprule
Detector & \oracle{} & \last{} & \valnormal{} & \hpfree{} \\
& {\scriptsize test-selected} & & & \\
\midrule
SIGNET & 74.7 & 70.7 & 71.6 & 69.3 \\
CVTGAD & 76.1 & 71.0 & 71.1 & 66.5 \\
OCGTL & 72.1 & 68.7 & 68.8 & 60.1 \\
GOOD-D & 72.3 & 67.7 & 67.6 & 66.8 \\
\midrule
\textit{Combined statistics} & 62.1 & 62.1 & 62.1 & 62.1 \\
\textit{Graphlet count} & 60.4 & 60.4 & 60.4 & 60.4 \\
\textit{Size only} & 57.3 & 57.3 & 57.3 & 57.3 \\
\bottomrule
\end{tabular}

\end{table}

\paragraph{The leaders remain ahead, while lower ranks are less stable.}
\label{sec:c2}
With tuned hyperparameters, SIGNET and CVTGAD remain the top two by AUROC under every checkpoint rule.
Table~\ref{tab:main-leaders} shows the four detectors with the highest mean \valnormal{} AUROC alongside the three fixed references defined in Section~\ref{sec:setup}.
Each cell reports mean UB9 AUROC under the corresponding protocol.
The references make no selection changes across columns and therefore retain the same scores.
The complete 21-detector comparison is in Appendix~\ref{app:additional-leaderboard}.
For the 12 UB-GOLD detectors, rankings under \oracle{} and \valnormal{} have Kendall's $\tau_b=0.76$ [0.36, 0.85].
The two rules give different top-three sets in the full UB9 comparison, but the sets differ in only 45\% of dataset-bootstrap samples.
Hyperparameters and training seeds also change ranks (Appendices~\ref{app:extra-kendall} and~\ref{app:followup-seeds}).
Under \last{} and \valnormal{}, OCGTL has the highest AUPRC and lowest FPR95 (Appendix~\ref{app:extra-secondary}).

\subsection{About half the gain is tied to selection on the reported data}
\label{sec:f2-mech}
\oracle{} may score higher by finding better checkpoints or by selecting unusually high test estimates.
We examine these sources for the 12 UB-GOLD detectors on eight UB9 datasets.
We exclude REDDIT-BINARY from these checks because of its computational and memory costs.
The \emph{disjoint labeled split} selects a checkpoint on one half of the test graphs and reports its performance as \eselect{} on the other.
The label-permutation check measures how much choosing the highest-scoring checkpoint can gain when labels are unrelated to the graphs.
The seed-transfer check selects an epoch in one seed and evaluates the checkpoint at the same epoch in another.

\begin{table}[t]
  \centering\footnotesize
  \caption{Sources of the checkpoint-selection gain.}
  \label{tab:mechanism}
  \setlength{\tabcolsep}{3pt}
  \begin{tabular}{@{}lrr@{}}
\toprule
Comparison & Mean & 95\% CI \\
\midrule
\multicolumn{3}{@{}l}{\emph{Full test sets, mean of seeds 0--2}} \\
\oracle{} $-$ \last{} & 5.32 & [4.06, 6.49] \\
Label-permutation best $-$ \last{} & 3.80 & [2.62, 4.98] \\
Seed-transfer $-$ \last{} & 2.55 & [1.16, 4.02] \\
\oracle{} $-$ seed-transfer & 2.77 & [1.97, 3.57] \\
\midrule
\multicolumn{3}{@{}l}{\emph{Disjoint labeled split, seed 0}} \\
\oracle{} $-$ \last{} & 6.50 & [5.11, 7.65] \\
\eselect{} $-$ \last{} & 2.86 & [0.88, 4.94] \\
\oracle{} $-$ \eselect{} & 3.64 & [1.70, 5.64] \\
\bottomrule
\end{tabular}

\end{table}

Table~\ref{tab:mechanism} reports mean AUROC differences and 95\% dataset-bootstrap intervals for the three checks.
For seed-transfer, the first difference is the AUROC at the epoch chosen in another seed minus the AUROC of \last{}.
The second is \oracle{} AUROC minus the AUROC at that transferred epoch.
Across 200 stratified half-splits, \eselect{} exceeds \last{} on the evaluation half by 2.86 points [0.88, 4.94].
\oracle{} selected on that evaluation half scores another 3.64 points [1.70, 5.64] above \eselect{}.
This additional gap estimates how much selecting on the reporting graphs raises AUROC and accounts for about half the total gain in this analysis.

Label-permutation yields a mean gain of 3.80 points [2.62, 4.98] over \last{}, about 71\% of the corresponding 5.32-point \oracle{}--\last{} gap (Table~\ref{tab:mechanism}).
Checkpoint selection can therefore raise reported AUROC even when test labels are unrelated to the graphs.
The label-permutation check takes the highest AUROC across all recorded evaluation points in the full run, without applying the original early-stopping rule.
This gain is an upper bound on label-permutation gains under the original early-stopping rule and it does not estimate how much of the observed gap comes from noise.

The gains retained in seed-transfer and the \emph{disjoint labeled split} suggest that better label-free checkpoint selection may be possible.
But both analyses still use anomaly labels to choose checkpoints.
The extra score from choosing checkpoints on the reporting graphs should not be interpreted as better performance on new test graphs.
Appendix~\ref{app:followup} gives the full diagnostics and method-specific exceptions.

In Type III, the mean \oracle{}--\last{} gap is 9.84 points [6.73, 13.07] (Table~\ref{tab:main-selection}).
The Type III hyperparameter gap is $-0.47$ points [$-1.67$, 0.83], without a clear average benefit (Appendix~\ref{app:t3}).
In a separate check, we include the validation graphs in training and compare \oracle{} with \last{}.
We also test repository-default hyperparameters and the setting in which the minority class is anomalous, with detailed results in Appendix~\ref{app:followup-robust}.
The checks cover different detector and dataset subsets, and the mean checkpoint gap remains positive in each subset.
Some GPU reruns produce different scores even at the same seed, so differences between runs also include device nondeterminism (Appendix~\ref{app:ports-devices}).

\begin{figure}[t]
  \centering
  \includegraphics[width=\linewidth]{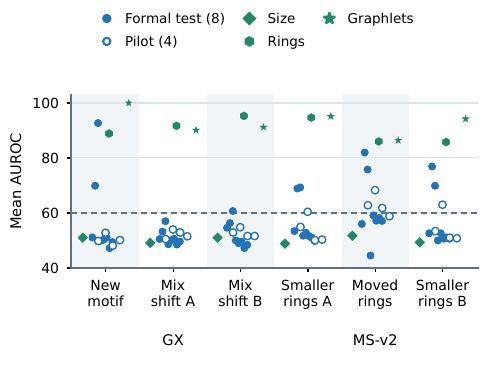}
  \caption{Responses to size-preserving structural changes.}
  \label{fig:typed-main}
  \vspace{-0.3cm}
\end{figure}

\subsection{High benchmark scores do not establish structural sensitivity}
\label{sec:f3}
For RQ3, we compare detectors on six size-preserving structural changes that pass the validity controls.
Figure~\ref{fig:typed-main} shows five-seed mean \valnormal{} AUROC for the 12 UB-GOLD detectors and simple references on each type.
Filled blue circles denote the eight formal-test detectors, open blue circles the four pilot detectors, and green symbols the references.
We define a detector as \emph{blind} to a type when its five-seed mean \valnormal{} AUROC is at most 60, marked by the dashed line.
Graphlet counts reach 86.4 to 100 and ring counts 85.8 to 95.3, while the size reference scores only 48.8 to 51.7.
Across the eight formal-test detectors and six anomaly types, 81\% of the 48 combinations have mean AUROC at most 60.
Most tested methods therefore miss changes that simple structural counts detect.
CVTGAD and GOOD-D are important exceptions, detecting all molecular edit types while responding weakly to GX motif-proportion changes.

For the 24 combinations of eight formal-test detectors and three molecular edit types, the share with mean AUROC at most 60 falls from 75\% to 12.5\% when the edits also increase graph size.
The blind-spot claim therefore covers only the tested size-preserving changes.
We also test structural encodings, other checkpoint rules, and different AUROC thresholds for classifying a detector as blind.
The structural limitation does not apply to every detector.
For example, the recent methods SEGO and GLASS detect all six types, with their results and other recent-method exceptions in Appendix~\ref{app:typed}.

\paragraph{Simple statistics also predict benchmark labels.}
\label{sec:c5}
The size reference reaches 98.7 AUROC on AIDS, and graphlet counts average 60.4 on UB9, while SIGNET and CVTGAD outperform every simple reference.
In a post hoc analysis, we match normal and anomalous test graphs by node and edge counts and recompute AUROC without retraining detectors or reselecting checkpoints.
Across the six datasets meeting the matching requirements, mean \valnormal{} AUROC for the eight formal-test UB-GOLD detectors falls from 61.1 on the original test sets to 53.8 on the matched subsets.
Because matching requires similar rather than identical node and edge counts, size differences can remain between paired graphs.
The subsets are selected by size rather than randomly sampled, so the AUROC decrease is descriptive.
The drop after size matching supports a contribution from graph size to benchmark performance, but does not show that detectors use only graph size or estimate how often they miss real anomalies.
Appendix~\ref{app:typed} gives the full structural results, degree and ring-count controls, and size-matching details, while Appendix~\ref{app:additional-structural} gives the extended figure.

\section{Recommendations and Limitations}
\label{sec:recommendations}
\paragraph{Recommendations.}
GLAD evaluations should report which data are used to choose hyperparameters, checkpoints, and score direction \citep{domainbed}.
Results selected using test labels should be clearly marked as diagnostic upper bounds.
Reports should state whether labeled anomalies are available for selection.
When these labels are used, final evaluation should use separate graphs, as in MUSE \citep{muse}.
Results from all training seeds should be reported, either individually or as a mean.
Reused baseline results should be accompanied by their source, code version, anomaly-class definition, and selection rule.
Benchmark comparisons should include controlled structural tests and simple size and count references.
Controlled tests check whether detectors identify specified structural changes, while simple references test whether graph size and small substructure counts predict benchmark labels.
Releasing checkpoint scores, per-graph scores, and configurations allows other researchers to compare checkpoint rules and run further analyses without retraining.
Appendix~\ref{app:additional-scope} lists the reporting items.

\vspace{-0.2cm}
\paragraph{Limitations.}
Our experiments cover a limited set of datasets and detectors and may not represent other GLAD settings.
The normal-only selection rules and repository defaults we test may be suboptimal without anomaly labels, and better label-free checkpoint rules could narrow the gap to test-label selection.
Comparisons of recent methods remain descriptive because we could not verify reproduction of every method's published results.
The blind rate is measured on these constructed anomaly types and does not estimate how often detectors miss real anomalies.
Our molecular tests cover only ring edits, while changes in motif proportions are tested only on synthetic graphs.

\section{Conclusion}
\label{sec:conclusion}
On UB9, test-label checkpoint selection raises every UB-GOLD detector's mean AUROC relative to \last{} and \valnormal{}.
For the eight recent methods with checkpoints, \oracle{} also outperforms both rules on average under UB-GOLD's selection rule.
For UB-GOLD, the mean gap persists across additional training seeds and when normal and anomalous graphs come from different datasets.
In the \emph{disjoint labeled split}, about half of \oracle{}'s gain over \last{} is associated with selecting and reporting on the same graphs.
Most UB-GOLD detectors miss many tested size-preserving structural changes detected by simple counts.
Under fixed configurations, SEGO and GLASS detect all six tested size-preserving anomaly types, while five other recent methods miss all six and four also miss the size-changing molecular variants.
A reliable assessment of progress in GLAD requires clear model-selection rules and conclusions about detection ability supported by experimental evidence.

\section*{Impact Statement}
This work aims to make GLAD evaluations reflect the performance available to users without anomaly labels.
Overstated benchmark results can lead users to choose weaker detectors in applications such as molecular screening.
The audit describes code at specified versions without judging author intent or establishing that it produced the published values.
Some papers disclose their use of test labels.
Appendix~\ref{app:defects-credit} also lists practices that help readers assess the reported results.
We welcome corrections to the audit.
Controlled structural tests do not establish performance on real anomalies.
All datasets are public, the social-network data contain only graph structure without node attributes, and we collected no data about people.

\setlength{\bibsep}{3pt plus 1pt minus 0.5pt}
\bibliography{refs}
\bibliographystyle{icml2026}
\newpage
\appendix
\raggedbottom
\setlength{\textfloatsep}{12pt plus 2pt minus 2pt}
\setlength{\floatsep}{10pt plus 2pt minus 2pt}
\setlength{\intextsep}{10pt plus 2pt minus 2pt}
\captionsetup{skip=6pt}
\onecolumn
\section*{Guide to the appendix}

The appendices provide complete results, experimental details, code checks and resource information that supplement the main text.

Table~\ref{tab:app-overview} lists the contents of each appendix and the corresponding main-text sections.
The full code locations, the summary of every amendment and the per-cell tables are in the supplementary study record released with the code \ifanon\else(\url{https://github.com/Lu-Yujie/Label-Free-GLAD})\fi{}.
We abbreviate IMDB-BINARY and REDDIT-BINARY as \mbox{IMDB-B} and \mbox{REDDIT-B}, except in Type III, which keeps UB-GOLD's dataset names.
A0 to A49 denote dated amendments to the study plan (Appendix~\ref{app:prereg-amend}), U and T with a number denote audit entries (Appendix~\ref{app:defects}), and a letter after the word part, as in part E, names a follow-up study in the amendment record rather than an appendix.
Appendix~\ref{app:compute-datasheet} lists the released resources.

\begin{table}[!htb]
  \centering\footnotesize\setlength{\tabcolsep}{4pt}
  \caption{Questions supported by each appendix.}
  \label{tab:app-overview}
  \begin{tabular}{@{}p{0.6cm}p{5.2cm}p{8.0cm}p{2.0cm}@{}}
    \toprule
    & Part & Question it answers & Main text \\
    \midrule
    \ref{app:prereg} & Study plans and amendments & Which plans were written before the results, and how are their dates recorded? & \S\ref{sec:design} \\
    \ref{app:ports} & Implementations and reproduction checks & Which code was used, and did it reproduce the released behavior? & \S\ref{sec:setup} \\
    \ref{app:defects} & Code and data audit & How do test-label selection, data loading and scoring affect results? & \S\ref{sec:f1} \\
    \ref{app:extra} & Complete UB9 results & What are the per-detector scores, checkpoint gaps and rankings? & \S\ref{sec:f2} \\
    \ref{app:followup} & Checkpoint selection and additional checks & How much gain remains on separate graphs or seeds, and do gaps persist under other settings? & \S\ref{sec:f2-mech} \\
    \ref{app:t3} & Anomalies from another dataset (Type III) & Do checkpoint gaps persist when normal and anomalous graphs come from different datasets? & \S\ref{sec:f2-mech} \\
    \ref{app:typed} & Controlled structural anomalies & How are the anomalies constructed and checked, and which changes do detectors detect? & \S\ref{sec:f3} \\
    \ref{app:kernels} & Statistical references and kernel baselines & How are references fitted, and what changes after scores are paired with the correct labels? & \S\ref{sec:setup}, \S\ref{sec:f3} \\
    \ref{app:compute} & Compute and released resources & What hardware, computation and files are needed to repeat the experiments? & \S\ref{sec:setup} \\
    \ref{app:additional} & Additional results and reporting guidance & Which full tables and figures supplement the main results, and what should future studies report? & \S\ref{sec:results}, \S\ref{sec:recommendations} \\
    \bottomrule
  \end{tabular}
\end{table}

\makeatletter
\providecommand{\appRaggedCells}{\let\app@apbr\@arrayparboxrestore
  \def\@arrayparboxrestore{\app@apbr\raggedright\let\\\tabularnewline}}
\makeatother
\section{Study plan and amendment records}
\label{app:prereg}

This appendix explains when each analysis was specified and what evidence supports that timing.
Pilot experiments developed the hypotheses, formal experiments tested specified questions, and post hoc analyses investigated the observed results.
An earlier version of this paper, called the first version below, reported the original detector and dataset comparisons and the controlled structural tests.
This paper is the second version and adds the follow-up studies described below.
In the terms of Section~\ref{sec:design}, the formal structural test, criterion G2(c) below, uses the eight detectors added after the pilot, the formal-test detectors, while the checkpoint and hyperparameter criteria G1, G2(a) and G2(b) cover all 12 UB-GOLD detectors.
The follow-up studies were designed after the formal results and are post hoc in that sense, although each was specified before its own results and the Type III replication has its own decision criterion, R1.
Earlier records call the formal tests and R1 confirmatory.

\subsection{From pilot experiments to formal tests}
\label{app:prereg-plan}

The original study plan was written on 2026-10-12 and has not been edited since.
Its header states 21:20 EDT and its file timestamp 20:51 EDT, a discrepancy recorded without changing the content in amendment A0.
By then, pilot experiments on SIGNET, OCGTL, OCGIN and GLocalKD had produced results using UB-GOLD hyperparameters.
The remaining eight detectors, GLADC, CVTGAD, GOOD-D, GraphDE and the four two-stage detectors, had produced no test metrics.

The plan asked how much performance on UB-GOLD folds depends on using test labels to select checkpoints and hyperparameters, and whether rankings change when those labels are unavailable.
It specified the 12 UB-GOLD detectors, kernel baselines and training-free references, taking GraphDE score direction from its paper and code.
UB9 formed the main dataset group, and DD, COLLAB, Tox21 p53 and Tox21 MMP formed a secondary group, the additional datasets mentioned in Section~\ref{sec:intro}.
A10 added MUTAG to the secondary group, and all experiments used UB-GOLD five-fold splits.

The plan specified \oracle{}, \last{}, \valnormal{} and \hpfree{}, with one run per detector, dataset, fold and setting.
Each run held out 20\% of the normal training graphs using a fixed seed.
AUROC was the primary metric, with AUPRC and FPR95 as secondary metrics.
Summaries first average the five folds and then weight datasets equally.
The plan also specified paired fold-bootstrap and dataset-bootstrap 95\% intervals, Kendall correlation between rankings, and Friedman tests with Holm correction.
Missing runs must be reported, and only detectors with complete UB9 results enter the decision.

The original decision required both G1 and G2.
G1 required at least eight deep detectors with complete UB9 runs, a mean \oracle{} minus \valnormal{} of at least 3.0 AUROC points, and a gap of at least 2.0 points for at least 75\% of those detectors.
G2 required at least one of the following three outcomes.
G2(a) required Kendall $\tau\leq0.8$ between the two leaderboards or a change in the top three.
G2(b) required \hpfree{} to score at least 2.0 points below \valnormal{} on average.
G2(c) required at least 75\% of the deep detectors to be blind to three typed anomalies detectable by a graphlet-count control.

These rules were not all specified before the pilot results.
The original plan labels G2(c) post hoc because its thresholds were set after the four pilot detectors had produced typed-anomaly results.
A9 changed G2(c) to a test on the other eight detectors and a new molecular generator.
This amendment preceded every typed-anomaly result from those detectors and the implementation of the generator.
After the pilot aggregates but before any G1 aggregate over all 12 detectors, A14 additionally required the \oracle{} comparison with \last{} to satisfy G1.
A31 changed only the decision date, from the planned date to completion of all specified experiments.
All criteria and their outcomes are reported, including the unmet G2(b).

The original criteria and A9 therefore preceded all results from the eight added detectors.
A14 was written after their runs had started but before any aggregate over them was computed, and it only made G1 stricter.
All criteria followed the pilot results from the other four detectors.
The four-detector pilot, its comparison of label-free checkpoint selectors, and its initial GX structural blind-spot observation are exploratory.
Analyses added after a method failed its reproduction check are also labeled post hoc.

\subsection{Follow-up experiments in the second version}
\label{app:prereg-v2}

The second version adds eight follow-up studies, identified as A, C, C2, D, E, F, G and H in the amendments.
B was reserved for Type III OGB pairs, which were not run.
A36 adopted written plans for D, A, C and E before their runs.
A file-format check had displayed one typed \last{} value outside the A9 decision, but no other typed \last{} or \oracle{} value had been read.
A40 added C2 before any extra-seed result, selecting its subset only by runtime and training objective.
A42 added F, G and H before any result from A, C, C2 or E and before any score from the new references.

Table~\ref{tab:prereg-v2} summarizes each follow-up question, its scale, the amendments specifying it, and its analysis status.
Run and first-analysis times come from logs and scheduler accounting and are EDT on 2026-10-16, except D on 2026-10-15.
A43 and A46 were written after the results they concern, as explained below.

\begin{table}[!htb]
  \centering\footnotesize\setlength{\tabcolsep}{3.5pt}\appRaggedCells
  \caption{Scope and timing of the follow-up studies.}
  \label{tab:prereg-v2}
  \begin{tabular}{@{}p{0.55cm}p{4.8cm}p{1.3cm}p{1.75cm}p{1.75cm}p{1.15cm}p{2.65cm}@{}}
    \toprule
    Part & Question & Data & Specified in & Runs & First stat. & Status \\
    \midrule
    D & Does the blind spot depend on the checkpoint rule? & Typed runs & A36 & & 23:00 & Post hoc \\
    A & Does leakage replicate in Type III? & 600 runs & A36, A37 & 02:12--04:22 & 09:18 & Decision criterion R1 \\
    C & How much does leakage vary over seeds 1 and 2? & 960 runs & A36, A41 & 04:21--09:49 & 09:52 & Descriptive \\
    C2 & Does leakage stay positive over seeds 3 to 5? & 180 runs & A40, A41 & 04:21--09:49 & 09:52 & Descriptive \\
    E & What does selection on a separate labeled half add? & Stored scores & A36, A45, A46 & & 10:21 & Post hoc \\
    F & Do test passes change training? & 120 runs & A42, A44, A47 & 10:36--12:45 & 12:51 & Descriptive \\
    G & Does the 20\% hold-out lower \last{}? & 60 runs & A42, A44, A47 & 10:36--12:45 & 12:51 & Descriptive \\
    H & Which typed changes do low-order statistics detect? & No training & A42, A43 & & 03:45 & Descriptive \\
    \bottomrule
  \end{tabular}
\end{table}

R1, the only follow-up test with a decision criterion, tests checkpoint leakage on five Type III TU pairs using the G1 thresholds.
We first average five trials within each pair and then weight pairs equally.
Both \oracle{} minus \valnormal{} and \oracle{} minus \last{} must average at least 3.0 points, with gaps of at least 2.0 points for at least 75\% of complete detectors.
A detector enters the decision only if it completes all five pairs under UB-GOLD settings.
At least eight such detectors are required.
All 12 are complete here, so 75\% corresponds to nine detectors.
A36 stated nine of 12, and A37 clarified before any Type III grid run that the requirement refers to the G1 fraction of complete detectors.

The plans also specified implementation choices.
SIGNET uses its Type III script values, correcting and marking only two misspelled flags.
The four pilot detectors use CPU for the extra seeds and their CPU runs as seed 0.
REDDIT-B receives no extra seeds because of cost.
Type III index lists were fixed and their sha256 hashes recorded before execution.
Type III runs for OGB pairs and recent methods would be added only on request.

A37, A41, A44 and A47 specified details that the plans had left open, before the corresponding study produced any result.
A45 fixed the run set for E after the results of A, C, C2 and H had been read, but before any E quantity was computed.
A43 and A46 followed their related results, recording the ridge term from the existing log-count controls for H and the rounding, random draws, AUROC arithmetic and per-seed reporting for E.
According to the study record, these choices were already in code saved before execution, at 02:59 before the 03:45 run for H and at 10:18 before the 10:21 run for E.
Neither later record changes a result.

A48, its correction and A49 concern presentation and add no run, test or decision rule.
Written on 2026-10-19 at 16:46 EDT after all follow-up results had been read, A48 retains G1, G2(a)--(c) and R1 and specifies three summaries of existing numbers before plotting.
The only derived quantity is the sum of two reported E differences.
The threshold curve is labeled post hoc.
The 17:49 correction fixes a criterion name.
A49 was written at 22:11, after all results and the full draft.
It allows descriptions of experimental coverage and procedures to be shortened according to their relevance to the research questions, but not according to result direction.
All results of the formal tests and R1, exceptions, limitations, timing disclosures, deviations and audit findings must remain, with a recorded destination for material moved elsewhere.

The first version has external registrations, but the written follow-up specifications have no external timestamp preceding their results.
The timing of the follow-up studies is supported by run logs and private-repository commits made within five minutes of each amendment.
We therefore describe follow-up analyses planned before their results as specified in advance, not as pre-registered.

\subsection{Amendments that affect interpretation}
\label{app:prereg-amend}

Every change after the original plan has a dated amendment stating its reason.
From 2026-10-12 at 21:37 EDT, amendments also record the clock time and test metrics then available.
A1--A8 were written together at 21:17:22 EDT, before any added-detector metric.
Their initially estimated times were later than the writing time and were corrected to the file time without changing the text.
Table~\ref{tab:amend} groups amendments by what they changed.
The supplementary study record summarizes every amendment.

\begin{table}[!htb]
  \centering\footnotesize\setlength{\tabcolsep}{3.5pt}\appRaggedCells
  \caption{Amendments grouped by role.}
  \label{tab:amend}
  \begin{tabular}{@{}p{2.1cm}p{2.65cm}p{0.75cm}p{2.15cm}p{7.15cm}@{}}
    \toprule
    Role & Amendments & Oct. & Effect on criteria & Timing relative to related results \\
    \midrule
    Record corrections & A0, A24 & 12, 13 & None & A0 records a timing discrepancy. A24 retracts an A23 factual error after GLASS's reproduction table \\
    Rule implementation & A1--A8, A12, A15, A26 & 12--13 & None & Before added-detector metrics (A1--A8), G1 aggregates (A12, A15) or real-data statistics (A26) \\
    Scope changes & A9--A11, A17, A28 & 12--13 & A9 changes evaluation of G2(c) & Before affected runs \\
    Anomaly generators & A9.1, A25, A28.1 & 13 & None & Before any deep detector ran on the generator concerned \\
    Implementation decisions & A13, A16, A18--A23, A27, A29, A30 & 13--14 & None & Port notes precede each method's first metric. A18, A22 and A27 resolve issues raised by results \\
    Decision rule & A14, A31 & 13, 14 & Stricter G1 (A14), date only (A31) & A14 follows pilot aggregates but precedes the 12-detector aggregate. A31 follows provisional statistics \\
    First-version analyses & A32--A35, A38, A39 & 14--16 & None & A32 follows first statistics but precedes its analyses. A33--A35 precede and A38--A39 follow all results \\
    Follow-up plans & A36, A37, A40--A42, A44, A45, A47 & 15--16 & A36 adds R1 & Before results of the governed part. A44 and A45 follow Type III results \\
    Decisions recorded later & A43, A46 & 16 & None & After results, with choices in code saved before execution \\
    Presentation & A48, its correction, A49 & 19 & None & After all results \\
    \bottomrule
  \end{tabular}
\end{table}

No decision criterion changed after its corresponding result was computed.
Alongside A14, two generator amendments affect interpretation.
When only generator yields and checks without deep detectors were available, A25 reduced the required number of anomalies per seed from 150 to 100.
This change used the precision needed to distinguish AUROC 60 from 85.
A28 retired the first molecular generator because graphlet count + LW remained below 85 on every type.
The only deep-detector outputs on typed anomalies then came from eight-epoch smoke runs.
These outputs had not been inspected, and an automatic check had tested only whether the values were finite.
The replacement generator had to be built without access to deep-detector results on typed anomalies and fixed before any deep detector ran on it.
MS-v2 met these requirements under A28.1.

Other amendments recorded after related results changed no criterion, threshold or selector.
Because results already existed, A18 rejected an alternative and retained the pre-registered \hpfree{} definition for LGKDE.
A22 marked a reproduction check not evaluable when no comparable published setting was found, and A27 applied the existing non-finite-score rule to HC-GLAD.
A32 was written after the frozen statistics code first produced output and after the first draft, labeled all its still-uncomputed analyses post hoc, and was committed within a minute.
A38 and A39 only changed wording and added a post hoc table.

\subsection{Registrations and timing evidence}
\label{app:prereg-ledger}
\label{app:prereg-verify}

The first version registered hashes of the study files with OSF at four milestones.
Each registration records sha256 hashes of the study files, including the pre-registration and the amendment file.
OSF recorded M0 on 2026-10-13 at 18:54 UTC, M1 on 2026-10-14 at 00:11 UTC, M2 that day at 04:14 UTC, and M7 on 2026-10-16 at 01:48 UTC.
M0 fixes the plan and A0--A21 before any added-detector G1 aggregate was inspected.
M1 precedes every pooled statistic from the frozen analysis code, M2 precedes every deep typed run, and M7 records the final results of the first version.

The registrations were embargoed until the first version appeared and are now public.
The released pre-registration is byte-identical to the original and can be checked against M0, and de-identified run outputs are public in a project on the same registry.
\ifanon
\else
The four registrations are \url{https://osf.io/2zq4h}, \url{https://osf.io/d9p8j}, \url{https://osf.io/e93q7} and \url{https://osf.io/ckuz4}.
\fi

Between milestones, amendments were committed to a private repository within minutes, and run logs record each job time and purpose.
These records allow amendment times to be compared with the jobs producing the affected results.
External registrations provide public timestamps for the registered hashes, whereas later follow-up timing relies on private records.

\section{Implementations and reproduction checks}
\label{app:ports}

This appendix records how our implementations differ from released code and whether their results match the published values.
Checkpoint comparisons within one run use the same training process.
Agreement with published results also depends on whether the implementation and configuration match the original method.
We first cover UB-GOLD and the nine recent methods, then version and device differences.
U1--U25 identify audit entries, and A0--A49 identify study amendments.

\subsection{UB-GOLD detectors}
\label{app:ports-ubgold}

\paragraph{Execution and recording.}
We retain an unmodified, MIT-licensed copy of UB-GOLD at c940457.
Our shared execution script runs explicitly marked, line-by-line copies of the training loops and calls the released models, scorers and helper functions.
The four pilot detectors use the loops from the pilot code.
For OCGIN and GLocalKD on BZR fold 0, CPU runs reproduce the pilot results bit for bit.
Splits come directly from UB-GOLD and are checked against cached indices, with trial $i$ using fold $i$ and seed $3407+i$.

Each fold holds out 20\% of the normal training graphs with a fixed seed and fits the model only on the remainder for the full budget, without early stopping.
We follow UB-GOLD's evaluation schedule to record test metrics and held-out normal-graph scores.
We also record per-graph test scores, except in the four pilot detectors' main-comparison runs (Appendix~\ref{app:followup}).
The added scoring of held-out graphs uses no gradients.
We save all random-number states before this scoring step and restore them afterward.
We retain random operations in the original test evaluations, so they still consume random numbers from the generator used for training.

\paragraph{Selection and scoring.}
We reconstruct \oracle{} from the saved run record, updating its checkpoint only when test AUROC strictly increases.
Ties retain the earlier checkpoint, so floating-point details can affect the selected epoch.
The released eight end-to-end detectors stop after 30 evaluations without improvement, whereas the four two-stage detectors do not stop early.
Our replay of this rule applies the 30-evaluation stop to the saved scores of all 12 detectors, including the two-stage detectors.
Without the stop for the two-stage detectors, the mean checkpoint gaps of the 12 UB-GOLD detectors on UB9 would be 5.13 instead of 5.12 points against \last{} and 4.14 instead of 4.13 against \valnormal{}.
We retain CVTGAD selection on normalized scores and reporting on raw scores (U15, A4).
We negate GraphDE scores because its paper defines them as probabilities of being normal (U14, A5).
For two-stage detectors, we pair each graph score with the label of the same graph.
This corrects the original code, which collects scores and labels in separate shuffled passes (U5, A6).

\paragraph{Hyperparameters.}
Tuned settings use the values in UB-GOLD per-dataset scripts.
For GraphDE on REDDIT-B, we omit the dangling flag that lacks a value and prevents parsing (A2).
Defaults reset all options except batch sizes to repository defaults (A1).
InfoGraph-OCSVM crashes at the default layer count, so it uses two layers, the most frequent value in the released configurations (U19).

\paragraph{Reproduction.}
The specified reproduction check requires tuned \oracle{} to fall within $\pm5$ AUROC points of the published value on at least two datasets, using the UB-GOLD score direction for GraphDE.
Across the 13 datasets with published values, GraphCL-OCSVM matches on 13, CVTGAD on 12, GOOD-D, GraphCL-IF, OCGIN, OCGTL and SIGNET on 11 each, GLADC on 10, and GLocalKD on nine.
GraphDE matches on eight, or four in its paper orientation, and each InfoGraph detector matches on three.
All 12 detectors meet this check, which does not imply close agreement on every result.

The following exceptions show where our results differ from the published values.
In the pilot, training on 80\% of normal graphs lowered SIGNET \oracle{} by about 0.8 points.
GLADC completes only two COLLAB folds because the other three diverge.
InfoGraph exceeds published values on most datasets, by up to 38.14 points, while several published values are below 50, consistent with the U5 label misalignment.
Replaying misaligned evaluation on the first datasets checked gives selected values of 56 to 67, although GraphCL, which shares the loop, matches most published values.
Other single-dataset differences reach $-37.19$ points for GLocalKD on DD, and the full comparison remains in the released results.
If an implementation error shifts all selection-rule results in a run by the same amount, it cannot explain differences between those rules.

\subsection{Recent methods}
\label{app:ports-recent}

\paragraph{Inclusion.}
\label{app:ports-inclusion}
The recent-method code survey closed on 2026-10-13 at 00:45 EDT, before any surveyed method ran.
It covered graph-level anomaly and out-of-distribution detection, mainly from 2023--2026, including preprints and UB-GOLD-listed methods, with code availability checked on October 12--13.
Coverage is incomplete because two bibliographic services blocked automated queries and anonymous submissions could not be checked.
The first tier prioritized 2024--2026 main-venue methods whose official code trains without anomaly labels and supports one-class evaluation.
The second tier included other venues or preprints, different native settings and costly builds.
We ported all methods in both tiers except CLINIC, added each through A10 or A11 before its runs, and subsequently removed none.

Table~\ref{tab:ports-inclusion} lists the decisions for all 46 candidates.
It records the survey coverage and does not judge the performance of methods we did not evaluate.
\begin{table}[!htb]
  \centering\footnotesize\setlength{\tabcolsep}{4pt}\appRaggedCells
  \caption{Inclusion and exclusion in the method survey.}
  \label{tab:ports-inclusion}
  \begin{tabular}{@{}p{2.55cm}p{13.85cm}@{}}
    \toprule
    Outcome & Candidates and reasons \\
    \midrule
    Included & A10 added MUSE (NeurIPS'24), LGKDE (ICML'26), GLADPro (KDD'25), SEGO (AAAI'25) and GLADMamba (ECML-PKDD'25). The first four were first-tier methods. GLADMamba already supported UB-GOLD's folds. A11 added second-tier GLASS (arXiv'26), ProtoGLAD (arXiv'26), HC-GLAD (Neural Netw.'26) and SUL/GUL (WWW'26), the latter without contamination on normal-only folds. \\
    Code read only & CLINIC (ICML'26), whose custom CUDA extensions made its build costly. \\
    Optional, not added & HimNet (ECML-PKDD'23) predates 2024 and evaluates the final epoch. GRAM (Neural Netw.'24) releases a demonstration outside these datasets and selects on training loss. FANFOLD (arXiv'24) does not run as published. RedOUT (NeurIPS'25) and BaCa (AAAI'26) lack anomaly-detection pretrained backbones. GOODAT (AAAI'24) has no training script and reuses a checkpoint across folds. FGS-GLAD (AAAI'26) targets distribution shift. DO2HSC's code address returns 404. \\
    No released code & M-GLAD (IJCAI'26, supplementary PDF only), NGLAD (TIFS'26), GenGLAD (NeurIPS'26), SDM (ICML'25), AGDiff (ICML'25), CGOD (WWW'25), PGOS (AAAI'26), SIGOOD (AAAI'26, README only), UniFORM (AAAI'25), TUAF (DASFAA'23). \\
    Supervised & GADTrans (TKDE'26), G-GLAD (WWW'26), GLADformer (ECML-PKDD'24), UniGAD (NeurIPS'24), RQGNN (ICLR'24), iGAD (NeurIPS'22) and GmapAD (KDD'23) use anomaly labels. SGOOD (CIKM'24) uses in-distribution class labels. \\
    Not checked & BNFlow, PopGLAD and MSRE (ICLR'27 submissions) had inaccessible pages. DGPA, X-GAD, MoEGAD, ODGNN, AAGOD, FedCIGAR (IJCAI'26) and D2GO were found but not read before the survey closed. \\
    \bottomrule
  \end{tabular}
\end{table}

\paragraph{Shared execution rules.}
We keep unmodified copies of the official repositories at the stated commits.
All changes use runtime overrides or explicitly marked copies in our adapters.
Adapters use the same splits, held-out graphs, seeds and record format.
Each method scores held-out graphs with its own scorer, without changing the training random-number states.
Tuned settings use the released per-dataset configurations.
For a missing option, we use its most frequent value in the released configurations.
Default settings use repository values, with specified alternatives for crashes or memory limits.
Both tuned and default settings were fixed before the method first produced a test metric.
\oracle{}-own replays each method's released rule, while checkpoint-selection gaps averaged across datasets and detectors consistently use the UB-GOLD rule (A26).

All runs retain the UB-GOLD anomaly class, which several papers reverse on BZR, COX2 and PROTEINS\_full.
The specified reproduction check requires the tuned own-rule five-fold mean to fall within $\pm5$ points of the published value on at least two comparable datasets, excluding references with a different anomaly class.
The check does not determine inclusion, and we found no porting error that would exclude a method.
Unlicensed code is run but not redistributed.
Six methods meet the check, LGKDE does not, GLADPro is contested, and SUL/GUL is not evaluable.

Table~\ref{tab:ports} lists each method's configurations, code changes and reproduction-check results.
The categories below consider code changes, available configurations and reproduction checks together.
\begin{table}[!htb]
  \centering\footnotesize\setlength{\tabcolsep}{3.5pt}\appRaggedCells
  \caption{Implementation and reproduction of the nine recent methods.}
  \label{tab:ports}
  \begin{tabular}{@{}p{1.8cm}p{4.0cm}p{5.3cm}p{4.9cm}@{}}
    \toprule
    Method & Tuned setting & Changes in our runs & Reproduction check and tier \\
    \midrule
    MUSE\newline 07da392, none & Released labeled-selection values. Modal for COX2, ENZYMES, MUTAG & Second-stage autoencoder with 500 epochs, width 32, learning rate $10^{-3}$. Adjacency target sized by node count & Met with own code and splits, 4/5. UB-GOLD folds without labeled selection, 2/7. \emph{Incomplete configuration} \\
    LGKDE\newline 1bba5ef, MIT & MUTAG, the only released configuration & DGL-to-PyG port. GPU copy semantics on every device (A18) & Not met, 1/4 (MUTAG), including own pipeline. \emph{Reproduction not met} \\
    GLADPro\newline 1f811b9, MIT & README for MUTAG, IMDB-B, PROTEINS (used for PROTEINS\_full). Modal elsewhere, defaults except 500 rather than 1000 epochs & Two forward-value-preserving compatibility edits. Deterministic CPU kernels & Contested, 0/3 against a third-party re-run. \emph{Incomplete configuration, or not met} \\
    GLADMamba\newline 2b2462c, none & Released scripts. Modal for PROTEINS\_full, MUTAG & No model changes. Unmodified commit archived from a public mirror & Met, 6/8. Both misses match under its released protocol. \emph{Incomplete configuration} \\
    SEGO\newline 45bdda7, MIT & No detection configuration. Modal values from 10 pair lines & Three PyG API edits, with message passing checked independently. Pair-loader batching option & Met, 2/5. Own pair task with released lines, 4/5. \emph{Incomplete configuration} \\
    GLASS\newline 3bbba39, MIT & Single released configuration, equal to defaults (A21, A23) & No model or scoring changes. All 150 epochs trained and stopping replayed. Inductive default scorer for label-free columns (A24) & Met, 5/7. Own pipeline reaches paper values on MUTAG, AIDS, IMDB-B only with test-selected scoring. \emph{Native} \\
    ProtoGLAD\newline f7d8ce9, none & Replay of 144-cell test-AUROC search. README for MUTAG & Reseed each cell and fold (A20). GPU operations run unchanged on CPU & Met, 3/4. Own protocol, 3/3. \emph{Minimal patch} \\
    HC-GLAD\newline 4238b20, none & Released scripts. Modal for NCI1, IMDB-B, MUTAG & Memory fallbacks: test batch 16 on NCI1, width 8 and two layers on REDDIT-B. Two non-finite tuned runs handled under A15 (A27) & Met, 7/7 against preprint. Modal IMDB-B 50.77 versus 60.92. \emph{Incomplete configuration} \\
    SUL/GUL\newline 1a1f2e4, none & Demonstration configuration, defaults with 500 epochs & Anchor count capped at training-set size for BZR, COX2, MUTAG (A19) & Not evaluable (A22). Post hoc paper-setting check, 2/2. \emph{Incomplete configuration, or not met} \\
    \bottomrule
  \end{tabular}
\end{table}

\paragraph{Implementation and reproduction categories.}
\label{app:ports-fidelity}
Each method is assigned to the lowest applicable category in the order below.
In the \emph{native} category, model, loss and scoring code are unchanged.
We add only data roots, splits, seeds, hold-out, evaluation without gradients, recording and resuming.
A \emph{minimal compatibility patch} addresses current libraries, CPU execution or crashing inputs, whereas a \emph{framework port} changes libraries.
Missing UB9 configurations or components incompatible with one-class use imply an \emph{incomplete configuration}, and a failed check implies \emph{reproduction not met}.
GLASS is native, and ProtoGLAD has a minimal compatibility patch.
GLADMamba and HC-GLAD have native code, and MUSE, GLADPro, SEGO and SUL/GUL have minimally patched code, but all six have incomplete configurations.
LGKDE is a framework port with a failed check, and a stricter interpretation also places GLADPro and SUL/GUL in the lowest tier.
The mean checkpoint gap for recent methods remains positive under both interpretations, but its size depends on which uncertain reproductions are included (Appendix~\ref{app:posthoc-fidelity}).

\paragraph{Reproduction results requiring further explanation.}
MUSE uses labeled anomalies to select the width, learning rate and stopping time of its second-stage autoencoder, labels unavailable in the one-class folds.
We therefore fixed the table values before any run, and equivalence checks against the unmodified functions give identical snapshots and test scores.
A diagnostic replay substituting test labels for validation labels brings four of seven datasets within $\pm5$ points but never enters the reported results.
The diagnostic indicates that labeled selection accounts for most of the gap.

The LGKDE DGL-to-PyG port gives bit-identical losses and scores at each of ten epochs on MUTAG and BZR.
The vectorized grid loop computes the same function, with relative loss differences at most $1.2\times10^{-15}$ over ten epochs.
Because the original CPU code swaps negative features in place, all devices use the copy semantics of the GPU required by its README.
The only released configuration uses learning rate 0.01, whereas the paper uses 0.001.
Both our implementation and the native pipeline match only MUTAG within $\pm5$ points, at 88.21 versus 91.63.
A post hoc run with the paper learning rate does not close the AIDS, DHFR and NCI1 gaps.
Tuned and default settings differ only in feature noise and stopping parameters, so this comparison tests only those choices.

The GLADPro paper was unavailable when the reproduction check was specified and run, so it used a third-party re-run with matching train and test sizes and was not repeated after the paper became available.
A literal count gives two of seven matches, but anomaly classes likely differ on COX2 and normal classes on ENZYMES.
On the three class-matched datasets, differences are $+5.40$ on MUTAG, $-11.03$ on IMDB-B and $-22.75$ on AIDS, making the check contested.
Another re-run with unspecified anomaly class matches four of five datasets.
The native code and splits give 87.08 on MUTAG and 62.45 on IMDB-B, close to our 87.48 and 59.59, so these gaps do not arise from the port.

Only GUL is released for SUL/GUL, and its contamination handling has no effect on normal-only training folds, so this evaluation does not test the paper's main claim.
The paper reports GUL only with fully contaminated training, making the primary reproduction check not evaluable.
Post hoc runs in that setting give 90.40 versus 93.38 on MUTAG and 66.34 versus 69.84 on IMDB-B.
We retain the released 500 epochs, whereas the paper specifies 1000 and unreleased grid-searched hyperparameters.

\subsection{The earlier UB-GOLD commit}
\label{app:ports-a17}

The UB-GOLD published values most likely precede the external pull request that changed batching for GLocalKD and GLADC.
A17 specified five-fold UB9 re-runs for both methods at pre-merge commit a96137b, totaling 90 runs.
Two parsing edits preserve behavior under A30, and the common runner uses the earlier loops that rebuild shuffled batches every epoch.
Each run uses the same device as its paired c940457 run.
CPU differences therefore also reflect the use of different random numbers, and GPU differences may include run-to-run variation.
This descriptive analysis enters neither G1 nor G2, and all 90 runs finish with paired mean-AUROC differences at most 0.22 points (Appendix~\ref{app:extra-version}).

\subsection{Devices and bit-level reproducibility}
\label{app:ports-devices}

Device limitations affect reproduction of a run, while all selection rules still read the same run record.
Fixed-seed CPU re-runs give identical losses and scores, and resuming is bit-identical for all ten UB-GOLD loop families and all checkpointed recent adapters, apart from GLASS's intrinsic randomness.
GPU smoke checks repeat one seed-0 cell for each of six loop families, with only GLADC bit-identical in both re-running and resuming.
CVTGAD, GOOD-D, GraphCL-OCSVM, GraphDE and InfoGraph-IF differ within the first few epochs, likely from nondeterministic GPU accumulation, although individual operations were not verified.

This variation limits reproduction of the main-experiment GPU results and also contributes to differences across GPU training seeds.
GraphCL-OCSVM on PROTEINS\_full differs from epoch 1 or 2 even on identical GPU re-runs.
Analyses F and G therefore flag its five folds and also report averages excluding this detector and dataset combination (A47).
Across devices, single GLocalKD folds differ by up to 31 points because dropout uses different random streams, and changing the ProtoGLAD device similarly acts like changing its seed.
With SciPy 1.15 and later, the GLASS sparse eigensolver draws a new starting vector, producing differences in 32 of 50 identical-configuration pairs.

Two tuned GraphDE main-grid runs on AIDS, folds 0 and 4, resumed on CPU after GPU epochs 385 and 384 of 400 and cannot be reproduced bit for bit on a single device.
The seed analysis uses GPU.
For SIGNET, OCGTL, OCGIN and GLocalKD, seed 0 uses the bit-identical CPU pilot path because the main-grid GPU runs did not store per-graph scores.
This choice preceded the extra-seed runs under A36, and the main results retain their original values.
Together with omitting REDDIT-B, this explains why seed-0 gaps of 4.49 against \valnormal{} and 5.48 against \last{} differ from the main-grid 4.13 and 5.12.

\providecommand{\audY}{$\bullet$}
\providecommand{\audN}{$\circ$}
\section{Code and data audit}
\label{app:defects}

This appendix gives the code and data evidence for Table~\ref{tab:audit}.
It records where selection uses test labels and which data or scoring problems affect the results.
The audit covers UB-GOLD at c940457, the nine re-evaluated recent methods, and five additional code bases.
Findings concern the stated versions of released code and do not establish that published results followed those paths or imply author intent.

Entries retain the audit-log identifiers U1--U25 and T1--T12.
\emph{Code} denotes inspected source, \emph{paper} the specified paper version or UB-GOLD public review record, \emph{history} commits, issues or pull requests, and \emph{data} our counts or runs on released data and code.
We rechecked every retained entry against the stated source version or paper.
When we cannot establish where a published value came from, we mark its origin as unknown.
The tables give the evidence for each finding and explain how we handle it.
The supplementary study record lists the full file paths and line numbers.

\subsection{UB-GOLD's Types I and II}
\label{app:defects-ubgold}

\paragraph{The shared checkpoint path.}
In U1, line 432 of \path{dataloader/data_loader.py} assigns the test loader to the validation loader, as do the Tox21 and out-of-distribution paths.
The evaluations called validation therefore use test labels.
In the history, f6be7ea first added a normal-only validation split on 2024-05-06, for which AUROC is undefined.
Later that day, f805ad7 added selection by the highest validation AUROC, 2e7ab46 replaced the split with the test loader on May 7, and the template covered all 12 detectors by May 27.
The selection path thus comes from the shared UB-GOLD template and need not match the original implementations of individual detectors (code, history).

In U2, each detector computes test AUROC at scheduled evaluations, saves the highest-scoring checkpoint, reloads it for prediction, and replaces it only on strict improvement.
The eight end-to-end detectors also stop after 30 evaluations without improvement, while the four two-stage detectors do not stop early.
On UB9, this selection averages 4.13 points above \valnormal{} and 5.12 above \last{}.
The paper does not describe this rule, and the ICLR 2025 proceedings version, both arXiv versions and their appendices contain none of the terms epoch, checkpoint, early stopping, patience or model selection.
Section 3.3 describes metrics, splits and test-set tuning, while Appendix D.3 describes software, hardware, search space and efficiency.
The 15 records returned by the public review search, comprising the submission, four reviews, nine comments and the meta-review, also omit the rule (code, paper).

Tables~\ref{tab:adf-ubgold} and~\ref{tab:adf-ubgold2} separate selection and scoring issues from data, reporting and runnability issues.
Each entry states the released behavior and its effect on this study, with file-level locations in the supplementary study record.

\begin{table}[!htb]
  \centering\footnotesize\setlength{\tabcolsep}{3.5pt}\appRaggedCells
  \caption{UB-GOLD paths affecting selection and scoring.}
  \label{tab:adf-ubgold}
  \begin{tabular}{@{}p{2.5cm}p{7.7cm}p{6.15cm}@{}}
    \toprule
    Entry and evidence & Released behavior & Effect and handling \\
    \midrule
    U3. Test-set tuning\newline paper, code & Random search on test performance, explicitly described as an upper bound in Section 3.3. Selected values appear in per-dataset scripts & Tuned settings add 1.53 points over defaults under \valnormal{}, 8.76 for OCGTL. Kernel searches add 1.24--3.60 points. \\
    U15. Different selection and reporting scores\newline code & CVTGAD selects on $(g-\mu_g)/\sigma_g+(n-\mu_n)/\sigma_n$ with adaptive scoring enabled, but returns $g+n$ & \oracle{} replays this difference (A4). \\
    U5. Misaligned labels\newline code, data & KernelGLAD and the four two-stage detectors collect scores and labels in separate passes over a shuffled test loader & Unchanged WL-IF averages 49.83 over 65 folds. Twenty-one of 26 IF values and 24 of 26 OCSVM values satisfy the chance band $|z|\leq3$. We align labels (A6). \\
    U6. Unreproduced published kernel values\newline data & PK-IF on DD is 75.29 ($z=14.78$). WL-IF on DD is 70.49 ($z=11.97$). WL-IF on PROTEINS\_full is 65.75 ($z=8.91$) & Aligned test-selected search reaches 47.68, 65.04 and 60.17, respectively. Released WL-IF gives 48.20 on DD and 49.48 on PROTEINS\_full. Published-value origins remain indeterminate. \\
    U12. Test-set fitting\newline code & KernelGLAD fits its kernel and IF or OCSVM on the test loader, ignoring the training loader & Released results are transductive. Ranked rows fit normal training graphs only. \\
    U14. GraphDE scoring and flags\newline code & Normal probability used as anomaly score. A tuple assignment enables BatchNorm. Test-time negative sampling. REDDIT-B script flag lacks a value & We negate scores (A5), retain the released orientation for reproduction, and omit the dangling flag (A2). Test scores remain stochastic. The released script cannot produce its published REDDIT-B value. \\
    U19. Two-stage fitting\newline code & IF or OCSVM refits per batch, leaving the final batch's fit. InfoGraph fits projected but scores unprojected embeddings. Its OCSVM input width requires two layers & We retain these behaviors because they define UB-GOLD's detectors. \\
    U23. Batch-dependent scoring\newline code & Graph-level contrastive losses in CVTGAD, GOOD-D, and SIGNET normalize over all graphs in the test batch. CVTGAD and GOOD-D also allocate an $N\times N$ node-loss matrix for all $N$ batch nodes, with comparisons masked to the same graph. Default test batch size 9999 puts the full test set together & Graph scores depend on batch composition. The node-loss matrix in CVTGAD and GOOD-D requires memory quadratic in the batch's total node count. Typed groups are each scored as one batch. \\
    U21. Evaluation changes training randomness\newline code & Test shuffling, GraphDE negative sampling and GOOD-D's shuffled training-loader pass consume the training random stream & We retain the released schedule and draws, then measure their effect with isolated test evaluations. \\
    \bottomrule
  \end{tabular}
\end{table}

\begin{table}[!htb]
  \centering\footnotesize\setlength{\tabcolsep}{3.5pt}\appRaggedCells
  \caption{UB-GOLD data, reporting and execution issues.}
  \label{tab:adf-ubgold2}
  \begin{tabular}{@{}p{2.5cm}p{7.7cm}p{6.15cm}@{}}
    \toprule
    Entry and evidence & Released behavior or data property & Effect and handling \\
    \midrule
    U7. Majority-class anomalies\newline code, data, paper & Class 0 is anomalous throughout, but is the majority in BZR (319/405), COX2 (365/467), PROTEINS\_full (663/1113), DD (691/1178) and COLLAB (2600/5000), contrary to Appendix B's minority-class description & The task differs from its description and scores may fall below 50. We retain UB-GOLD's class for comparability. \\
    U8. Inverted Tox21 tasks\newline code, data & First-appearance label mapping makes HSE and PPAR-gamma, whose training files start with an inactive molecule, train on 423 and 219 active molecules, respectively, while test labels still call them anomalous & These two tasks are excluded because their results cannot be interpreted. \\
    U10. WL-identical classes\newline data & Across five folds, 152/500 IMDB-B test anomalies are WL-identical to a training normal. In the dataset, 318 graphs match a graph of the other class. No fold shares graph indices between train and test & This data property limits WL-bounded detectors. It is not a split error. \\
    U11. Size separates classes\newline data & Node count alone scores 99.2--100.0 on every AIDS fold and 29.5--34.9 on every NCI1 fold & A high score can reflect size, and a one-sided size score can reverse across datasets. \\
    U13. Constant propagation kernel\newline code & IMDB-B, REDDIT-B, COLLAB and Tox21 use constant node features & The PK matrix is constant and every PK result is 50 by construction. \\
    \midrule
    U9. Variance reported as $\pm$\newline code & The reporting function returns \texttt{statistics.variance} times 100 & Interpreting it as standard deviation understates fold variation. We report standard deviations. \\
    U4. Search scripts fail\newline code & A U+2019 quotation mark causes an import syntax error. Kernel search also reads undefined \texttt{model.max\_AUC} & Released scripts cannot recover the published hyperparameters. We emulate the stated kernel search. \\
    U17. GLADC memory and divergence\newline code, data, history & Whole-batch label conversion requires test batch size 1. Maximum-size padding and preprocessed batches raise memory to 229\,GB on one REDDIT-B run. COLLAB folds 1, 3 and 4 diverge with NaN standard deviations under both settings and on CPU and GPU & Divergence does not arise from tuned settings. We mark these runs diverged and impute nothing (A12). \\
    U20. Unrunnable benchmark components\newline code, history & PR~\#3 comments out the Type IV loader, still called by the entry point and all 16 drugOOD/GOOD script lines. README-listed GOODAT and AAGOD are not implemented & Part of the benchmark cannot be reproduced, so Type IV is excluded. \\
    \bottomrule
  \end{tabular}
\end{table}

Three other verified entries affect only execution or concern features unused by the evaluated detectors.
U16 repeats full-training-set k-means for every GOOD-D batch and fails with CPU faiss.
T5 takes Type III padding size from the entire second dataset, reaching 620 nodes for ENZYMES+PROTEINS.
T6 retains edge attributes in Type III training and removes them at test time, but none of the 12 detectors uses them.
In this study, U1 and U2 identify checkpoint selection using test labels, and U3 identifies the disclosed hyperparameter selection using test labels.
We align graph scores and labels, retain the UB-GOLD anomaly class, and exclude the two inverted Tox21 tasks.

\subsection{Pull request \#3 and the earlier commit}
\label{app:defects-pr3}

The UB-GOLD published values date from 2024, whereas c940457 includes external PR~\#3, merged on 2025-08-19 as 63bc598 over a96137b of 2024-10-02.
Changes introduced by that pull request should therefore be distinguished from the original benchmark.
U18 includes one-time batch construction without subsequent reshuffling for GLocalKD and GLADC, and hard-coded CUDA transfers.
GLocalKD also references \texttt{preprocessed} in its decision function outside the \texttt{fit()} scope where it is defined, causing a NameError.
The PR also adds U17 preprocessing, comments out the U20 loader, supplies abstract-method \texttt{pass} bodies in \path{mybase.py}, and corrects the SIGNET readout flag.

As of 2026-10-13, the repository contained only issue \#1 of 2024-08-08 on incomplete \path{mybase.py} and issue \#2 of 2024-11-21 on installation, with HEAD still c940457 of 2026-06-03.
None of the findings above had been reported there, and the PR did not disclose the U18 and U20 behaviors it introduced, recorded as U22.
The pre-merge a96137b also fails to parse because five abstract methods are empty and loader line 423 contains a stray \texttt{1}, recorded as U24.
Bytecode committed in May 2024 records different source sizes, and GLocalKD bytecode lacks \texttt{preprocessed}, consistent with earlier per-batch processing.
We therefore infer, but cannot confirm, that the 2024 numbers came from a different local tree.

The PR also changes when loaders consume random numbers, recorded as U25.
At the same CPU seed, c940457 consumes two extra random numbers for GLADC and four for GLocalKD before the first training step.
Dropout and GLADC parameter noise then differ immediately, and batches differ from epoch two.
GPU random draws use the CUDA generator and are unaffected by this particular shift.
Current-version runs therefore differ from the procedure most likely used for published values, but the 90 earlier-version runs give paired mean differences of at most 0.22 points and do not explain the reported leakage (code, history, data).

\subsection{Selection and dataset construction in Type III}
\label{app:t3-code}

Type III trains on an in-distribution dataset and uses graphs from a second dataset as test anomalies.
These experiments repeat the checkpoint-leakage test in a different setting, but their dataset construction also limits what AUROC can show.
Table~\ref{tab:adf-t3} lists the selection rules separately from the dataset properties.
PROTEINS denotes the TU dataset of that name.
T9 concerns paper text not verified against the PDF and is omitted.

\begin{table}[!htb]
  \centering\footnotesize\setlength{\tabcolsep}{3.5pt}\appRaggedCells
  \caption{Audit of UB-GOLD's Type III scenario.}
  \label{tab:adf-t3}
  \begin{tabular}{@{}p{2.5cm}p{7.7cm}p{6.15cm}@{}}
    \toprule
    Entry and evidence & Released behavior & Effect and handling \\
    \midrule
    T1. Validation is test\newline code & Validation data and loader are the combined test set & The same test-AUROC checkpoint and stopping channel is present. \\
    T2. Fixed anomaly prefix\newline code, data & The first $n$ second-dataset graphs are used, with $n$ equal to the ID test count. The first 60 PROTEINS graphs have class 1 and the first 150 IMDB-BINARY graphs class 0 & Every test set is half anomalous and shares its anomalies across trials. ENZYMES+PROTEINS and IMDB-MULTI+IMDB-BINARY use a single second-dataset class. \\
    T3. Overlapping trials\newline code & Trial $i$ uses seed $3407+i$ and a fresh random 90/10 ID split & ID test sets overlap and anomalies are shared. Trial spread understates uncertainty. Our intervals resample pairs. \\
    T4. Constant node features\newline code, data & Both TU datasets receive constant features with equal width & Tasks become structure-only and molecules lose atom labels. Size can separate pairs, for example AIDS at 15.7 mean nodes versus DHFR at 42.4. \\
    T7. SIGNET flags\newline code & All ten Type III script lines are commented out and still fail when enabled because two flags have incorrect spellings & Published values cannot come from these lines as released. We correct only \texttt{readout} and \texttt{encoder\_layers} spellings and mark the rows (A36). \\
    T8. Misaligned labels\newline code, data & The test loader shuffles and two-stage labels and scores use separate passes & With alignment, both InfoGraph variants exceed the published values on every pair. None is within five points. \\
    T11. Conflicting published cells\newline paper & On AIDS+DHFR, Table 3 gives InfoGraph-OCSVM 84.33 and IF 83.27. Table 8 gives 44.33 and 63.27 (arXiv 2406.15523v2) & Only the tens digits differ. We use Table 3 and retain the alternatives. The correct values are indeterminate. \\
    \bottomrule
  \end{tabular}
\end{table}

Two checks support the reproduction comparison.
T10 verifies all 60 transcribed values for 12 detectors on five TU pairs against Table 3 of arXiv 2406.15523v2.
In T12, an independent split implementation matches UB-GOLD indices under torch 2.4.0, 2.5.1 and 2.14.1.
We do not have the original 2.1.0 environment, so these checks support, but do not prove, identical indices in that environment.
T2--T4 limit conclusions about arbitrary graphs from the second dataset.
The checkpoint rules are still compared using the same data and training run.
We handle T7 and T8 with marked flag corrections and aligned labels, respectively.

\subsection{Selection in the recent methods}
\label{app:defects-recent}

This subsection specifies the recent-method selection rules used by \oracle{}-own and supports the code-audit counts in the main text.
Table~\ref{tab:adf-repos} gives the inspected commits.
The supplementary study record gives the repository paths and source lines.
MUSE, GLADMamba, ProtoGLAD, HC-GLAD and SUL/GUL have no license, so their code is not redistributed.

\begin{table}[!htb]
  \centering\footnotesize\setlength{\tabcolsep}{3.5pt}\appRaggedCells
  \caption{Selection information in recent released code.}
  \label{tab:adf-repos}
  \begin{tabular}{@{}p{1.85cm}p{9.35cm}p{5.15cm}@{}}
    \toprule
    Method & Selection in released code & Paper statement \\
    \midrule
    MUSE\newline 07da392 & Model selection and stopping use a disjoint validation set with 5\% of anomalies. Results average both normal-class choices & Labeled validation and validation-AUROC tuning stated (arXiv 2410.20366v1, Section 6.1, Appendix D.5). \\
    LGKDE\newline 1bba5ef & Only train and test loaders. Checkpoint and stopping use test $(\mathrm{AUROC}+\mathrm{AUPR})/2$. One configuration is read. Only MUTAG is released & Early stopping on a validation set that the code does not construct. \\
    GLADPro\newline 1f811b9 & Reports the epoch with the highest mean test AUROC over splits. Explanation AUROCs use their own test-mean maxima. No hyperparameter loop & Grid ranges without selection data or metric. No epoch rule (KDD'25, Appendix C.3). \\
    GLADMamba\newline 2b2462c & Reports the epoch with the highest mean test AUROC across folds. No hyperparameter loop & No rule stated (arXiv 2503.17903v2). \\
    SEGO\newline 45bdda7 & Reports the last scheduled test evaluation. Retains no best value or checkpoint. No search & No rule stated (repository paper PDF). \\
    GLASS\newline 3bbba39 & Test AUROC selects fusion weights, $\alpha$, checkpoint (patience 40), and the best of six scorers. Reference evaluation compares 24 rules from three scores and eight weights and loads only a test-selected checkpoint & v1 gives no rule. v2 specifies a final epoch only for a separate private-reference experiment without a released entry point and identifies the other path as legacy evaluation-driven selection. \\
    ProtoGLAD\newline f7d8ce9 & Searches $(h,\psi_1,\psi_2,\tau)$ by mean test AUROC across folds. No checkpoints & No own-hyperparameter rule stated (arXiv 2602.10708v1). \\
    HC-GLAD\newline 4238b20 & Records test AUROC, disables last-epoch reporting, and reports the best mean epoch. No hyperparameter loop & No rule stated (arXiv 2407.02057v2). \\
    SUL/GUL\newline 1a1f2e4 & Reports the epoch with the highest mean test AUROC over splits. No hyperparameter loop & 1000 epochs and grid search without data or metric. No checkpoint rule (WWW'26, Appendix C.2). \\
    \bottomrule
  \end{tabular}
\end{table}

\paragraph{Additional findings affecting interpretation.}
Four papers reuse baseline rows from other studies.
LGKDE uses the test-tuned UB-GOLD Table 8, GLASS uses UB-GOLD, GLADMamba uses GOOD-D and CVTGAD, and HC-GLAD arXiv v2 uses GOOD-D.
Of the 90 overlapping GLASS v2 baseline cells, 89 equal UB-GOLD values, with OCGTL on REDDIT-B at 88.03 rather than 89.92.
GLASS itself makes the least frequent class anomalous and tests on 20\% held-out normals plus all anomalies, so copied baselines target a different class on the five majority-anomaly datasets.
Its eigendecomposition starting vector is not seed-controlled, causing differences in 32 of 50 identical-configuration pairs and up to 8.33 points under \valnormal{}.
Its default scorer uses only normal reference graphs and is inductive, despite having ``test'' in its name.

SEGO creates and zeros its tree-encoder optimizer but never updates it, leaving the encoder at random initialization.
At test time, the tree term repeats the graph loss but standardizes it with tree-loss statistics.
The graph term uses other batch graphs as negatives, making its score depend on batch composition.
The anomaly loader omits \texttt{follow\_batch} and fails with a KeyError, and the code makes class 0 anomalous while the paper describes the minority or real anomalous class.
MUSE's \texttt{to\_dense\_adj} fails when a graph ends with isolated nodes, and the supplementary study record gives full locations for these issues.

Three further checks help explain the reported results.
On AIDS, GLADPro's own rule selects the first epoch at 55.25, compared with 33.52 at the final epoch.
In every fold, 99--100\% of evaluation points lie below 50, so the reported value describes an almost untrained model.
GLADMamba standardizes scores using current training-loss statistics, but scores held-out graphs in chunks of a different size.
This often makes \valnormal{} select early evaluations and triggers the A14 descriptive flag under default settings.
ProtoGLAD's tuned settings exceed defaults by 3.86 points on UB9, and its batch script runs both normal-class choices and writes them side by side without selecting an anomaly direction.

HC-GLAD stores \texttt{hyperedge\_index} as an ordinary attribute without \texttt{\_\_inc\_\_}, so PyG shifts hyperedge IDs by node count.
Graphs with more hyperedges than nodes then share hyperedge IDs with the next graph.
Such graphs comprise 85--100\% of MUTAG, BZR, COX2 and DHFR, 41\% of AIDS, 23\% of ENZYMES and 22\% of PROTEINS\_full.
Test order consequently changes scores, and we retain the released code.
A diagnostic with 20 test orders on fold 0 after 100 epochs changes AUROC by about 2.5 points, for example 48.2--50.8 on COX2, but the effect was not quantified on the main grid.

Seven of the nine code bases use test labels to select a reported epoch, checkpoint or hyperparameter.
MUSE uses labeled anomalies in a disjoint validation set, and SEGO reports the final evaluation.
Apart from MUSE, the papers omit the implemented rule, describe a different rule, or state grid search without identifying the selection data and metric.
We cannot determine whether published values came from these paths, so we report the rules diagnostically as \oracle{}-own alongside label-free rules.

\subsection{Other inspected code bases and papers}
\label{app:defects-other}

Five additional code bases from 2023--2026 complete the 14 recent implementations in the main-text audit, with reasons for not re-running them in Table~\ref{tab:ports-inclusion}.
In Table~\ref{tab:defects-other}, B means both final-epoch and best-test averages are reported, ? denotes an unresolved choice, and $c$ denotes a reconstruction threshold.
\begin{table}[!htb]
  \centering\footnotesize\setlength{\tabcolsep}{3pt}\appRaggedCells
  \caption{Selection rules in the additional code bases.}
  \label{tab:defects-other}
  \begin{tabular}{@{}p{2.1cm}p{1.55cm}ccc p{7.1cm}p{3.3cm}@{}}
    \toprule
    & & \multicolumn{3}{c}{Test labels select} & & \\
    \cmidrule(lr){3-5}
    Code base & Venue & Ckpt. & HP & Other & Released-code evidence & Paper statement \\
    \midrule
    FGS-GLAD\newline 3deaf13 & AAAI'26 & \audY & \audN & \audY$^c$ & Selects and reloads teacher and student by best test ROC-AUC. Test errors set the reconstruction threshold. No search & 200 epochs. Early stopping on validation loss \\
    CLINIC\newline 51005e2 & ICML'26 & B & ? & \audN & Test-only evaluation. Prints last and best averages. Loops over $k_2\in\{2,3\}$. No license & Five-run mean AUC. Candidate sets without data. $k_2\in\{3,5\}$ (PMLR 306) \\
    HimNet\newline 8c81fcd & ECML-PKDD'23 & \audN & \audN & \audN & Test AUROC evaluated at the final epoch only, including the Tox21 script & No rule stated (arXiv 2307.00755v1) \\
    GRAM\newline 4046124 & Neural Netw.'24 & \audN & \audN & \audN & Lowest-training-loss checkpoint. Test evaluation once after training & No rule stated (arXiv 2311.06153v2) \\
    FANFOLD\newline be2c09b & arXiv'24 & \audN & \audN & \audN & Teacher, flow and student selected by training losses. Test evaluated once & No rule stated (arXiv 2407.00383v1) \\
    \bottomrule
  \end{tabular}
\end{table}

FGS-GLAD raises the confirmed count of code bases using test-label selection from seven to eight.
CLINIC evaluates only test data, so any selection would use test labels, but which reported average its paper uses remains indeterminate.
HimNet, GRAM and FANFOLD show that the practice is not universal, and these five code bases contain no hyperparameter search apart from the CLINIC loop.
The main-text count applies only to the code versions we audited.

We also inspected four further sources.
GOODAT at a10dd97 has no license and releases 15 checkpoints without training scripts.
It loads one \texttt{ad\_AIDS.pt} before adapting on each fold's test set.
Its original training graphs and possible overlap across folds are unknown.
SDM and AGDiff release no code.
They describe grid search for optimal performance but do not identify validation data, so the data used for selection are unknown.
M-GLAD at c5f9ecb contains only a supplementary PDF, reporting GLADPro AUPR of 36.77 on BZR and 29.66 on COX2, compared with 77--89 for other methods.
UB-GOLD test anomaly proportions are 77.8--79.0\% and 77.7--78.5\%, respectively, suggesting a reversed anomaly class, but this remains an inference.
The DO2HSC code address returned 404 on 2026-10-13.

\subsection{Practices that support interpretation}
\label{app:defects-credit}

The audit also records practices that support accurate interpretation.
UB-GOLD explicitly calls test-tuned values upper bounds and explained that choice in public review.
The CVTGAD README states that it reports the best test epoch and explains why.
OCGTL stops on normal-only validation loss and motivates unbiased evaluation without true anomalies, while OCGIN uses defaults for a fixed 25 epochs.
MUSE discloses labeled validation disjoint from test.
GLASS v2 adds a fixed-final-epoch re-run without target-metric access, and DO2HSC reports final-epoch results.
These differences show that the audited methods do not all use the same selection practice.

\section{Complete results on UB9}
\label{app:extra}

This appendix gives the full selection and ranking results for Section~\ref{sec:f2}.
Appendix~\ref{app:followup} tests why the checkpoint gaps arise and whether they persist under other settings.
Unless stated otherwise, AUROC scores are expressed as percentages, and differences as percentage points.
We first average UB-GOLD's five folds within each dataset, then average datasets with equal weight.
Brackets give 95\% percentile intervals from 4000 bootstrap resamples of datasets, with detectors held fixed.
Scripts generate all tables from the released per-run files.

\subsection{Leaderboard under a common selection rule}
\label{app:extra-leaderboard}

In Table~\ref{tab:leaderboard}, \oracle{} uses each recent method's own test-label selection rule when one is available.
Table~\ref{tab:extra-leaderboard} compares these scores with those obtained by applying UB-GOLD's rule to every method.
The table includes 21 detectors, four UB-GOLD kernel baselines, and five statistical references.
Ranks in parentheses are computed over all 30 rows.
The own-rule column replays the released selection rule, using UB-GOLD's rule for MUSE and SEGO because they provide none.
All checkpoint gaps averaged across datasets and detectors use UB-GOLD's rule.
The \valnormal{} column also reports dataset-bootstrap intervals.
Kernels and references have no checkpoint selection, so their first four protocol columns coincide, with test-selected hyperparameters used for the kernels.

\begin{table}[!ht]
  \centering\footnotesize
  \setlength{\tabcolsep}{4pt}
  \caption{UB9 leaderboard with all 30 rows.}
  \label{tab:extra-leaderboard}
  \begin{tabular}{llcccccc}
\toprule
 & & \oracle{} & \oracle{} & \last{} & \valnormal{} & \valnormal{} & \hpfree{} \\
Detector & Group & own rule & UB-GOLD rule & & & 95\% CI & \\
\midrule
SIGNET & UB-GOLD & 74.7 (2) & 74.7 (2) & 70.7 (2) & 71.6 (1) & [65.4, 78.4] & 69.3 (1) \\
CVTGAD & UB-GOLD & 76.1 (1) & 76.1 (1) & 71.0 (1) & 71.1 (2) & [64.5, 78.3] & 66.5 (4) \\
OCGTL & UB-GOLD & 72.1 (7) & 72.1 (6) & 68.7 (4) & 68.8 (3) & [59.4, 79.6] & 60.1 (17) \\
GOOD-D & UB-GOLD & 72.3 (6) & 72.3 (5) & 67.7 (6) & 67.6 (4) & [60.1, 76.7] & 66.8 (2) \\
OCGIN & UB-GOLD & 70.2 (8) & 70.2 (7) & 64.5 (7) & 64.8 (6) & [55.7, 75.6] & 63.9 (10) \\
InfoGraph-OCSVM & UB-GOLD & 67.7 (12) & 67.7 (12) & 61.3 (16) & 64.7 (7) & [53.9, 76.1] & 64.1 (8) \\
InfoGraph-IF & UB-GOLD & 69.5 (9) & 69.5 (9) & 59.8 (18) & 64.1 (10) & [53.1, 75.7] & 63.4 (11) \\
GraphCL-IF & UB-GOLD & 67.1 (13) & 67.1 (13) & 63.1 (11) & 64.0 (11) & [55.8, 72.6] & 63.9 (9) \\
GLADC & UB-GOLD & 68.2 (11) & 68.2 (11) & 63.1 (10) & 63.4 (14) & [53.0, 74.5] & 64.2 (7) \\
GraphCL-OCSVM & UB-GOLD & 68.9 (10) & 68.9 (10) & 62.5 (13) & 62.9 (15) & [52.9, 74.6] & 62.7 (13) \\
GLocalKD & UB-GOLD & 66.4 (15) & 66.4 (14) & 62.5 (14) & 62.5 (16) & [54.6, 71.4] & 61.1 (15) \\
GraphDE & UB-GOLD & 54.4 (27) & 54.4 (28) & 51.0 (26) & 52.3 (26) & [45.9, 58.7] & 53.4 (24) \\
\midrule
GLASS & recent & 72.8 (4) & 70.1 (8) & 63.8 (8) & 64.9 (5) & [56.0, 73.9] & 64.5 (5) \\
SEGO & recent & 72.7 (5) & 72.7 (4) & 68.4 (5) & 64.5 (8) & [58.7, 71.1] & 66.7 (3) \\
HC-GLAD & recent & 66.5 (14) & 65.7 (15) & 63.1 (12) & 64.2 (9) & [54.0, 75.5] & 64.3 (6) \\
GLADMamba & recent & 73.2 (3) & 74.2 (3) & 69.8 (3) & 63.7 (12) & [54.7, 73.9] & 63.3 (12) \\
ProtoGLAD & recent & 63.4 (17) & 63.4 (18) & 63.4 (9) & 63.4 (13) & [56.4, 73.7] & 59.6 (18) \\
MUSE & recent & 65.6 (16) & 65.6 (16) & 57.9 (21) & 59.0 (19) & [53.2, 66.9] & 54.8 (23) \\
LGKDE & recent & 52.8 (28) & 56.8 (24) & 54.5 (25) & 54.1 (25) & [48.6, 59.5] & 51.5 (25) \\
SUL/GUL & recent & 58.3 (22) & 56.2 (25) & 48.3 (28) & 49.7 (27) & [43.9, 55.0] & 49.7 (27) \\
GLADPro & recent & 59.0 (20) & 64.9 (17) & 46.8 (29) & 46.3 (29) & [38.9, 52.9] & 48.9 (28) \\
\midrule
WL-IF & kernel & 58.6 (21) & 58.6 (21) & 58.6 (19) & 58.6 (20) & [49.0, 69.5] & 56.8 (21) \\
PK-IF & kernel & 54.5 (26) & 54.5 (27) & 54.5 (24) & 54.5 (24) & [51.2, 58.2] & 50.9 (26) \\
PK-OCSVM & kernel & 48.8 (29) & 48.8 (29) & 48.8 (27) & 48.8 (28) & [42.3, 54.5] & 47.2 (29) \\
WL-OCSVM & kernel & 43.2 (30) & 43.2 (30) & 43.2 (30) & 43.2 (30) & [31.0, 54.2] & 42.0 (30) \\
\midrule
Combined statistics & reference & 62.1 (18) & 62.1 (19) & 62.1 (15) & 62.1 (17) & [52.6, 73.6] & 62.1 (14) \\
Graphlet count + LW & reference & 60.4 (19) & 60.4 (20) & 60.4 (17) & 60.4 (18) & [49.6, 72.4] & 60.4 (16) \\
Graphlet count + IF & reference & 58.2 (23) & 58.2 (22) & 58.2 (20) & 58.2 (21) & [48.1, 69.5] & 58.2 (19) \\
Size only & reference & 57.3 (24) & 57.3 (23) & 57.3 (22) & 57.3 (22) & [47.6, 69.7] & 57.3 (20) \\
WL histogram + IF & reference & 55.8 (25) & 55.8 (26) & 55.8 (23) & 55.8 (23) & [47.6, 64.5] & 55.8 (22) \\
\bottomrule
\end{tabular}

\end{table}

The leaders are the same as in the main text, but changing the checkpoint rule changes some recent methods' scores.
GLADPro scores 59.0 under its own rule and 64.9 under UB-GOLD's.
Wide \valnormal{} intervals reflect dataset variation, as illustrated by SIGNET's 57.4 on IMDB-B and 91.9 on AIDS.
A small rank difference does not by itself show that one detector is better.
Appendix~\ref{app:extra-kendall} compares the full rankings using Kendall's $\tau_b$ and its intervals.

\subsection{Dataset variation and selector failures}
\label{app:extra-perdataset}

Table~\ref{tab:extra-ub9-valnormal} reports five-fold mean AUROC on each UB9 dataset under \valnormal{}, with the last column matching the UB9 mean in Table~\ref{tab:extra-leaderboard}.
A section sign marks a constant kernel matrix, whose AUROC is 50 by construction.

\begin{table}[!ht]
  \centering\footnotesize
  \caption{Per-dataset UB9 results under \valnormal{}.}
  \label{tab:extra-ub9-valnormal}
  \begin{tabular}{lcccccccccc}
\toprule
Detector & AIDS & BZR & COX2 & DHFR & ENZ. & NCI1 & PROT. & IMDB-B & RDT-B & Mean \\
\midrule
SIGNET & 91.9 & 77.6 & 70.3 & 72.3 & 60.4 & 70.1 & 64.1 & 57.4 & 80.2 & 71.6 \\
CVTGAD & 91.0 & 72.9 & 55.6 & 60.0 & 66.4 & 74.0 & 64.3 & 69.9 & 85.3 & 71.1 \\
OCGTL & 99.3 & 54.1 & 54.3 & 51.5 & 63.4 & 76.9 & 69.3 & 62.2 & 88.6 & 68.8 \\
GOOD-D & 91.7 & 66.3 & 54.4 & 59.6 & 59.2 & 57.8 & 66.1 & 63.9 & 89.6 & 67.6 \\
OCGIN & 92.7 & 64.8 & 50.1 & 52.5 & 62.7 & 49.2 & 63.7 & 56.4 & 91.2 & 64.8 \\
InfoGraph-OCSVM & 99.6 & 62.0 & 56.7 & 60.0 & 60.0 & 33.8 & 74.2 & 56.9 & 79.3 & 64.7 \\
InfoGraph-IF & 99.6 & 64.7 & 55.4 & 59.4 & 57.2 & 32.5 & 73.8 & 55.6 & 79.2 & 64.1 \\
GraphCL-IF & 85.7 & 60.1 & 57.2 & 51.3 & 59.9 & 44.6 & 72.7 & 61.0 & 83.7 & 64.0 \\
GLADC & 97.3 & 56.4 & 57.3 & 51.3 & 56.2 & 37.0 & 70.4 & 64.9 & 79.9 & 63.4 \\
GraphCL-OCSVM & 98.6 & 62.0 & 55.7 & 49.9 & 57.1 & 39.3 & 72.9 & 51.0 & 79.5 & 62.9 \\
GLocalKD & 89.1 & 64.3 & 43.6 & 56.0 & 54.9 & 61.5 & 64.9 & 50.3 & 78.2 & 62.5 \\
GraphDE & 65.8 & 39.3 & 52.0 & 56.3 & 63.0 & 45.8 & 37.6 & 47.1 & 63.5 & 52.3 \\
\midrule
GLASS & 89.0 & 47.4 & 50.1 & 54.2 & 53.0 & 63.7 & 71.9 & 71.1 & 83.3 & 64.9 \\
SEGO & 81.7 & 61.8 & 51.5 & 57.6 & 57.7 & 70.7 & 62.0 & 58.8 & 78.7 & 64.5 \\
HC-GLAD & 98.5 & 68.2 & 56.5 & 55.5 & 62.1 & 37.7 & 73.5 & 48.1 & 78.2 & 64.2 \\
GLADMamba & 93.7 & 56.6 & 54.4 & 61.1 & 54.5 & 43.9 & 74.9 & 53.4 & 81.1 & 63.7 \\
ProtoGLAD & 98.3 & 57.6 & 50.9 & 57.4 & 61.4 & 69.6 & 56.9 & 53.9 & 64.9 & 63.4 \\
MUSE & 85.8 & 50.6 & 52.8 & 49.7 & 58.6 & 56.6 & 55.5 & 54.8 & 66.8 & 59.0 \\
LGKDE & 59.1 & 45.7 & 43.6 & 41.8 & 55.8 & 62.4 & 57.0 & 54.3 & 67.3 & 54.1 \\
SUL/GUL & 53.7 & 32.4 & 41.7 & 46.2 & 51.2 & 50.2 & 50.2 & 63.6 & 57.8 & 49.7 \\
GLADPro & 25.2 & 42.0 & 55.4 & 42.5 & 57.9 & 59.0 & 33.2 & 50.3 & 51.0 & 46.3 \\
\midrule
WL-IF & 91.8 & 37.6 & 55.2 & 41.2 & 61.9 & 50.1 & 60.2 & 56.4 & 73.6 & 58.6 \\
PK-IF & 54.2 & 49.0 & 58.4 & 47.1 & 62.2 & 60.2 & 59.7 & 50.0$^\S$ & 50.0$^\S$ & 54.5 \\
PK-OCSVM & 35.4 & 53.7 & 56.2 & 47.7 & 60.5 & 55.8 & 30.0 & 50.0$^\S$ & 50.0$^\S$ & 48.8 \\
WL-OCSVM & 8.4 & 51.7 & 48.0 & 34.5 & 58.8 & 67.2 & 19.5 & 47.5 & 53.3 & 43.2 \\
\midrule
Combined statistics & 96.5 & 54.4 & 58.6 & 43.3 & 51.9 & 50.5 & 56.3 & 63.1 & 84.3 & 62.1 \\
Graphlet count + LW & 93.5 & 44.8 & 57.5 & 36.3 & 54.3 & 50.2 & 55.0 & 66.2 & 85.7 & 60.4 \\
Graphlet count + IF & 91.4 & 41.2 & 51.7 & 37.0 & 46.3 & 52.2 & 62.8 & 62.8 & 78.4 & 58.2 \\
Size only & 98.7 & 42.5 & 50.9 & 41.9 & 43.7 & 53.7 & 57.8 & 52.0 & 74.1 & 57.3 \\
WL histogram + IF & 78.5 & 37.4 & 54.6 & 41.4 & 61.2 & 49.5 & 56.7 & 49.6 & 73.5 & 55.8 \\
\bottomrule
\end{tabular}

\end{table}

Every detector varies substantially across datasets.
Most score well on AIDS, where size alone reaches 98.7, while several fall below 50 on NCI1, BZR, or PROTEINS\_full.
Simple graph statistics can therefore help distinguish normal from anomalous graphs in these benchmarks, consistent with Section~\ref{sec:f3}.
Results for other protocols and the secondary datasets DD, COLLAB, Tox21 p53, Tox21 MMP, and MUTAG are available in the supplementary study record and released files.
MUTAG also illustrates an important failure of normal-validation selection.
SEGO and GLADMamba score 14.5 and 23.9 under \valnormal{}, compared with 76.8 and 66.8 under \last{}.
Criterion G1 therefore requires \oracle{} to outperform both label-free rules, so a weak comparator alone cannot satisfy the criterion.

\subsection{Checkpoint leakage by detector}
\label{app:extra-inflation}

\begin{figure}[!ht]
  \centering
  \includegraphics[width=\textwidth]{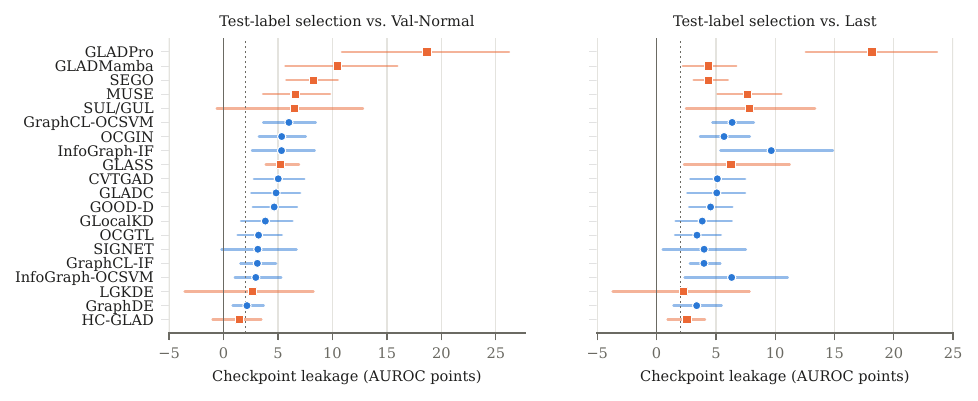}
  \caption{Checkpoint leakage against both label-free rules.}
  \label{fig:inflation-both}
\end{figure}

Figure~\ref{fig:inflation-both} and Table~\ref{tab:extra-inflation} give the complete detector-level checkpoint comparisons for Section~\ref{sec:f2}.
The left comparison is \oracle{} under UB-GOLD's rule minus \valnormal{}, and the right is \oracle{} minus \last{}, both with dataset-bootstrap intervals.
Circles mark UB-GOLD detectors, squares mark recent methods, and the dotted line marks the 2-point threshold of G1.
The table's $\geq 2$ columns count datasets in detector rows and detectors in summary rows.
The final column instead uses each recent method's own test-label selection rule.

\begin{table}[!ht]
  \centering\footnotesize
  \setlength{\tabcolsep}{4pt}
  \caption{Checkpoint leakage by detector on UB9.}
  \label{tab:extra-inflation}
  \begin{tabular}{lccccc}
\toprule
 & \multicolumn{2}{c}{\oracle{} $-$ \valnormal{}} & \multicolumn{2}{c}{\oracle{} $-$ \last{}} & Own rule \\
\cmidrule(lr){2-3}\cmidrule(lr){4-5}
Detector & Mean [95\% CI] & $\geq 2$ & Mean [95\% CI] & $\geq 2$ & $-$ \valnormal{} \\
\midrule
SIGNET & +3.15 [$-$0.19, 6.70] & 5/9 & +4.01 [0.52, 7.52] & 4/9 & -- \\
CVTGAD & +5.02 [2.79, 7.41] & 7/9 & +5.12 [2.85, 7.46] & 7/9 & -- \\
OCGTL & +3.21 [1.29, 5.34] & 4/9 & +3.41 [1.56, 5.41] & 5/9 & -- \\
GOOD-D & +4.65 [2.69, 6.74] & 7/9 & +4.55 [2.74, 6.38] & 6/9 & -- \\
OCGIN & +5.34 [3.25, 7.54] & 7/9 & +5.69 [3.66, 7.86] & 8/9 & -- \\
InfoGraph-OCSVM & +2.96 [1.03, 5.29] & 4/9 & +6.33 [2.37, 11.05] & 4/9 & -- \\
InfoGraph-IF & +5.32 [2.62, 8.37] & 6/9 & +9.67 [5.39, 14.87] & 9/9 & -- \\
GraphCL-IF & +3.10 [1.55, 4.82] & 5/9 & +3.99 [2.81, 5.38] & 8/9 & -- \\
GLADC & +4.82 [2.55, 7.03] & 6/9 & +5.07 [2.59, 7.46] & 6/9 & -- \\
GraphCL-OCSVM & +6.00 [3.65, 8.46] & 8/9 & +6.37 [4.71, 8.20] & 9/9 & -- \\
GLocalKD & +3.84 [1.61, 6.32] & 5/9 & +3.85 [1.61, 6.33] & 5/9 & -- \\
GraphDE & +2.15 [0.84, 3.68] & 3/9 & +3.38 [1.43, 5.49] & 4/9 & -- \\
\midrule
GLASS & +5.23 [3.88, 6.93] & 8/9 & +6.26 [2.31, 11.22] & 5/9 & +7.91 \\
SEGO & +8.25 [5.76, 10.49] & 8/9 & +4.38 [3.12, 6.01] & 9/9 & +8.25 \\
HC-GLAD & +1.44 [$-$1.01, 3.47] & 6/9 & +2.56 [0.93, 4.08] & 6/9 & +2.28 \\
GLADMamba & +10.44 [5.68, 15.95] & 7/9 & +4.38 [2.21, 6.71] & 6/9 & +9.45 \\
ProtoGLAD & \multicolumn{5}{c}{no checkpoints, one model per run, excluded from the pooled rows} \\
MUSE & +6.57 [3.63, 9.77] & 6/9 & +7.68 [5.13, 10.52] & 9/9 & +6.57 \\
LGKDE & +2.65 [$-$3.57, 8.26] & 6/9 & +2.29 [$-$3.70, 7.84] & 5/9 & $-$1.27 \\
SUL/GUL & +6.49 [$-$0.62, 12.80] & 7/9 & +7.85 [2.48, 13.36] & 7/9 & +8.59 \\
GLADPro & +18.68 [10.87, 26.20] & 9/9 & +18.16 [12.58, 23.64] & 9/9 & +12.75 \\
\midrule
UB-GOLD detectors (pooled) & +4.13 [3.32, 4.79] & 12/12 & +5.12 [3.97, 6.24] & 12/12 & -- \\
Recent methods (pooled) & +7.47 [6.15, 8.83] & 7/8 & +6.69 [5.75, 7.61] & 8/8 & -- \\
All checkpointed (pooled) & +5.47 [4.83, 6.02] & 19/20 & +5.75 [4.75, 6.67] & 20/20 & -- \\
\bottomrule
\end{tabular}

\end{table}

All 12 UB-GOLD detectors gain at least 2 points on average over both label-free rules, including the eight formal-test detectors.
Individual intervals include zero for SIGNET, HC-GLAD, LGKDE, and SUL/GUL against \valnormal{}, and for LGKDE against \last{}.
Criterion G1 concerns the mean gain and the proportion of detectors reaching the threshold, without requiring every individual interval to exclude zero.
With its own rule, GLASS's gain against \valnormal{} rises from 5.23 to 7.91, while GLADPro's falls from 18.68 to 12.75.
LGKDE's own rule selects checkpoints 1.27 points below the label-free choice.
ProtoGLAD has no checkpoints and is excluded from mean checkpoint-leakage estimates.

\subsection{Hyperparameter gaps by detector}
\label{app:extra-hpleak}

\begin{figure}[!ht]
  \centering
  \includegraphics{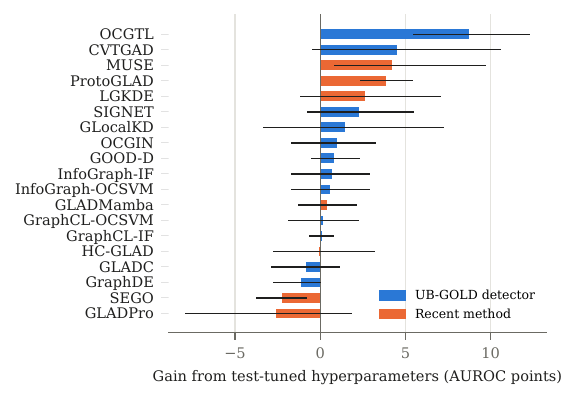}
  \caption{Hyperparameter gap by detector.}
  \label{fig:hpleak}
\end{figure}

Figure~\ref{fig:hpleak} examines how the mean hyperparameter gap varies across detectors.
Each point subtracts \valnormal{} AUROC with repository defaults from \valnormal{} AUROC with UB-GOLD's or the method's released per-dataset hyperparameters.
Intervals are computed by resampling datasets.
We train a separate run for each hyperparameter setting and use the same checkpoint rule in both.
GLASS and SUL/GUL each release one configuration and are omitted.

The mean gap over the 12 UB-GOLD detectors is 1.53 points, below the 2-point threshold of G2(b), with gains concentrated in a few detectors.
Adding checkpoint leakage, \oracle{} exceeds \hpfree{} by 5.66 points [4.27, 6.86] for the 12 UB-GOLD detectors, the sum of the 4.13-point checkpoint gap and the 1.53-point hyperparameter gap.
Table~\ref{tab:leaderboard} gives each detector's \oracle{} and \hpfree{} scores.
OCGTL gains 8.76 points, whereas SEGO, GLADPro, and GraphDE lose performance, and GLADC and HC-GLAD also score slightly higher with defaults.
SEGO's and GraphDE's intervals lie below zero, while GLADPro's interval $[-7.89,1.82]$ includes zero.
Test-set tuning does not benefit every detector, and this comparison combines the effects of tuning with those of access to test labels.
The mean gap is no longer positive in Type III (Appendix~\ref{app:t3}).

\subsection{Agreement between leaderboards}
\label{app:extra-kendall}

Table~\ref{tab:extra-kendall} extends Section~\ref{sec:c2} with Kendall's $\tau_b$~\citep{kendall1938} between protocol rankings.
Results cover the 12 UB-GOLD detectors, nine recent methods, all 21 detectors, and all 30 rows, with both fold- and dataset-bootstrap intervals.
The last column indicates whether the top-three set changes.
\oracle{} uses each method's own rule, \oracle{}$_\text{UB-GOLD}$ uses the common UB-GOLD rule, and \oracle{}$_\text{defaults}$ applies that common rule to default-hyperparameter runs.

\begin{table}[!ht]
  \centering\footnotesize
  \setlength{\tabcolsep}{4pt}
  \caption{Kendall's $\tau_b$ between UB9 rankings.}
  \label{tab:extra-kendall}
  \begin{tabular}{llcccc}
\toprule
Rows ranked & Protocols compared & $\tau_b$ & Fold-bootstrap CI & Dataset-bootstrap CI & Top 3 changed \\
\midrule
UB-GOLD (12) & \oracle{} vs \valnormal{} & 0.76 & [0.61, 0.85] & [0.36, 0.85] & yes \\
 & \oracle{} vs \last{} & 0.70 & [0.58, 0.79] & [0.42, 0.85] & yes \\
 & \valnormal{} vs \last{} & 0.70 & [0.58, 0.85] & [0.42, 0.97] & no \\
 & \oracle{}$_\text{UB-GOLD}$ vs \valnormal{} & 0.76 & [0.61, 0.85] & [0.36, 0.85] & yes \\
 & \oracle{} vs \hpfree{} & 0.45 & [0.33, 0.58] & [0.27, 0.67] & no \\
 & \valnormal{} vs \hpfree{} & 0.52 & [0.39, 0.70] & [0.39, 0.76] & yes \\
 & \oracle{}$_\text{defaults}$ vs \hpfree{} & 0.42 & [0.33, 0.58] & [0.33, 0.85] & yes \\
\midrule
Recent (9) & \oracle{} vs \valnormal{} & 0.61 & [0.39, 0.72] & [0.33, 0.78] & yes \\
 & \oracle{} vs \last{} & 0.67 & [0.56, 0.83] & [0.44, 0.78] & no \\
 & \valnormal{} vs \last{} & 0.72 & [0.56, 0.89] & [0.56, 0.94] & yes \\
 & \oracle{}$_\text{UB-GOLD}$ vs \valnormal{} & 0.56 & [0.22, 0.67] & [0.17, 0.67] & yes \\
 & \oracle{} vs \hpfree{} & 0.56 & [0.50, 0.78] & [0.44, 0.83] & yes \\
 & \valnormal{} vs \hpfree{} & 0.94 & [0.61, 1.00] & [0.61, 0.94] & no \\
 & \oracle{}$_\text{defaults}$ vs \hpfree{} & 0.67 & [0.50, 0.78] & [0.33, 0.78] & yes \\
\midrule
All detectors (21) & \oracle{} vs \valnormal{} & 0.65 & [0.52, 0.72] & [0.49, 0.73] & yes \\
 & \oracle{} vs \last{} & 0.70 & [0.62, 0.75] & [0.51, 0.80] & no \\
 & \valnormal{} vs \last{} & 0.70 & [0.60, 0.82] & [0.60, 0.88] & yes \\
 & \oracle{}$_\text{UB-GOLD}$ vs \valnormal{} & 0.65 & [0.52, 0.71] & [0.46, 0.71] & yes \\
 & \oracle{} vs \hpfree{} & 0.58 & [0.50, 0.68] & [0.45, 0.69] & yes \\
 & \valnormal{} vs \hpfree{} & 0.70 & [0.57, 0.75] & [0.57, 0.81] & yes \\
 & \oracle{}$_\text{defaults}$ vs \hpfree{} & 0.56 & [0.49, 0.65] & [0.43, 0.70] & yes \\
\midrule
All rows (30) & \oracle{} vs \valnormal{} & 0.77 & [0.70, 0.81] & [0.59, 0.80] & yes \\
 & \oracle{} vs \last{} & 0.77 & [0.71, 0.81] & [0.60, 0.81] & no \\
 & \valnormal{} vs \last{} & 0.83 & [0.77, 0.89] & [0.75, 0.92] & yes \\
 & \oracle{}$_\text{UB-GOLD}$ vs \valnormal{} & 0.77 & [0.69, 0.79] & [0.55, 0.79] & yes \\
 & \oracle{} vs \hpfree{} & 0.69 & [0.65, 0.76] & [0.53, 0.75] & yes \\
 & \valnormal{} vs \hpfree{} & 0.79 & [0.73, 0.83] & [0.67, 0.85] & yes \\
 & \oracle{}$_\text{defaults}$ vs \hpfree{} & 0.70 & [0.66, 0.74] & [0.57, 0.77] & yes \\
\bottomrule
\end{tabular}

\end{table}

The study plan specifies a ranking comparison between \oracle{} and \valnormal{} for the 12 UB-GOLD detectors.
It gives $\tau_b=0.76$, with a dataset-bootstrap interval extending down to 0.36.
The point estimate meets criterion G2(a), but the top three change in only 45\% of dataset-bootstrap samples.
The two label-free rules themselves give $\tau_b=0.70$.
Changing hyperparameters produces more reordering, with $\tau_b=0.45$ between \oracle{} and \hpfree{} and 0.52 between \valnormal{} and \hpfree{}.
The protocol changes the rankings, but the direction of an individual detector's rank change remains uncertain.
Appendix~\ref{app:followup-seeds} compares these changes with those caused by changing the training seed.

\subsection{Secondary metrics and tests between detectors}
\label{app:extra-secondary}

We next examine whether the leakage and leading methods depend on the use of AUROC.
The study plan specifies AUPRC and FPR95 as secondary metrics, together with Friedman tests and Holm correction for each protocol.
FPR95 is the proportion of normal test graphs flagged when 95\% of anomalies are detected, so lower values are better.
We average folds and datasets as for AUROC, and also average detectors for overall gaps.
\oracle{} uses UB-GOLD's rule throughout.

For every UB-GOLD detector, test-label checkpoint selection gives higher AUPRC and lower FPR95 than \valnormal{}.
SIGNET's FPR95 difference is only 0.08, at 71.16 versus 71.24.
Test-label selection also improves the reported AUPRC and FPR95, but the leading detector depends on the metric.
Under \valnormal{}, SIGNET, CVTGAD, and OCGTL are the top three of the 21 detectors in all three metrics, while OCGTL leads AUPRC and FPR95 under both label-free rules.
The main-text lead of SIGNET and CVTGAD is specific to AUROC, the primary metric, and full AUPRC and FPR95 results for all 30 rows are in the released files.

Table~\ref{tab:extra-friedman} reports Friedman tests on the nine UB9 datasets, their Iman--Davenport form, and the Nemenyi critical difference of average ranks~\citep{demsar2006}.
It also counts Wilcoxon signed-rank pairs significant at 0.05 after Holm correction.
Here \oracle{} uses UB-GOLD's rule for every method, unlike the main leaderboard.
For the 12 UB-GOLD detectors, the overall-test $p$-value rises from $p=3.0\times10^{-4}$ under \oracle{} to 0.027 under \valnormal{} and 0.068 under \hpfree{}.
No pair is significant after Holm correction under any protocol, and the smallest unadjusted pairwise $p$-value is 0.0039.
These nine-dataset tests do not distinguish individual detectors or establish equivalence, so the main text emphasizes effect sizes, bootstrap intervals, and $\tau_b$.

\begin{table}[!ht]
  \centering\footnotesize
  \setlength{\tabcolsep}{4pt}
  \caption{Friedman tests on UB9 by protocol.}
  \label{tab:extra-friedman}
\begin{tabular}{llccccc}
\toprule
Detectors & Protocol & $\chi^2_F$ & $p$ & Iman--Davenport $p$ & Nemenyi CD & Holm-significant pairs \\
\midrule
12 UB-GOLD & \oracle{}$_\text{UB-GOLD}$ & 34.54 & $3.0\times10^{-4}$ & $4.2\times10^{-5}$ & 5.55 & 0 of 66 \\
12 UB-GOLD & \last{} & 26.57 & 0.0053 & 0.0024 & 5.55 & 0 of 66 \\
12 UB-GOLD & \valnormal{} & 21.65 & 0.027 & 0.019 & 5.55 & 0 of 66 \\
12 UB-GOLD & \hpfree{} & 18.64 & 0.068 & 0.056 & 5.55 & 0 of 66 \\
\midrule
21 & \oracle{}$_\text{UB-GOLD}$ & 69.09 & $2.6\times10^{-7}$ & $2.1\times10^{-9}$ & 10.44 & 0 of 210 \\
21 & \last{} & 75.23 & $2.5\times10^{-8}$ & $4.6\times10^{-11}$ & 10.44 & 0 of 210 \\
21 & \valnormal{} & 58.88 & $1.1\times10^{-5}$ & $6.0\times10^{-7}$ & 10.44 & 0 of 210 \\
21 & \hpfree{} & 52.62 & $9.2\times10^{-5}$ & $1.3\times10^{-5}$ & 10.44 & 0 of 210 \\
\bottomrule
\end{tabular}
\end{table}

\subsection{Flagged runs and sensitivity}
\label{app:extra-flags}

We record non-finite scores, many identical scores, and training failures to check their effect on the comparisons in Section~\ref{sec:f2}.
The first two flags were defined before computing the affected averages.
A selected evaluation point is flagged as \emph{non-finite} if its test scores are non-finite.
This flag applies only to the final point used by \last{}, because amendment A15 prevents \oracle{} and \valnormal{} from selecting such points.
\last{} scores this output at chance because undefined scores cannot rank test graphs.
A selected point with fewer than $\max(10,0.05\,n_\text{test})$ distinct finite scores is flagged as \emph{tie-collapsed}.
These runs retain their original metrics, and a sensitivity analysis excludes them.
Training is recorded as diverged if two attempts in the method's own code fail in the same way with the same seed, without patching or imputation.
Only GLADC on COLLAB in folds 1, 3, and 4 diverges this way under both hyperparameter settings on CPU and GPU.

Table~\ref{tab:extra-flags} counts flagged runs across datasets, folds, and hyperparameter settings.
Table~\ref{tab:extra-sensitivity} recomputes the mean checkpoint and hyperparameter gaps and detector counts after excluding flagged runs.
Detecting identical scores requires per-graph outputs, which are unavailable for kernels, references, and some pilot-detector runs.
No flag therefore does not necessarily mean that no scores are tied.

\begin{table}[!ht]
  \centering\footnotesize
  \caption{Detectors with flagged or diverged runs.}
  \label{tab:extra-flags}
  \begin{tabular}{lcccc}
\toprule
 & & Non-finite at & Tie- & Diverged \\
Detector & Runs & selected point & collapsed & (excluded) \\
\midrule
OCGTL & 135 & 0 & 5 & 0 \\
GLADC & 119 & 0 & 0 & 6 \\
GraphCL-OCSVM & 130 & 0 & 3 & 0 \\
GraphDE & 130 & 0 & 13 & 0 \\
HC-GLAD & 100 & 2 & 2 & 0 \\
ProtoGLAD & 100 & 0 & 3 & 0 \\
SUL/GUL & 100 & 10 & 30 & 0 \\
GLADPro & 100 & 0 & 15 & 0 \\
\bottomrule
\end{tabular}

\end{table}

\begin{table}[!ht]
  \centering\footnotesize
  \setlength{\tabcolsep}{4pt}
  \caption{Specified quantities after excluding flagged runs.}
  \label{tab:extra-sensitivity}
  \begin{tabular}{lcccc}
\toprule
Quantity (12 UB-GOLD detectors) & All runs & Drop non-finite & Drop tie-collapsed & Drop either \\
 & & (12 runs) & (71 runs) & (71 runs) \\
\midrule
Leakage, \oracle{} $-$ \valnormal{} & +4.13 [3.32, 4.79] & +4.13 [3.32, 4.79] & +4.13 [3.32, 4.79] & +4.13 [3.32, 4.79] \\
\quad detectors with $\geq 2$ points & 12/12 & 12/12 & 12/12 & 12/12 \\
Leakage, \oracle{} $-$ \last{} & +5.12 [3.97, 6.24] & +5.12 [3.97, 6.24] & +5.11 [3.97, 6.23] & +5.11 [3.97, 6.23] \\
\quad detectors with $\geq 2$ points & 12/12 & 12/12 & 12/12 & 12/12 \\
Hyperparameter gap & +1.53 [0.64, 2.42] & +1.53 [0.64, 2.42] & +1.54 [0.65, 2.43] & +1.54 [0.65, 2.43] \\
\bottomrule
\end{tabular}

\end{table}

Excluding the 71 tie-collapsed runs or 12 non-finite runs changes the mean leakage and hyperparameter gap by at most 0.01 points.
All 12 UB-GOLD detectors still gain at least 2 points over both rules, and most flags concern recent methods excluded from these quantities.
A separate study amendment, A14, defines a descriptive flag for \valnormal{} degeneracy.
A detector is flagged if at least 80\% of its runs select within the first 10\% of training and its mean \valnormal{} AUROC is at least 5 points below \last{}.
Only GLADMamba with default hyperparameters meets this condition, and with tuned hyperparameters it still selects this early in 73\% of UB9 runs.
Its drop from third under \last{} to twelfth under \valnormal{} therefore mainly reflects checkpoint-selector behavior.

\subsection{Run-to-run variation}
\label{app:extra-noise}

In Appendix~\ref{app:followup-seeds}, the mean checkpoint gaps change little across extra seeds, even though individual runs vary substantially.
For a detector on one dataset and fold, the across-seed standard deviation averages 0.8 to 4.1 points, while mean checkpoint leakage changes by only a few tenths.
Even with fixed seeds, GLASS draws a fresh starting vector in SciPy's sparse eigensolver, and ProtoGLAD differs between CPU and GPU (Appendix~\ref{app:ports-devices}).
The main comparisons therefore rely on averages across folds, datasets, and detectors rather than individual runs.

\section{Sources and stability of checkpoint leakage}
\label{app:followup}
\label{app:posthoc}

This appendix tests why the checkpoint gaps in Section~\ref{sec:f2} arise and whether they persist under other settings, extending Section~\ref{sec:f2-mech}.
Unless stated otherwise, the analyses cover the 12 UB-GOLD detectors, are descriptive, and leave the original decision criteria unchanged.
The Type III replication in Appendix~\ref{app:t3} is the only follow-up with a decision criterion fixed before its runs (R1).

\paragraph{Timing of analysis specifications.}
Identifiers such as A32 and A36 refer to study-record amendments in Appendix~\ref{app:prereg}, and C, C2, E, F, and G identify analyses in the follow-up plan.
A32 specified five post hoc analyses on 2026-10-14 at 20:55 EDT, after the frozen statistical code had run and the first paper draft existed.
The record states that none had yet been computed and was committed to the private repository within a minute.
The five analyses use stored outputs without retraining to compare checkpoint rules, permute labels, change dataset coverage, match graph sizes, and check uncertain reproductions.
Size matching is reported in Appendix~\ref{app:posthoc-size}.
A36 and A40--A47 specified parts C, C2, E, F, and G of the second-version follow-up before the corresponding results, except that A46 confirmed implementation details for part E afterward.
Table~\ref{tab:prereg-v2} records when these plans were written relative to their results.
They lack external registration predating their results, so we call them specified in advance rather than pre-registered.

\paragraph{Statistical conventions.}
AUROC scores are expressed as percentages, and differences as percentage points.
We average five folds within each dataset, then datasets equally, and finally detectors.
Brackets give 95\% percentile intervals from 4000 bootstrap resamples of datasets with detectors fixed.
Study amendment A48 restricts extra seeds to descriptive analyses and excludes them from the formal tests.
The UB9 main-comparison runs of the four pilot detectors, SIGNET, OCGTL, OCGIN, and GLocalKD, stored checkpoint-level results but no per-graph scores.
They can enter analyses of recorded checkpoint results but not the main-comparison post hoc label-permutation analysis.

Table~\ref{tab:robust-matrix} summarizes what each check changes and holds fixed, and Table~\ref{tab:robust-full} collects the numerical results.
We first examine whether the gaps persist across settings, then analyze alternative selectors and selection optimism, and finally address sensitivity to the protocol and study coverage.

\begin{table}[!ht]
  \centering\footnotesize
  \caption{Checks of checkpoint leakage.}
  \label{tab:robust-matrix}
  \appRaggedCells
  \setlength{\tabcolsep}{4pt}
  \begin{tabular}{@{}p{2.9cm}p{3.1cm}p{2.9cm}p{5.0cm}p{2.1cm}@{}}
    \toprule
    Check & What changes & What is held fixed & What it shows & Status \\
    \midrule
    Other label-free selectors (\ref{app:posthoc-selectors}) & checkpoint rule & runs, hyperparameters & no tested rule closes the gap & descriptive (A32) \\
    Label-permutation check (\ref{app:posthoc-null}) & test labels & runs, trajectories & noise can yield more than half the observed gap, as an upper bound & descriptive (A32, A36) \\
    Extra seeds (\ref{app:followup-seeds}) & training seed & folds, protocol & mean leakage stays stable while individual runs vary & descriptive (A36, A40) \\
    Seed-transfer check (\ref{app:followup-seeds}) & seed used to select the epoch & detector, dataset, fold & about half the gap transfers & descriptive (A36) \\
    Disjoint labeled split (\ref{app:followup-split}) & selection graphs & runs, stored scores & disjoint labeled selection recovers part of the gap, leaving optimism & descriptive (A36, A45) \\
    Isolated test evaluations (\ref{app:followup-iso}) & evaluation random draws & training protocol, seed & mean changes are below one seed SD and the gap persists & descriptive (A42) \\
    All normal graphs (\ref{app:followup-allnormals}) & hold-out fraction & detector, seed & \last{} is essentially unchanged and the gap persists & descriptive (A42) \\
    Minority class (\ref{app:extra-orientation}) & anomaly class & detectors, hyperparameters & the gap remains positive & descriptive (A33) \\
    Earlier commit (\ref{app:extra-version}) & UB-GOLD version & data, protocol & changes are small & descriptive (A17) \\
    Dataset robustness (\ref{app:posthoc-datasets}) & datasets included in the mean & runs & the gap remains positive after any one dataset is removed & descriptive (A32) \\
    Type III (Appendix~\ref{app:t3}) & scenario & detectors, protocol & leakage is larger and the hyperparameter gap is no longer positive & decision criterion (R1) \\
    \bottomrule
  \end{tabular}
\end{table}

\subsection{Checkpoint leakage across settings}
\label{app:followup-robust}

Table~\ref{tab:robust-full} adds results for default hyperparameters, an earlier code version, extra seeds, and subsets to Table~\ref{tab:robust}.
Each row gives the detector count, dataset or pair count, mean \oracle{} gaps against \last{} and \valnormal{}, and the number of detectors gaining at least 2 points on average.
The four-detector, three-dataset subset averages 12 detector--dataset combinations.
We also report the mean after excluding the only combination run on GPU.
Each row retains its original interval definition.
Type III intervals resample only five dataset pairs and therefore have limited resolution.
Anomaly-class checks marked with a dagger resample 15 dataset--fold combinations.
Their intervals are optimistic because folds share training graphs.
Entries without brackets have no interval, a dash marks an inapplicable count, and n/a marks \valnormal{} when no graphs are held out.

\begin{table}[!ht]
  \centering\footnotesize
  \caption{Checkpoint leakage across all settings.}
  \label{tab:robust-full}
  \begin{tabular}{lcccccc}
\toprule
 & & & \multicolumn{2}{c}{\oracle{} $-$ \last{}} & \multicolumn{2}{c}{\oracle{} $-$ \valnormal{}} \\
\cmidrule(lr){4-5}\cmidrule(lr){6-7}
Setting & Det. & Units & Mean [95\% CI] & $\geq 2$ & Mean [95\% CI] & $\geq 2$ \\
\midrule
\multicolumn{7}{l}{\emph{Main grid, UB9, seed 0}} \\
Tuned hyperparameters (main result) & 12 & 9 & 5.12 [3.97, 6.24] & 12 & 4.13 [3.32, 4.79] & 12 \\
Default hyperparameters (\hpfree{} runs) & 12 & 9 & 5.11 [3.94, 6.21] & 12 & 4.43 [3.21, 5.52] & 11 \\
GLocalKD at commit c940457 (main grid) & 1 & 9 & 3.85 [1.61, 6.33] & -- & 3.84 [1.61, 6.32] & -- \\
GLocalKD at commit a96137b & 1 & 9 & 4.12 & -- & 4.11 & -- \\
GLADC at commit c940457 (main grid) & 1 & 9 & 5.07 [2.59, 7.46] & -- & 4.82 [2.55, 7.03] & -- \\
GLADC at commit a96137b & 1 & 9 & 4.89 & -- & 4.75 & -- \\
\midrule
\multicolumn{7}{l}{\emph{Extra training seeds, UB9 without REDDIT-B}} \\
Seed 0 & 12 & 8 & 5.48 [4.22, 6.63] & 12 & 4.49 [3.82, 5.16] & 12 \\
Seed 1 & 12 & 8 & 5.16 [3.71, 6.50] & 12 & 4.31 [3.58, 5.17] & 12 \\
Seed 2 & 12 & 8 & 5.33 [4.18, 6.37] & 12 & 4.37 [3.66, 5.00] & 12 \\
Mean of seeds 0 to 2 & 12 & 8 & 5.32 [4.06, 6.49] & 12 & 4.39 [3.71, 5.07] & 12 \\
\multicolumn{7}{l}{\emph{Six seeds (0 to 5) on the subset}} \\
Mean over seeds & 4 & 3 & 5.55 & -- & 5.53 & -- \\
SD over seeds & 4 & 3 & 0.42 & -- & 0.52 & -- \\
Minimum over seeds & 4 & 3 & 4.83 & -- & 4.99 & -- \\
Seeds with a positive gap & 4 & 3 & 6 of 6 & -- & 6 of 6 & -- \\
\midrule
\multicolumn{7}{l}{\emph{Second scenario}} \\
Type III, five dataset pairs & 12 & 5 pairs & 9.84 [6.73, 13.07] & 10 & 8.11 [5.36, 11.00] & 12 \\
\midrule
\multicolumn{7}{l}{\emph{Anomaly class on BZR, COX2 and PROTEINS\_full}} \\
UB-GOLD's anomaly class (class 0) & 12 & 3 & 5.78 [4.97, 6.62]$^\dagger$ & 12 & 4.83 [4.09, 5.58]$^\dagger$ & 11 \\
Minority class (class 1) & 12 & 3 & 2.96 [2.41, 3.52]$^\dagger$ & 8 & 5.00 [4.24, 5.82]$^\dagger$ & 9 \\
\midrule
\multicolumn{7}{l}{\emph{Subset, seed 0, all 12 cells}} \\
Original runs & 4 & 3 & 6.05 & -- & 6.31 & -- \\
Isolated test evaluations & 4 & 3 & 6.01 & -- & 6.14 & -- \\
All normal graphs & 4 & 3 & 5.72 & -- & n/a & -- \\
\multicolumn{7}{l}{\emph{Subset, seed 0, without the GPU cell (11 cells)}} \\
Original runs & 4 & 3 & 5.97 & -- & 6.64 & -- \\
Isolated test evaluations & 4 & 3 & 6.02 & -- & 6.14 & -- \\
All normal graphs & 4 & 3 & 5.76 & -- & n/a & -- \\
\bottomrule
\end{tabular}

\end{table}

Every mean gap is positive, ranging from 2.96 points against \last{} with minority-class anomalies to 9.84 in Type III.
Extra seeds, isolated test evaluations, all-normal training, the earlier commit, and default hyperparameters each change a gap by at most half a point.
With defaults, the two gaps are 5.11 and 4.43, so checkpoint leakage does not require test-tuned hyperparameters.
Not every setting has all 12 detectors reaching 2 points.
Against \valnormal{}, 11 detectors reach the threshold with defaults and in the three-dataset check that retains the original anomaly class.
With minority-class anomalies, the counts are 8 against \last{} and 9 against \valnormal{}.
In Type III, 10 detectors reach the threshold against \last{}.
Within the tested settings, the mean gap persists after changing training seeds, the Types I/II scenario, anomaly classes, random-number sharing between evaluation and training, the normal hold-out, and the code version affected by the pull request.
Limitations include no extra seeds on REDDIT-B, one seed for isolation and all-normal training, only five Type III pairs, and GPU nondeterminism.

\subsection{Other label-free selectors}
\label{app:posthoc-selectors}
\label{app:extra-selectors}

A weak label-free comparison rule could enlarge the gap, so A32 specified seven checkpoint selectors before their computation.
S1 is \last{} and S2 is \valnormal{}.
S3 minimizes the median held-out score, S4 minimizes training loss, and S5 selects the checkpoint nearest half the training budget.
S6 averages the recorded test AUROCs over checkpoints, giving the expected AUROC when a checkpoint is selected uniformly at random.
S7 minimizes the mean held-out score after standardizing scores with normal-training statistics.
S3 requires per-graph held-out scores and covers only the eight formal-test UB-GOLD detectors.
S7 requires standardized scores, which CVTGAD records and GLADMamba, SEGO, and GLASS already use as outputs, so S7 equals S2 for the latter three.

\begin{figure}[!ht]
  \centering
  \includegraphics{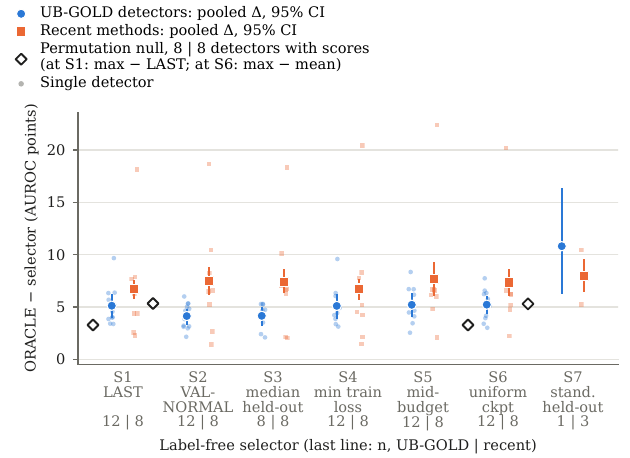}
  \caption{Gaps between \oracle{} and label-free selectors. Permutation null denotes the label-permutation check.}
  \label{fig:posthoc-selectors}
\end{figure}

Figure~\ref{fig:posthoc-selectors} shows \oracle{} minus each label-free selector for individual detectors and their means, with intervals.
The figure shows UB-GOLD detectors and the eight recent methods with checkpoints separately, and includes the label-permutation results from Appendix~\ref{app:posthoc-null}.
The last line below the axis gives the detector coverage of each group.

Every interval for a mean gap lies above zero, meeting the condition in A32.
Across the six broadly available selectors, UB-GOLD mean gaps range from 4.13 [3.32, 4.79] under S2 to 5.22 [4.34, 6.08] under S6.
For UB-GOLD, S7 covers only CVTGAD and gives a gap of 10.81 [6.28, 16.34].
Recent-method gaps range from 6.69 [5.75, 7.61] under S1 to 7.97 [6.45, 9.60] under S7.
Among selectors that cover all 12 UB-GOLD detectors, \valnormal{} gives the smallest mean gap.
On the eight detectors that S3 covers, S3 gives a slightly smaller gap than \valnormal{}, 4.15 versus 4.25.
Each of the eight formal-test UB-GOLD detectors gains at least 2 points under S1 and S2, so the G1 result extends beyond the four detectors whose results informed the criterion.

Individual intervals include zero for SIGNET under S2, LGKDE under S1--S4, HC-GLAD under S2--S4, and SUL/GUL under S2, S3, and S5.
The best rule also depends on the detector.
Held-out scores work well for both InfoGraph detectors but leave gaps of 6.64--10.44 points for GLADMamba and SEGO, compared with 4.22--4.50 for \last{} and training loss.
CVTGAD's S7 selects a checkpoint within the first 10\% of training in 91\% of runs, as in the early-selection pattern recorded for GLADMamba in A14.

An earlier pilot specified six alternative rules before running.
They use node- and edge-count-preserving pseudo-anomalies, an average over the three checkpoints selected as best by those pseudo-anomalies, checkpoint ensembling after \citet{robod}, a consensus checkpoint after \citet{glam}, transductive AUCp~\citep{aucp}, and \last{}.
None exceeds \valnormal{} on average over the four pilot detectors, scoring 0.00--0.51 points lower while \oracle{} scores 4.26 points higher.
AUCp improves OCGIN and GLocalKD but lowers SIGNET by 6.5 points.
The supplementary study record gives individual-rule results, and better label-free selectors remain possible.
The next analyses measure how much selecting on the reporting graphs raises the reported score.

\subsection{Label-permutation check}
\label{app:posthoc-null}

Taking the maximum of noisy test evaluations can raise the reported score even without a systematically better checkpoint.
For runs with per-graph scores, we permute the test labels 200 times while keeping the anomaly count fixed.
For each permutation, we subtract \last{} from the highest recorded AUROC over the full run.
We average these differences over permutations, then over datasets and detectors as for the observed gap.
No checkpoint is systematically better under this label permutation, so the resulting gap shows how much selection on noise alone can produce.
The maximum uses every recorded evaluation rather than stopping after 30 evaluations without improvement.
It therefore gives an upper bound for UB-GOLD's rule, not the fraction of the observed gap caused by noise.

The A32 analysis uses the UB9 main-comparison runs of eight formal-test UB-GOLD detectors with per-graph scores and eight recent methods with checkpoints.
For UB-GOLD detectors, the null gain is 3.28 [2.20, 4.45], compared with an observed gap of 5.56, giving a ratio of 0.59.
The corresponding recent-method values are 5.34 [4.02, 6.77], 6.69, and 0.80.
The null is at least as large as the observed gap for 6 of the 16 methods, namely CVTGAD, GOOD-D, LGKDE, GLADMamba, SEGO, and HC-GLAD.
GLADPro and SUL/GUL have the most evaluation points and the largest null gains, at 9.05 and 7.54, while the two InfoGraph detectors have the smallest ratios, at 0.34 and 0.25.

The extra-seed analysis C stores per-graph scores from three seeds for all 12 UB-GOLD detectors on eight datasets.
As specified in A41, runs with the same detector, dataset, and fold share the same 200 label permutations across seeds.
Averaged across seeds, the null gain is 3.80 [2.62, 4.98] against an observed gap of 5.32 [4.06, 6.49], giving the ratio of 0.71 used in the main text.
The three seeds give null gains of 3.75, 3.82, and 3.81, with ratios from 0.69 to 0.74.
This analysis uses different detectors, datasets, and pilot-detector seed-0 runs, so it supplements rather than replaces the main-comparison analysis.
Both show that selection on noise can produce gains exceeding half the observed gap, subject to the upper-bound limitation of the unrestricted maximum.

\subsection{Extra training seeds and the seed-transfer check}
\label{app:followup-seeds}

\paragraph{Design and devices.}
To examine training randomness, A36 adopted the extra-seed analysis C, and A41 confirmed its analysis details before any extra-seed results.
C adds two seeds for all 12 tuned detectors on the eight UB9 datasets other than REDDIT-B.
Fold $f$ of seed $k$ calls \texttt{set\_seed}$(3407+f+1000k)$, leaves all other settings unchanged, and stores per-graph scores.
REDDIT-B is excluded because it would require about 840 of the estimated 1140 core-hours and nodes with 512\,GB memory.
Before any extra-seed results, A40 added the extension C2, assigning seeds 3--5 to a subset chosen without using results.
The selected datasets are NCI1, PROTEINS\_full, and IMDB-B, the largest available dataset in each TU category.
The selected detectors are GraphCL-OCSVM, OCGIN, GOOD-D, and GLocalKD, the least costly detector in each of four training-objective groups.
C adds 960 runs to 480 seed-0 runs for a total of 1440, and C2 adds 180 more.

For each detector, dataset, and fold, the extra seeds use the device from the main comparison, with two specified exceptions.
The four pilot detectors use CPU because their main-grid GPU runs have no per-graph scores.
Their seed 0 uses CPU runs from the pilot code path that match our evaluation code bit for bit, without changing the main table.
Two main-grid GraphDE runs on AIDS switched from GPU to CPU, and their extra seeds use GPU, which handled most epochs.
Together with excluding REDDIT-B, these choices give seed-0 gaps of 4.49 against \valnormal{} and 5.48 against \last{}, compared with 4.13 and 5.12 in the main grid.
Except for GLADC, GPU runs can differ even with the seed fixed.
Differences across GPU training seeds therefore also include variation from device execution (Appendix~\ref{app:ports-devices}).

\paragraph{Mean stability and run variation.}
Mean gaps across datasets and detectors remain nearly constant across seeds, with every detector gaining at least 2 points over both rules in every seed.
Reporting the best of three seeds instead of their mean raises \last{} by 2.57 [2.06, 3.03], with nine detectors gaining at least 2 points from this choice alone.
Table~\ref{tab:followup-seeds-det} reports each detector's gaps over three seeds.
For each detector, it also reports the mean and maximum across-seed standard deviation over its 40 dataset--fold combinations.
The mean standard deviation across dataset--fold combinations ranges from 0.83 for OCGTL under \oracle{} to 4.12 for GraphCL-OCSVM under \last{}.
The maximum is 17.65 for GLocalKD under \valnormal{}.
Stable average leakage therefore does not imply stable individual runs.

\begin{table}[!ht]
  \centering\footnotesize
  \setlength{\tabcolsep}{4pt}
  \caption{Checkpoint leakage and seed spread by detector.}
  \label{tab:followup-seeds-det}
  \begin{tabular}{lcc@{\hspace{12pt}}ccc@{\hspace{12pt}}ccc}
\toprule
 & \multicolumn{2}{c}{\oracle{} minus, three seeds} & \multicolumn{3}{c}{Mean cell SD} & \multicolumn{3}{c}{Largest cell SD} \\
\cmidrule(lr){2-3}\cmidrule(lr){4-6}\cmidrule(lr){7-9}
Detector & \last{} & \valnormal{} & \oracle{} & \last{} & \valnormal{} & \oracle{} & \last{} & \valnormal{} \\
\midrule
SIGNET & 3.37 & 2.73 & 1.95 & 2.46 & 1.88 & 7.02 & 6.55 & 7.10 \\
OCGTL & 3.63 & 3.49 & 0.83 & 1.48 & 1.41 & 3.26 & 4.19 & 4.19 \\
OCGIN & 5.72 & 5.40 & 1.63 & 2.47 & 2.36 & 5.34 & 6.87 & 6.87 \\
GLocalKD & 5.31 & 5.30 & 2.88 & 3.74 & 3.75 & 10.89 & 17.12 & 17.65 \\
GLADC & 5.61 & 5.52 & 1.09 & 2.00 & 1.92 & 3.49 & 6.38 & 5.88 \\
CVTGAD & 5.43 & 5.37 & 2.05 & 2.48 & 2.50 & 7.87 & 6.89 & 6.52 \\
GOOD-D & 4.35 & 4.37 & 3.26 & 3.25 & 3.84 & 10.48 & 9.77 & 15.81 \\
GraphDE & 2.33 & 2.30 & 0.85 & 0.93 & 1.03 & 5.19 & 5.72 & 4.25 \\
GraphCL-IF & 4.62 & 4.00 & 2.32 & 3.47 & 3.03 & 7.06 & 13.02 & 8.81 \\
GraphCL-OCSVM & 6.32 & 5.63 & 1.98 & 4.12 & 2.98 & 8.99 & 14.67 & 8.81 \\
InfoGraph-IF & 10.13 & 5.51 & 2.26 & 3.66 & 2.13 & 5.94 & 14.01 & 5.37 \\
InfoGraph-OCSVM & 7.04 & 3.05 & 1.30 & 2.35 & 2.44 & 3.91 & 6.05 & 9.00 \\
\midrule
All 12 detectors & 5.32 & 4.39 & 1.87 & 2.70 & 2.44 & 10.89 & 17.12 & 17.65 \\
\bottomrule
\end{tabular}

\end{table}

\paragraph{Transfer of the selected epoch.}
We select the \oracle{} epoch in seed $s$ and evaluate the checkpoint at that epoch in seed $s'$ for the same detector, dataset, and fold.
We subtract the second seed's \last{} and average the six ordered seed pairs as specified in A41.
Table~\ref{tab:followup-seeds-transfer} gives the transferred gain, the within-seed gain, and their difference.
This difference is the part not retained across seeds.
It may include both selection noise and a real checkpoint advantage limited to one training run, so it is not a pure noise estimate.

\begin{table}[!ht]
  \centering\footnotesize
  \caption{Transfer of the \oracle{} epoch across seeds.}
  \label{tab:followup-seeds-transfer}
  \begin{tabular}{lccc}
\toprule
 & Transfer to another seed & Within seed & Run-specific part \\
Detector & $-$ \last{} & \oracle{} $-$ \last{} & within $-$ transfer \\
\midrule
SIGNET & $-$2.69 & 3.37 & 6.06 \\
OCGTL & 3.02 & 3.63 & 0.61 \\
OCGIN & 3.96 & 5.72 & 1.76 \\
GLocalKD & 4.20 & 5.31 & 1.11 \\
GLADC & 3.32 & 5.61 & 2.29 \\
CVTGAD & 1.32 & 5.43 & 4.11 \\
GOOD-D & $-$0.83 & 4.35 & 5.18 \\
GraphDE & 0.19 & 2.33 & 2.14 \\
GraphCL-IF & 1.82 & 4.62 & 2.80 \\
GraphCL-OCSVM & 2.79 & 6.32 & 3.53 \\
InfoGraph-IF & 7.15 & 10.13 & 2.98 \\
InfoGraph-OCSVM & 6.34 & 7.04 & 0.70 \\
\midrule
Mean of 12 [95\% CI] & 2.55 [1.16, 4.02] & 5.32 [4.06, 6.49] & 2.77 [1.97, 3.57] \\
Detectors $\geq 2$ & 7 of 12 & 12 of 12 & 8 of 12 \\
\bottomrule
\end{tabular}

\end{table}

The mean transferred gain is 2.55 [1.16, 4.02], about half the within-seed gap of 5.32, leaving a run-specific part of 2.77 [1.97, 3.57].
More than half the advantage transfers for both InfoGraph detectors, OCGTL, OCGIN, GLocalKD, and GLADC, consistent with their selected earlier epochs also tending to work better in other runs.
For SIGNET and GOOD-D, checkpoints chosen using another seed score below \last{}, so none of the observed gain is retained across seeds.
This analysis separates gains retained across seeds from gains limited to one run, but does not determine a single cause of the full gap.

\paragraph{Rankings and six seeds.}
With the rule fixed, cross-seed leaderboard $\tau_b$ ranges from 0.85 to 0.94, and the top three change in two of the three seed pairs under every rule.
Within a seed, \oracle{} versus \valnormal{} gives $\tau_b=0.73$--0.82, with intervals reaching 0.39.
Changing the checkpoint rule reorders detectors only somewhat more than changing the seed.
The seed-0 value is 0.73 rather than the main-grid 0.76 because this analysis covers eight datasets and uses pilot CPU runs.
On the six-seed subset, the mean gap against \last{} is 5.55, with a seed standard deviation of 0.42 and a minimum of 4.83.
Against \valnormal{}, the corresponding values are 5.53, 0.52, and 4.99, and both mean gaps are positive in all six seeds.
Except for GLocalKD on NCI1, cell means against \last{} range from 0.60 for GLocalKD on IMDB-B to 12.69 for GLocalKD on PROTEINS\_full.
The maximum against \valnormal{} is 13.82 for GraphCL-OCSVM on IMDB-B, and these cells are positive in every seed.
For GLocalKD on NCI1, both label-free rules select the final epoch, and the earlier \oracle{} epoch differs by less than 0.01 AUROC points, leaving both gaps effectively zero.

\subsection{Disjoint labeled split}
\label{app:followup-split}

The cross-seed analysis changes the training run but keeps the test graphs.
Here we keep the run and use different graphs for selection and evaluation.
It measures the benefit of labeled selection and the optimism from selecting and reporting on the same graphs.
A36 specified the labeled-split analysis E.
A45 fixed its run set after inspecting the Type III (A), extra-seed (C and C2), and low-order-reference (H) results, but before computing any statistic for part E.
The primary set contains the 480 seed-0 runs from part C, with seeds 1 and 2 used only for descriptive repeats.
Each run uses 200 stratified splits.
For each label class with $n$ graphs, $\lceil n/2\rceil$ graphs enter the selection half.
Runs with the same detector, dataset, and fold share the splits across seeds.
UB-GOLD's rule selects a checkpoint on the selection half, and \eselect{} is that checkpoint's AUROC on the other, evaluation half.
We also recompute \oracle{} by applying UB-GOLD's rule to the evaluation half, evaluate \last{} and \valnormal{} there, and average splits before datasets and detectors.
Some halves are small, with 40 graphs and 31 or 32 anomalies in the BZR evaluation half, and 60 graphs with 10 anomalies in ENZYMES.

A46 confirmed rounding, random-draw order, AUROC arithmetic, and per-seed reporting after the results, although these choices were already fixed in code saved before running.
UB-GOLD changes checkpoints only for a strictly larger AUROC, making the selected epoch sensitive to floating-point operation order.
Scikit-learn's trapezoid arithmetic can separate theoretically identical AUROCs by one unit in the last place and select a different epoch from an exact rank-sum formula.
The half-set analysis therefore follows \texttt{roc\_auc\_score}'s operation order rather than substituting exact arithmetic.
Recalculation on the full test sets reproduces the selected epochs and AUROCs of \oracle{}, \last{}, and \valnormal{} bit for bit for all 1440 runs across the three seeds.

Table~\ref{tab:followup-split-pooled} summarizes protocol scores and differences on the evaluation half, and Table~\ref{tab:followup-split-det} gives seed-0 detector differences.
\oracle{} and \eselect{} use the same rule, but only \oracle{} uses the reporting graphs to select a checkpoint.
Their difference estimates the extra reported score from selecting on those graphs in this design.
For seed 0, optimism is 3.64 [1.70, 5.64], while \eselect{} exceeds \last{} by 2.86 [0.88, 4.94] and \valnormal{} by 1.86 [0.36, 3.38].
A48 added evaluation-half \oracle{} minus \last{} to the mechanism display as the sum of two reported differences.
For seeds 0, 1, and 2, this gap is 6.50 [5.11, 7.65], 6.28 [4.69, 7.57], and 6.37 [5.05, 7.52], respectively.
The seed-0 value exceeds 5.48 on the same runs' full test sets, consistent with greater optimism when maximizing noisier half-set evaluations.
This sum explains the gap on the evaluation half, not the exact full-test gap in the main comparison.

\begin{table}[!ht]
  \centering\footnotesize
  \caption{Mean results of disjoint labeled selection.}
  \label{tab:followup-split-pooled}
  \begin{tabular}{lccc}
\toprule
Quantity on the evaluation half & Seed 0 & Seed 1 & Seed 2 \\
\midrule
\eselect{} & 64.50 [58.01, 73.69] & 64.34 [57.91, 73.49] & 64.31 [57.81, 73.49] \\
\oracle{} & 68.14 [62.03, 76.37] & 68.01 [61.98, 76.23] & 68.01 [61.85, 76.23] \\
\last{} & 61.64 [55.82, 70.06] & 61.73 [55.99, 70.16] & 61.64 [55.89, 69.97] \\
\valnormal{} & 62.64 [56.42, 71.63] & 62.57 [56.58, 71.41] & 62.60 [56.34, 71.64] \\
\midrule
\oracle{} $-$ \eselect{} & 3.64 [1.70, 5.64] & 3.67 [1.76, 5.66] & 3.70 [1.73, 5.75] \\
\eselect{} $-$ \last{} & 2.86 [0.88, 4.94] & 2.61 [0.53, 4.71] & 2.67 [0.85, 4.46] \\
\eselect{} $-$ \valnormal{} & 1.86 [0.36, 3.38] & 1.76 [0.25, 3.36] & 1.71 [0.34, 3.05] \\
\bottomrule
\end{tabular}

\end{table}

\begin{table}[!ht]
  \centering\footnotesize
  \caption{Disjoint labeled selection by detector, seed 0.}
  \label{tab:followup-split-det}
  \begin{tabular}{lccc}
\toprule
Detector & \oracle{} $-$ \eselect{} & \eselect{} $-$ \last{} & \eselect{} $-$ \valnormal{} \\
\midrule
SIGNET & 6.33 [3.23, 9.43] & $-$1.45 [$-$5.94, 3.68] & $-$2.11 [$-$6.85, 3.56] \\
OCGTL & 3.53 [1.44, 5.79] & 1.36 [$-$1.54, 4.31] & 1.07 [$-$1.98, 4.08] \\
OCGIN & 3.65 [1.85, 5.38] & 3.64 [0.67, 6.85] & 3.24 [0.26, 6.57] \\
GLocalKD & 2.03 [0.54, 3.75] & 3.86 [0.75, 7.73] & 3.86 [0.75, 7.73] \\
GLADC & 2.66 [0.75, 4.80] & 3.79 [1.78, 5.95] & 3.49 [1.46, 5.78] \\
CVTGAD & 5.06 [2.44, 7.97] & 1.56 [$-$0.73, 4.24] & 1.50 [$-$0.93, 4.35] \\
GOOD-D & 6.34 [3.12, 9.70] & 0.28 [$-$3.20, 3.60] & 0.41 [$-$3.25, 4.14] \\
GraphDE & 2.82 [0.99, 5.13] & 0.81 [0.17, 1.69] & 0.48 [0.10, 0.98] \\
GraphCL-IF & 3.04 [1.11, 5.15] & 2.22 [0.62, 3.86] & 1.32 [$-$0.55, 3.10] \\
GraphCL-OCSVM & 4.04 [1.81, 6.35] & 3.70 [1.26, 5.96] & 3.22 [0.40, 6.15] \\
InfoGraph-IF & 3.04 [1.11, 4.93] & 8.44 [3.44, 14.46] & 3.58 [0.44, 7.24] \\
InfoGraph-OCSVM & 1.12 [0.39, 2.05] & 6.15 [1.49, 11.51] & 2.30 [0.05, 5.07] \\
\bottomrule
\end{tabular}

\end{table}

Optimism is positive for every detector, from 1.12 for InfoGraph-OCSVM to 6.34 for GOOD-D, with all 12 intervals above zero.
Benefits of labeled selection are less uniform, with gains over \last{} of 8.44 [3.44, 14.46] for InfoGraph-IF, 0.28 for GOOD-D, and $-1.45$ [$-5.94$, 3.68] for SIGNET, and the same pattern in seeds 1 and 2.
Detectors whose selected epochs work better across seeds also benefit most from separate labeled graphs, whereas SIGNET and GOOD-D show no meaningful gain.
Of this design's 6.50-point gap, 3.64 points are same-graph optimism unavailable to a selector without the reporting graphs, and 2.86 points are gains from better checkpoints found with labels.
A larger labeled selection set might increase the gain from finding better checkpoints.
The split, cross-seed, and label-permutation analyses therefore answer related but different questions.

\subsection{Isolated test evaluations}
\label{app:followup-iso}

UB-GOLD periodically evaluates test graphs during training, and operations such as shuffled loaders and augmentations share random-number generators with training and may change later updates.
Test-score computation uses no labels, but the training trajectory can depend on the frequency of test evaluations.
The main grid retains this released behavior, and the isolation analysis F tests its effect on leakage and \hpfree{}.
A42 specified F before the Type III (A), extra-seed (C and C2), or labeled-split (E) results, and A44 and A47 confirmed its implementation before running.
Before each evaluation block, we save the random-number generator states used by training in Python, NumPy, CPU PyTorch, and CUDA PyTorch.
After the block finishes, we restore those saved states.
This includes held-out scoring and the pretraining test-loader passes of GLocalKD and GLADC.
Smoke tests for each type of training loop produce the same training losses as runs with no test evaluations.
The 120 runs cover both hyperparameter settings of the seed-0 subset.
Tuned GraphCL-OCSVM on PROTEINS\_full uses GPU and differs from epoch 1 or 2 even in identical reruns, so we also report averages without this cell.

Across all 12 cells, isolation changes \oracle{}, \last{}, and \valnormal{} by 0.28, 0.32, and 0.46 points, below their respective six-seed standard deviations of 0.36, 0.57, and 0.56.
Without the GPU cell, the changes are 0.32, 0.28, and 0.82.
The mean gaps against \last{} and \valnormal{} move from 6.05 and 6.31 to 6.01 and 6.14, respectively.
Without the GPU combination, the gap against \valnormal{} falls from 6.64 to 6.14 because \valnormal{} scores higher.
Cell changes have both signs and are mostly within two seed standard deviations, ranging from 3.29 for GLocalKD on PROTEINS\_full under \oracle{} to $-3.54$ for the GPU cell under \valnormal{}.
The GPU cell's gap against \valnormal{} rises from 2.69 to 6.12, while OCGIN on NCI1 falls from 2.83 to 0.37, largely canceling in the mean.
\hpfree{} changes by 0.42 on average, with cell changes from $-1.26$ to 3.58.
Isolation has effects comparable to changing the seed and leaves both mean leakage and the overall label-free reference conclusions intact.

\subsection{Training on all normal graphs}
\label{app:followup-allnormals}

The main grid holds out 20\% of normal training graphs for \valnormal{}, whereas UB-GOLD trains on all normal graphs.
If withholding graphs harmed the final checkpoint more than the best checkpoint, it could enlarge the gap against \last{}.
In the pilot, SIGNET's gap is 4.47 on UB-GOLD's exact UB9 training sets versus 4.01 with the hold-out, with full training raising \oracle{} and \last{} by 0.84 and 0.39.
The all-normal analysis G was specified in A42 and confirmed in A44 and A47 before adding 60 tuned seed-0 subset runs.
A marked copy of the run function skips withholding and held-out scoring without changing training random draws.
Smoke tests confirm bit-for-bit reproduction of stored CPU runs with the 20\% hold-out and training sets identical to UB-GOLD's without it.
Larger training sets also change batch order, so this comparison changes both training-set size and the sequence of random draws.

Averaged over 12 cells, all-normal training changes \oracle{} by $-0.32$ and \last{} by 0.01, moving the gap from 6.05 to 5.72.
Without the GPU cell, these values are $-0.29$, $-0.08$, and 5.76.
Cell changes in \last{} range from $-2.67$ for OCGIN on NCI1 to 3.95 for GraphCL-OCSVM on IMDB-B, whose seed spread is 5.27.
The few larger changes have no common direction, and the hold-out neither lowers mean \last{} nor creates the gap.
This check does not assess \valnormal{}, which requires held-out graphs.

\subsection{An earlier UB-GOLD commit}
\label{app:extra-version}

An external pull request in 2025 changed the training batches and data-loader random draws of GLocalKD and GLADC.
Our runs use c940457, which includes the change, while UB-GOLD's published values most likely use earlier code.
To assess sensitivity to the UB-GOLD code version (Appendix~\ref{app:defects-pr3}), we specified descriptive reruns at the preceding commit a96137b, matching folds, hyperparameters, seeds, hold-out splits, and paired device types.
Two edits only repair parsing errors without changing behavior.
On CPU, the pull request changes the position in the random-number sequence by two draws for GLADC and four for GLocalKD.
The comparison therefore changes both batches and the sequence of random draws.
Table~\ref{tab:extra-version} gives UB9 means at both commits and paired differences over 45 folds per detector.

\begin{table}[!ht]
  \centering\footnotesize
  \caption{GLocalKD and GLADC at two UB-GOLD commits.}
  \label{tab:extra-version}
  \begin{tabular}{llcccc}
\toprule
Detector & Quantity & c940457 & a96137b & Paired difference & 95\% CI (dataset bootstrap) \\
\midrule
GLocalKD & \oracle{} & 66.36 & 66.41 & +0.05 & [$-$0.01, 0.14] \\
GLocalKD & \last{} & 62.51 & 62.29 & $-$0.22 & [$-$0.62, 0.10] \\
GLocalKD & \valnormal{} & 62.52 & 62.30 & $-$0.22 & [$-$0.62, 0.10] \\
GLocalKD & \oracle{} $-$ \valnormal{} & 3.84 & 4.11 & +0.27 & -- \\
GLocalKD & \oracle{} $-$ \last{} & 3.85 & 4.12 & +0.27 & -- \\
\midrule
GLADC & \oracle{} & 68.22 & 68.11 & $-$0.12 & [$-$0.26, 0.01] \\
GLADC & \last{} & 63.15 & 63.21 & +0.06 & [$-$0.31, 0.59] \\
GLADC & \valnormal{} & 63.40 & 63.36 & $-$0.04 & [$-$0.63, 0.57] \\
GLADC & \oracle{} $-$ \valnormal{} & 4.82 & 4.75 & $-$0.07 & -- \\
GLADC & \oracle{} $-$ \last{} & 5.07 & 4.89 & $-$0.18 & -- \\
\bottomrule
\end{tabular}

\end{table}

All 90 runs finish, mean AUROC changes by at most 0.22 points, and all six difference intervals include zero.
Leakage changes by at most 0.27 points for either detector against either label-free rule.

\subsection{The minority class as anomaly}
\label{app:extra-orientation}

UB-GOLD treats class 0 as anomalous, although it is the majority class on BZR, COX2, and PROTEINS\_full, contrary to the paper's convention.
This check was specified before running and reruns all 12 detectors on these three datasets with class 1 anomalous.
UB-GOLD's split code returns the same test folds after swapping labels, changing the test labels and normal training graphs.
Seeds, hold-out, devices, and UB-GOLD hyperparameters tuned for the opposite anomaly orientation remain fixed.
The check tests whether the sign of leakage depends on the anomaly class under these configurations, without retuning for the new orientation.

With class 1 anomalous, \oracle{} exceeds \valnormal{} by 5.00 [4.24, 5.82], compared with 4.83 [4.09, 5.58] for class 0.
These intervals resample folds because only three datasets are available.
The gap against \last{} falls from 5.78 to 2.96 [2.41, 3.52].
With the minority class anomalous, nine detectors gain at least 2 points against \valnormal{} and eight against \last{}, compared with 11 and 12 for class 0.
Absolute scores change more, with OCGTL's \oracle{} rising from 64.5 to 76.5 while GLADC and the four two-stage detectors fall to 44.9--51.4.
GraphCL-OCSVM and both InfoGraph detectors score below 40 and below \last{} under \valnormal{}, leaving gaps of 7.7--14.0 against \valnormal{}.
With class 1, size alone scores 63.9 on BZR and 65.5 on PROTEINS\_full, compared with 42.5 and 57.8 for class 0.
The anomaly class affects scores and the gap against \last{}, but mean leakage remains positive.

\subsection{Dataset sensitivity and scope of the intervals}
\label{app:posthoc-datasets}

\begin{figure}[!ht]
  \centering
  \includegraphics[width=\textwidth]{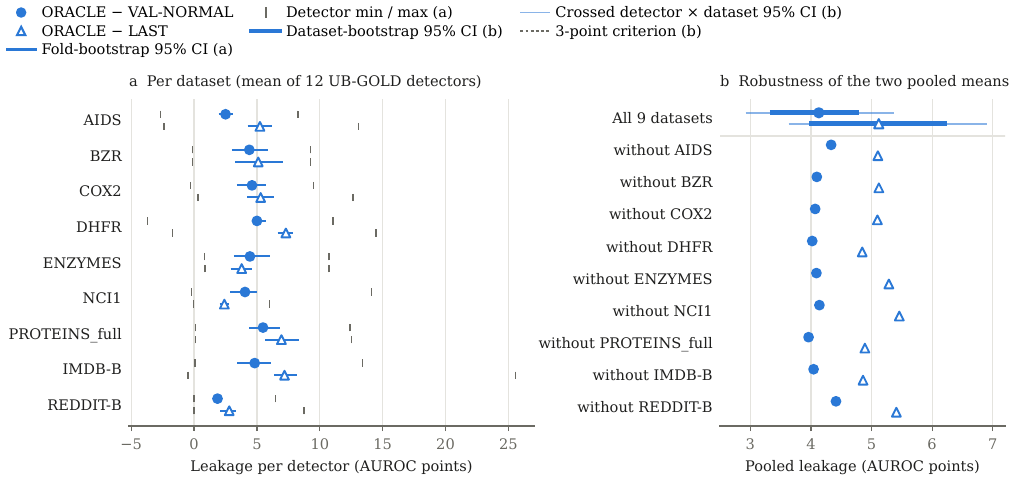}
  \caption{Dataset sensitivity of checkpoint leakage.}
  \label{fig:posthoc-forest}
\end{figure}

Figure~\ref{fig:posthoc-forest} tests whether one dataset drives leakage and compares the sources of variation represented by the intervals.
Panel a shows the two gaps for each dataset averaged across detectors, and panel b shows overall means under two interval definitions and after removing each dataset.

Every dataset has positive mean leakage, with at least 11 detectors showing a positive gap on each dataset.
Dataset means range from 1.86 on REDDIT-B to 5.48 on PROTEINS\_full against \valnormal{}, and from 2.41 on NCI1 to 7.30 on DHFR against \last{}.
Individual detector-dataset differences range from $-3.71$ to 25.57.
After removing any one dataset, mean gaps remain 3.96--4.41 against \valnormal{} and 4.85--5.46 against \last{}, above the criterion's 3-point threshold.
Resampling detectors as well as datasets widens the intervals to [2.92, 5.36] and [3.63, 6.90], with the first extending below 3.
This treats the evaluated detectors as a sample of the GLAD-detector population.
Our intervals therefore describe how results for the evaluated detectors, all of UB-GOLD's deep detectors, vary with the UB9 datasets included.
They do not support conclusions about the full detector population.

\subsection{Recent-method gaps after excluding uncertain reproductions}
\label{app:posthoc-fidelity}

Finally, we check whether the mean gap for recent methods depends on uncertain reproductions, as documented in Appendix~\ref{app:ports-fidelity}.
The full set gives gaps of 7.47 [6.15, 8.83] against \valnormal{} and 6.69 [5.75, 7.61] against \last{}.
Removing LGKDE, which fails its reproduction check, gives 8.16 [6.71, 9.58] and 7.32 [5.73, 8.93].
Also removing GLADPro, whose check is contested, and SUL/GUL, whose check is not evaluable, gives 6.38 [5.39, 7.43] and 5.05 [3.65, 6.69].
The mean gap changes but remains positive.
These descriptive results apply only to the tested settings.

\section{Anomalies from another dataset (Type III)}
\label{app:t3}

The UB9 main grid uses UB-GOLD's Types I and II, where normal and anomalous graphs come from the same dataset.
This appendix tests whether checkpoint leakage persists when anomalies come from another dataset.
UB-GOLD calls this Type III scenario inter-dataset out-of-distribution detection, using one TU dataset for normal graphs and another for anomalies.
The results extend the checkpoint comparison in Section~\ref{sec:f2-mech} and the simple-reference comparison in Section~\ref{sec:c5} to anomalies from another dataset.
This replication is the only follow-up study with a decision criterion, R1, with its code audit in Appendix~\ref{app:t3-code}.
PROTEINS here denotes the TU dataset of that name, distinct from PROTEINS\_full in UB9.

\subsection{Datasets, training, and intervals}

Before any Type III run, amendment A36 fixed the scenario, protocol, and R1 in part A of the follow-up plan.
A37 confirmed implementation details after a smoke test that computed no metrics.
We use UB-GOLD's five TU pairs, listing the normal in-distribution (ID) dataset before the anomaly-source dataset and retaining UB-GOLD's names, including IMDB-BINARY.
Trial $i$ uses seed $3407+i$ and the split function copied verbatim from c940457.
The function randomly splits ID graphs 90/10 into training and test sets, then adds an equally sized block from the beginning of the second dataset's file as anomalies.
Normal test graphs vary across trials, but every trial reuses the same anomaly block.
The anomaly blocks for ENZYMES+PROTEINS and IMDB-MULTI+IMDB-BINARY each contain graphs from only one class.
They contain 60 PROTEINS graphs and 150 IMDB-BINARY graphs, respectively.
UB-GOLD replaces every node feature within a TU pair with a constant, leaving structure-only tasks.

All trial indices were fixed by hash before running, and an independent implementation and PyTorch 2.4.0, 2.5.1, and 2.14.1 produce identical lists.
We could not run UB-GOLD's specified PyTorch 2.1 environment, so the checks above support but do not prove that its splits are identical.
Each trial holds out 20\% of ID training graphs for \valnormal{} and trains the 12 UB-GOLD detectors on the rest with the changes documented in Appendix~\ref{app:ports}.
Tuned configurations use each detector's Type III script line, and defaults follow the main-grid rule.
SIGNET's Type III script lines are commented out and contain two misspelled flags that prevent parsing.
We correct only those spellings, retain all other values, and mark SIGNET's rows with an asterisk.

As in the main comparison, CVTGAD's \oracle{} selects using normalized scores and is marked with a dagger.
GraphDE uses its paper's score direction and is marked with a double dagger.
Two-stage detectors use scores aligned with the graph labels.
All 600 runs, covering 12 detectors, five pairs, five trials, and two hyperparameter settings, use bit-reproducible CPU execution.
The size-only and graphlet-count references fit all ID training graphs of each trial, using graphlets of up to four nodes on IMDB-MULTI+IMDB-BINARY.
We average the five trials within each pair, weight pairs equally, and then average detectors.
Intervals use 4000 bootstrap resamples of the five pairs with detectors fixed.
Only a few dataset pairs are available, so these intervals have limited resolution.
Resampling trials would give optimistic intervals because trials overlap and reuse the same anomaly graphs.

\subsection{Decision criterion and detector results}

The decision criterion R1 reuses the main-grid leakage thresholds.
Both \oracle{} minus \valnormal{} and \oracle{} minus \last{} must average at least 3 points across detectors.
For each comparison, at least 75\% of complete detectors must gain at least 2 points on average.
A complete detector finishes all five pairs with tuned hyperparameters, and R1 is evaluated only if at least eight detectors are complete.
A36 expressed the count as 9 of 12, and A37 clarified before any grid run that the same fraction applies to complete detectors (Appendix~\ref{app:prereg-v2}).

Table~\ref{tab:t3-detectors} reports four protocols, two checkpoint gaps, and the hyperparameter gap for each detector, first averaging five trials and then pairs.
The final rows give the detector mean, pair-bootstrap interval, and count gaining at least 2 points.
All 12 detectors complete every pair.
\oracle{} exceeds \valnormal{} by 8.11 [5.36, 11.00], and all 12 detectors gain at least 2 points.
It exceeds \last{} by 9.84 [6.73, 13.07], with 10 detectors reaching 2 points.
GLADC at 1.34 and CVTGAD at 1.99 fall below the threshold against \last{}.
Both conditions hold, so R1 is met, and both gaps exceed their UB9 counterparts of 4.13 and 5.12.

\begin{table}[!htb]
  \centering\scriptsize
  \setlength{\tabcolsep}{3pt}
  \caption{Type III results by detector.}
  \label{tab:t3-detectors}
  \begin{tabular}{lccccccc}
\toprule
 & \multicolumn{4}{c}{AUROC} & \multicolumn{2}{c}{\oracle{} minus} & \valnormal{} \\
\cmidrule(lr){2-5}\cmidrule(lr){6-7}\cmidrule(lr){8-8}
Detector & \oracle{} & \last{} & \valnormal{} & \hpfree{} & \last{} & \valnormal{} & minus \hpfree{} \\
\midrule
SIGNET$^\ast$ & 82.60 & 65.93 & 71.44 & 73.04 & 16.67 & 11.16 & $-$1.61 \\
OCGTL & 77.84 & 59.45 & 62.72 & 64.54 & 18.39 & 15.12 & $-$1.82 \\
OCGIN & 79.40 & 75.15 & 74.15 & 74.02 & 4.25 & 5.25 & 0.13 \\
GLocalKD & 77.99 & 65.92 & 66.69 & 77.45 & 12.07 & 11.31 & $-$10.76 \\
GLADC & 76.81 & 75.47 & 70.39 & 71.71 & 1.34 & 6.41 & $-$1.32 \\
CVTGAD$^\dagger$ & 84.19 & 82.20 & 81.96 & 82.26 & 1.99 & 2.23 & $-$0.30 \\
GOOD-D & 84.54 & 80.64 & 80.35 & 81.23 & 3.90 & 4.19 & $-$0.88 \\
GraphDE$^\ddagger$ & 69.54 & 51.79 & 53.81 & 50.50 & 17.74 & 15.73 & 3.30 \\
GraphCL-IF & 78.79 & 72.03 & 74.02 & 72.02 & 6.76 & 4.78 & 2.00 \\
GraphCL-OCSVM & 81.27 & 68.46 & 73.02 & 68.81 & 12.81 & 8.25 & 4.22 \\
InfoGraph-IF & 80.51 & 67.91 & 72.47 & 75.18 & 12.60 & 8.04 & $-$2.71 \\
InfoGraph-OCSVM & 79.50 & 69.91 & 74.66 & 70.50 & 9.60 & 4.85 & 4.15 \\
\midrule
Mean of 12 & 79.42 & 69.57 & 71.31 & 71.77 & 9.84 & 8.11 & $-$0.47 \\
95\% CI & & & & & [6.73, 13.07] & [5.36, 11.00] & [$-$1.67, 0.83] \\
Detectors $\geq 2$ & & & & & 10 & 12 & 3 \\
\bottomrule
\end{tabular}

\end{table}

Against \last{}, OCGTL, GraphDE, and SIGNET have the largest gaps, at 16.67--18.39, while GLADC, CVTGAD, and GOOD-D have the smallest.
For GLADC, the gap is only 1.34 against \last{} but 6.41 against \valnormal{}, showing that the choice of comparison rule matters.
Unlike UB9, tuned configurations average 0.47 points below defaults under \valnormal{}, with an interval of [$-1.67$, 0.83].
Only three detectors gain at least 2 points from tuning.
GLocalKD scores 10.76 points higher with defaults.
The benefit of UB-GOLD hyperparameters depends on the scenario, whereas checkpoint leakage occurs in both.
With both checkpoint and hyperparameter choices changed, \oracle{} exceeds \hpfree{} by 7.64 [4.57, 11.42].

\paragraph{Noise and rankings.}
We compute the label-permutation check on tuned runs as in Appendix~\ref{app:posthoc-null}, using the run index fixed in A37.
For eight detectors, the null gain is less than half the observed \oracle{} minus \last{} gap, with ratios from 0.16 for GraphDE to 0.43 for SIGNET.
The ratios are 0.56 for GLADC, 0.80 for OCGIN, 1.90 for CVTGAD, and 1.20 for GOOD-D.
For the latter two, selecting on permuted labels can produce gains at least as large as those observed.
For most detectors, labels identify checkpoints that score clearly above the final checkpoint on these test graphs.
Because trials reuse the same anomaly graphs, this does not establish whether the selected checkpoints would improve on other graphs from the second dataset.

Leaderboard $\tau_b$ between \oracle{} and \valnormal{} is 0.58 [0.27, 0.76], below UB9's 0.76.
The point estimate therefore indicates a larger ranking change.
CVTGAD and GOOD-D lead every protocol, with SIGNET third under \oracle{} and GLADC, InfoGraph-OCSVM, and GLocalKD third under \last{}, \valnormal{}, and \hpfree{}, respectively.
The top-three set under \oracle{} differs from every label-free protocol.

\subsection{Pair variation and simple references}

Table~\ref{tab:t3-pairs} gives each pair's 12-detector mean under three protocols, the two leakage gaps, and two training-free references.
Leakage is positive on every pair, ranging from 5.81 on ENZYMES+PROTEINS to 14.56 on BZR+COX2 against \last{}.
For some dataset pairs, graph size alone distinguishes normal from anomalous graphs well.
On AIDS+DHFR, size alone reaches 89.11, close to the \valnormal{} detector mean of 90.77, with mean graph sizes of 15.7 and 42.4 nodes.
On ENZYMES+PROTEINS, size alone reaches 66.73, above every protocol's detector mean, including 61.09 under \oracle{}.

\begin{table}[!htb]
  \centering\scriptsize
  \setlength{\tabcolsep}{3pt}
  \caption{Type III results and references by dataset pair.}
  \label{tab:t3-pairs}
  \begin{tabular}{lccccccc}
\toprule
 & \multicolumn{3}{c}{Mean of 12 detectors} & \multicolumn{2}{c}{\oracle{} minus} & \multicolumn{2}{c}{References} \\
\cmidrule(lr){2-4}\cmidrule(lr){5-6}\cmidrule(lr){7-8}
Pair & \oracle{} & \last{} & \valnormal{} & \last{} & \valnormal{} & Size only & Graphlet count \\
\midrule
PTC\_MR + MUTAG & 79.12 & 65.35 & 67.42 & 13.77 & 11.70 & 48.46 & 50.90 \\
AIDS + DHFR & 96.68 & 87.97 & 90.77 & 8.71 & 5.91 & 89.11 & 83.79 \\
BZR + COX2 & 82.06 & 67.50 & 69.57 & 14.56 & 12.49 & 59.30 & 20.92 \\
ENZYMES + PROTEINS & 61.09 & 55.28 & 55.25 & 5.81 & 5.83 & 66.73 & 47.64 \\
IMDB-MULTI + IMDB-BINARY & 78.13 & 71.76 & 73.52 & 6.37 & 4.61 & 65.52 & 79.00 \\
\bottomrule
\end{tabular}

\end{table}

Graphlet counts reach 79.00 on IMDB-MULTI+IMDB-BINARY, above the \valnormal{} detector mean of 73.52, while both references remain near chance on PTC\_MR+MUTAG.
On BZR+COX2, graphlet counts score 20.92, ranking COX2 graphs as more typical of the normal training data than held-out BZR graphs.
Node features are constant, so detectors can use only graph structure.
Even so, size alone matches or exceeds the detector means on two of five pairs.
A high Type III score alone therefore does not establish sensitivity to more detailed graph structure.

\subsection{Reproduction of UB-GOLD's published values}

Part A compares \oracle{} under UB-GOLD's selection rule with the published Table 3 to check whether the replication roughly measures the same scenario.
This checks approximate agreement with published results and does not determine whether the new results are valid.
A detector passes if at least two of five pairs lie within 5 points of the published values.
Table~\ref{tab:t3-repro} places published scores alongside our five-trial means and counts the qualifying pairs.
GraphDE, marked with a pilcrow, uses UB-GOLD's score direction in this table to match the published results.
A section sign marks two AIDS+DHFR cells where UB-GOLD's Tables 3 and 8 disagree, and the final row gives the Table 8 values.

\begin{table}[!htb]
  \centering\scriptsize
  \setlength{\tabcolsep}{3pt}
  \caption{Reproduction of UB-GOLD's Type III values.}
  \label{tab:t3-repro}
  \begin{tabular}{@{}l@{\hspace{10pt}}c@{\hspace{6pt}}c@{\hspace{10pt}}c@{\hspace{6pt}}c@{\hspace{10pt}}c@{\hspace{6pt}}c@{\hspace{10pt}}c@{\hspace{6pt}}c@{\hspace{10pt}}c@{\hspace{6pt}}c@{\hspace{10pt}}c@{}}
\toprule
 & \multicolumn{2}{c}{PTC\_MR} & \multicolumn{2}{c}{AIDS} & \multicolumn{2}{c}{BZR} & \multicolumn{2}{c}{ENZYMES} & \multicolumn{2}{c}{IMDB-MULTI} & Pairs \\
 & \multicolumn{2}{c}{+ MUTAG} & \multicolumn{2}{c}{+ DHFR} & \multicolumn{2}{c}{+ COX2} & \multicolumn{2}{c}{+ PROTEINS} & \multicolumn{2}{c}{+ IMDB-BINARY} & within \\
\cmidrule(l{2pt}r{2pt}){2-3}\cmidrule(l{2pt}r{2pt}){4-5}\cmidrule(l{2pt}r{2pt}){6-7}\cmidrule(l{2pt}r{2pt}){8-9}\cmidrule(l{2pt}r{2pt}){10-11}
Detector & Pub. & Ours & Pub. & Ours & Pub. & Ours & Pub. & Ours & Pub. & Ours & 5 points \\
\midrule
SIGNET$^\ast$ & 84.63 & 83.58 & 96.82 & 97.26 & 89.11 & 87.90 & 62.42 & 61.73 & 71.10 & 82.52 & 4 \\
OCGTL & 62.51 & 75.23 & 98.95 & 99.19 & 81.86 & 80.90 & 67.18 & 67.56 & 66.73 & 66.31 & 4 \\
OCGIN & 79.27 & 74.78 & 95.68 & 95.96 & 87.27 & 81.39 & 61.77 & 64.29 & 80.98 & 80.55 & 4 \\
GLocalKD & 77.54 & 74.57 & 94.33 & 93.95 & 80.55 & 80.64 & 59.36 & 59.57 & 81.25 & 81.24 & 5 \\
GLADC & 80.12 & 82.20 & 95.42 & 97.43 & 83.21 & 77.82 & 56.95 & 49.81 & 78.28 & 76.77 & 3 \\
CVTGAD & 79.44 & 79.39 & 99.10 & 97.95 & 96.32 & 97.20 & 64.31 & 63.97 & 80.23 & 82.44 & 5 \\
GOOD-D & 82.05 & 81.12 & 99.27 & 99.36 & 95.16 & 96.31 & 63.84 & 64.74 & 80.94 & 81.18 & 5 \\
GraphDE$^\P$ & 58.28 & 32.69 & 94.58 & 93.49 & 65.26 & 56.34 & 54.48 & 48.04 & 52.67 & 28.74 & 1 \\
GraphCL-IF & 79.74 & 79.19 & 98.48 & 98.26 & 76.01 & 77.79 & 63.23 & 60.14 & 78.62 & 78.58 & 5 \\
GraphCL-OCSVM & 78.10 & 82.73 & 97.11 & 97.67 & 78.98 & 86.89 & 59.76 & 61.03 & 74.45 & 78.04 & 4 \\
InfoGraph-IF & 53.78 & 81.61 & 83.27$^\S$ & 94.45 & 64.65 & 87.44 & 33.52 & 60.16 & 51.32 & 78.90 & 0 \\
InfoGraph-OCSVM & 55.88 & 80.38 & 84.33$^\S$ & 96.57 & 64.29 & 82.17 & 46.01 & 59.66 & 56.26 & 78.74 & 0 \\
\midrule
\multicolumn{12}{l}{$^\S$UB-GOLD Table 8 instead: InfoGraph-IF 63.27, InfoGraph-OCSVM 44.33} \\
\bottomrule
\end{tabular}

\end{table}

Nine of 12 detectors pass the check.
Both InfoGraph detectors exceed the published values on all five pairs, consistent with UB-GOLD's mismatch between two-stage scores and graph labels, a pattern also seen on UB9 (Appendices~\ref{app:kernels-pairing} and \ref{app:ports-ubgold}).
GraphDE lies within 5 points only on AIDS+DHFR.
SIGNET's published values cannot come from its Type III lines as released, but correcting the flags gives agreement within 5 points on four pairs.
Other differences have both signs, with OCGTL at 75.23 versus the published 62.51 on PTC\_MR+MUTAG and GLADC at 49.81 versus 56.95 on ENZYMES+PROTEINS.
Whether UB-GOLD's PyTorch 2.1 runs used identical splits remains unproven and is still a possible source of differences.

R1 shows that checkpoint leakage replicates with a larger gap when anomalies come from another dataset.
These tests use a fixed anomaly block from the second dataset and constant node features.
Two pairs use anomalies from only one class, and size alone matches or exceeds the detector means on two pairs.
These properties limit conclusions about detecting anomalies from a different distribution.
They do not change the comparison of checkpoint rules on the same training runs.
Following A36, we evaluate neither the Type III OGB pairs nor recent methods in this scenario.

\section{Controlled structural anomalies}
\label{app:typed}

This appendix tests whether the detectors in Section~\ref{sec:f3} identify known changes in graph structure.
Normal and anomalous benchmark graphs may differ in several properties, so high AUROC alone does not show which change a detector identifies.
We use two generators to change motif composition in synthetic graphs and ring structure in real molecules.
Simple references check whether these changes can be detected.
Only the eight formal-test UB-GOLD detectors enter the formal typed-anomaly decision.
We call these the formal-test detectors below.
Results for the other detectors and all follow-up analyses are descriptive.

\subsection{Data checks and decision rules}
\label{app:typed-rules}

An anomaly type can test structural detection only if its changes are detectable without relying only on graph size or the label pairs on adjacent nodes.
We fit three reference methods on each seed's normal training graphs, without gradient training.
Each uses Mahalanobis distance with the Ledoit--Wolf covariance estimate~\citep{ledoitwolf}.
\emph{Graphlet count + LW} uses the log counts of the 29 connected graphlets with three to five nodes and the log node and edge counts.
\emph{Size only} uses the log node count, log edge count and density.
\emph{Edge-label pair + LW} uses the log counts of unordered node-label pairs joined by an edge.

We assess each anomaly type using mean AUROC over five seeds.
A valid type must give graphlet count + LW at least 85 and size only at most 60.
If edge-label pair + LW reaches at least 85, we exclude the type from the decision because adjacent node-label pairs alone can distinguish its anomalies.
We apply these checks without using deep-detector scores to select test graphs.

Each detector uses the seed's normal training graphs, holding out 20\% for \valnormal{}.
Its configuration is the most common UB-GOLD configuration for that detector across the TU datasets, fixed before any typed run.
We score each type's anomalies together with the test normals as a separate test set.
This is needed because some detectors' scores depend on the other graphs in the test batch.
A detector is \emph{blind} to a type if its mean \valnormal{} AUROC over five seeds is at most 60.

The formal decision tests whether many detectors miss changes from both generators, rather than only a few detector--type pairs.
Amendment A9 fixed three required conditions before the formal experiments.
At least 75\% of detector-type pairs across valid size-preserving types from the synthetic generator GX and molecular generator MS-v2 must be blind.
Within MS-v2, at least 60\% of pairs must be blind, and at least five of the eight detectors must be blind on most valid types.
The blind-spot hypothesis is refuted if fewer than 40\% of MS-v2 pairs are blind, and the outcome is inconclusive if fewer than two MS-v2 types are valid.
We also compare size-preserving and size-changing versions of the same molecular edits to determine the claim's scope.
If changing size reduces the blind rate by at least 25 percentage points, the claim is restricted to size-preserving changes.

The hypothesis arose from pilot GX results for SIGNET, OCGTL, OCGIN and GLocalKD, so these four detectors are excluded from the formal decision.
When A9 was written, the other eight UB-GOLD detectors had not been run on any typed data, and the molecular generator did not yet exist.
We also report the four pilot detectors and nine recent methods to show the broader pattern without including them in the formal decision.
Appendix~\ref{app:prereg} records the criteria and their timing.

\subsection{Motif changes in synthetic graphs}
\label{app:typed-gx}

GX directly controls the types and proportions of motifs in a graph.
Each graph contains a preferential-attachment tree with 20 to 30 nodes and ten motifs, each connected by one edge to a different tree node.
Two motifs are stars with four leaves, and the other eight have five nodes and six edges, taking the house or bowtie form with equal probability in normal graphs.
A graph with $n_b$ backbone nodes therefore has $n_b+50$ nodes and $n_b+65$ edges.
All anomaly types share the normal size distribution, without matching each anomaly to an individual normal graph of exactly the same size.

T1a replaces two ring motifs with $K_{2,3}$, which never appears in normal graphs but also has five nodes and six edges.
T3a increases the house proportion among ring motifs to 0.9, while T3b reduces it to 0.1.
Node labels encode roles, with 5\% redrawn at random, and each node has a two-dimensional Gaussian attribute whose mean depends on its role.
The primary feature variant F2 concatenates the one-hot label and the two attributes, while the sensitivity variant F1 uses only the label.
Each of seeds 0 to 4 contains 500 training normals, including 100 held out for validation, plus 250 test normals and 250 anomalies per type.
GX was written for an earlier experiment and frozen before any GX result was observed.

\subsection{Ring changes in real molecules}

\paragraph{Design objective.}
\label{app:typed-msv1}
The formal decision requires a second generator, independent of GX and based on real molecules.
The initial MolSwap v1 (MS-v1) used the 1637 connected NCI1 molecules in UB-GOLD's normal class.
Its three local edits preserved node count, edge count and every node's label.
They converted a six-membered ring into a five-membered ring with a pendant atom, opened a ring and closed another in an acyclic region, or rewired a four-atom chain into a star.
The generator and data were frozen and hashed before any control or detector run, as recorded in A9.1.

The three size-preserving MS-v1 types scored only 53.5, 77.8 and 42.4 under graphlet count + LW, so none passed the validity check.
We therefore stopped using MS-v1 before any full deep-detector run, as recorded in A28.
The only deep-detector runs were short smoke tests whose test metrics were never inspected, so MS-v1 provides no evidence about detector capability.
The control scores indicated that a single local edit was too small compared with variation among normal molecules.

\paragraph{MS-v2 design and selection.}
\label{app:typed-msv2-design}
Before MS-v2 was developed, A28 restricted design decisions to control results.
The designer could not access deep-detector results on the typed data.
Each design rule had to be recorded before its corresponding scores were computed.
Each candidate applies one family of edge swaps to every parent molecule, with at most $k$ swaps per molecule.
Every candidate preserves node count, edge count and the node-label array.
Every node's label-degree pair must already occur in a normal NCI1 molecule.
An anomaly cannot be identical to its parent or any normal molecule.

Before scoring candidates, A25 fixed the following requirements.
Candidates had to pass the validity checks above, score below 85 under edge-label pair + LW and at most 60 on unedited parents under graphlet count + LW, and generate at least 100 anomalies per seed.
Among qualifying candidates, the rule preferred the smallest $k$.
Two iterations tested 26 candidates from six families.
The retained MS1 shrinks a six-membered ring into a four-membered ring and a two-atom chain, MS2 relocates one or two rings into acyclic regions, and MS3 shrinks a six-membered ring into a three-membered ring and a chain.
Only 3.6\% and 6.2\% of normal molecules contain four- and three-membered rings, respectively.
Candidates changing common five- and six-membered ring proportions or converting fused junctions into spiro junctions never reached 85, with a best value of 84.8, so proportion shifts are tested only on GX.
MS-v2 code, data and control outputs were frozen and hashed before full formal deep-detector runs on GX or MS-v2, as recorded in A28.1.

\paragraph{Validity and independent checks.}
\label{app:typed-msv2-valid}
The final three rows of Table~\ref{tab:typed-full} report the controls.
All six size-preserving types pass the validity checks, and none reaches the exclusion threshold for adjacent node-label pairs.
MS2 is closest to the validity threshold, with a five-seed mean of 86.4 and a lowest seed of 84.4.
Every type is judged by its five-seed mean.
Unedited parents score 46.2 to 53.1 under graphlet count + LW.
Parent selection alone is therefore not detected by this reference.
Each MS-v2 seed has 327 test normals, with 139 to 173 anomalies for MS2 and 237 to 307 for MS1 and MS3.

An independent script recomputed all properties from the saved files without calling a detector, and every anomaly passed.
Every anomaly is simple, connected and non-isomorphic to any normal molecule.
The size-preserving variant retains its parent's node count, edge count and node-label array, and the checker confirms the intended edit by replaying the recorded swaps.
The size-changing variant joins the size-preserving anomaly to an unchanged donor molecule by one edge.
In negative controls, no type's checker accepted another type's anomalies, and only one actual ring relocation was accepted among 400 random edge swaps of normal molecules.
Two independent regenerations produced byte-identical files.

\subsection{Detection results and scope}
\label{app:typed-results}

Table~\ref{tab:typed-full} reports five-seed mean \valnormal{} AUROC for every detector, type and variant, with bold values exceeding 60 and therefore not classified as blind.
For the eight formal-test detectors, only the size-preserving columns enter the main decision.
The size-changing columns determine whether the finding is restricted to size-preserving changes.
Rows for pilot detectors and recent methods are descriptive.

\begin{table}[!htb]
  \centering\footnotesize
  \setlength{\tabcolsep}{4pt}
  \caption{Detection of typed anomalies.}
  \label{tab:typed-full}
  \begin{tabular}{l*{9}{c}}
\toprule
& \multicolumn{3}{c}{GX, size-preserving} & \multicolumn{3}{c}{MS-v2, size-preserving} & \multicolumn{3}{c}{MS-v2, size-changing} \\
\cmidrule(lr){2-4}\cmidrule(lr){5-7}\cmidrule(lr){8-10}
Detector & T1a & T3a & T3b & MS1 & MS2 & MS3 & MS1 & MS2 & MS3 \\
\midrule
\multicolumn{10}{l}{\emph{Formal-test detectors (enter the decision)}} \\
GLADC & 51.1 & 50.5 & 54.6 & 53.4 & 56.0 & 52.6 & \textbf{85.7} & \textbf{87.0} & \textbf{85.7} \\
CVTGAD & \textbf{69.9} & 53.2 & 56.3 & \textbf{68.9} & \textbf{82.0} & \textbf{76.9} & \textbf{85.0} & \textbf{88.5} & \textbf{87.4} \\
GOOD-D & \textbf{92.7} & 57.0 & \textbf{60.7} & \textbf{69.3} & \textbf{75.8} & \textbf{69.9} & \textbf{85.3} & \textbf{86.6} & \textbf{84.8} \\
GraphDE & 49.9 & 48.6 & 50.0 & 51.7 & 44.5 & 50.0 & 48.0 & 43.5 & 46.1 \\
GraphCL-IF & 50.3 & 50.1 & 49.0 & 52.8 & 59.1 & 52.6 & \textbf{77.9} & \textbf{80.9} & \textbf{77.8} \\
GraphCL-OCSVM & 51.1 & 50.6 & 49.6 & 51.6 & 57.2 & 50.6 & \textbf{74.8} & \textbf{78.0} & \textbf{74.7} \\
InfoGraph-IF & 47.2 & 48.5 & 47.2 & 51.0 & 58.2 & 51.1 & \textbf{86.1} & \textbf{88.1} & \textbf{85.6} \\
InfoGraph-OCSVM & 49.4 & 49.6 & 48.4 & 49.8 & 57.1 & 50.6 & \textbf{84.9} & \textbf{87.3} & \textbf{84.6} \\
\midrule
\multicolumn{10}{l}{\emph{Pilot detectors used to form the hypothesis, descriptive results}} \\
SIGNET & 49.8 & 50.5 & 52.9 & 54.9 & \textbf{62.8} & 53.4 & \textbf{75.0} & \textbf{70.2} & \textbf{75.7} \\
OCGTL & 52.8 & 54.0 & 54.8 & \textbf{60.4} & \textbf{68.3} & \textbf{63.0} & \textbf{89.6} & \textbf{91.9} & \textbf{90.2} \\
OCGIN & 48.1 & 52.9 & 51.6 & 50.1 & \textbf{61.8} & 51.0 & \textbf{77.7} & \textbf{82.9} & \textbf{78.1} \\
GLocalKD & 50.1 & 51.5 & 51.6 & 50.3 & 58.8 & 50.8 & \textbf{83.4} & \textbf{87.0} & \textbf{83.8} \\
\midrule
\multicolumn{10}{l}{\emph{Recent methods (descriptive)}} \\
MUSE & 49.5 & 51.3 & 50.2 & 49.1 & 49.9 & 48.7 & 50.1 & 55.6 & 49.7 \\
LGKDE & \textbf{67.0} & \textbf{64.2} & \textbf{68.6} & 54.1 & 56.1 & 53.6 & 48.3 & 49.0 & 48.5 \\
GLADPro & 49.7 & 47.4 & 52.0 & 49.4 & 49.7 & 49.2 & 25.0 & 24.0 & 24.8 \\
GLADMamba & \textbf{78.8} & 54.8 & 53.3 & \textbf{60.6} & \textbf{66.5} & \textbf{61.5} & \textbf{82.6} & \textbf{85.1} & \textbf{82.2} \\
SEGO & \textbf{95.2} & \textbf{63.6} & \textbf{65.2} & \textbf{69.2} & \textbf{79.8} & \textbf{75.4} & \textbf{82.6} & \textbf{86.1} & \textbf{85.3} \\
GLASS & \textbf{82.7} & \textbf{78.8} & \textbf{61.0} & \textbf{68.8} & \textbf{70.2} & \textbf{83.0} & \textbf{87.0} & \textbf{87.9} & \textbf{90.6} \\
ProtoGLAD & 51.3 & 48.7 & 55.6 & 49.7 & 48.0 & 49.0 & 46.8 & 53.0 & 47.9 \\
HC-GLAD & 48.8 & 54.7 & 43.5 & 50.7 & 57.5 & 52.4 & \textbf{82.5} & \textbf{84.3} & \textbf{83.4} \\
SUL/GUL & 49.8 & 50.6 & 49.6 & 49.7 & 50.1 & 50.2 & 50.6 & 50.4 & 50.6 \\
\midrule
\multicolumn{10}{l}{\emph{Training-free controls used to determine type validity}} \\
Graphlet count + LW & 100.0 & 90.1 & 91.2 & 95.1 & 86.4 & 94.2 & 95.5 & 88.3 & 95.1 \\
Size only & 51.0 & 49.1 & 51.0 & 48.8 & 51.7 & 49.3 & 67.4 & 72.2 & 68.0 \\
Edge-label pair + LW & 51.2 & 50.2 & 50.7 & 52.3 & 59.0 & 51.2 & 76.6 & 81.1 & 76.3 \\
\bottomrule
\end{tabular}

\end{table}

The eight formal-test detectors are blind in 81\% of the 48 pairs across both generators and 75\% of the 24 MS-v2 pairs.
Six detectors are blind on most MS-v2 types.
All three specified conditions are therefore met.
With GX F1 features, the blind share is 77\% and the decision is unchanged.
CVTGAD and GOOD-D are important exceptions, detecting T1a and all three MS-v2 types while remaining blind or close to the threshold on the proportion shifts T3a and T3b.

The two exceptions also share a relevant implementation feature, a separate structural encoder that uses precomputed random-walk and degree encodings.
The loader at c940457 attaches these encodings to every graph's nodes in \path{dataloader/data_loader.py}, lines 126--147.
Only CVTGAD and GOOD-D pass them to a separate structural encoder, in \path{GAOOD/nn/cvtgad.py}, lines 16--23 and 64--67, and \path{GAOOD/nn/good_d.py}, lines 41--42 and 68--72.
Appendix~\ref{app:extra-ablation} tests whether these encodings help detect the changes by removing them in an ablation.

The pilot detectors show a weaker pattern, with 79\% blind pairs across both generators but only 58\% on MS-v2.
OCGTL detects all three molecular types, while SIGNET and OCGIN detect MS2.
The pilot and formal-test results differ, supporting the use of a new generator and detectors not used to form the hypothesis.

The size-changing variants reduce the formal-test MS-v2 blind rate from 75\% to 12.5\%, a drop of 62.5 percentage points exceeding the required 25.
The claim is therefore restricted to size-preserving changes.
Every formal-test detector except GraphDE detects all three size-changing types.
Their misses on size-preserving variants therefore do not mean that they cannot learn from these molecules at all.
GraphDE is blind in every cell, consistent with its near-chance UB9 performance.
The size-changing variants are not valid composition tests because size only already scores 67.4 to 72.2.

Among recent methods, SEGO and GLASS detect all six size-preserving types, GLADMamba follows CVTGAD and GOOD-D's pattern, and LGKDE detects only the GX types.
The other five, MUSE, GLADPro, ProtoGLAD, HC-GLAD and SUL/GUL, score 43.5 to 57.5 on every size-preserving type.
HC-GLAD detects the size-changing variants, as most UB-GOLD detectors do, while LGKDE, MUSE, ProtoGLAD and SUL/GUL remain near chance.
GLADPro scores 24.0 to 25.0, ranking larger anomalous molecules as more normal.
LGKDE, MUSE, GLADPro, ProtoGLAD and SUL/GUL therefore miss MS-v2 anomalies even when size changes.
Their low scores alone therefore do not show a failure specific to detecting structural changes.

\subsection{Does checkpoint selection explain the misses?}
\label{app:typed-ckpt}

Scores may use different scales at different checkpoints.
\valnormal{} may therefore select a checkpoint with a low mean normal score but poor separation of normal and anomalous graphs.
We recompute \last{} and \oracle{} results from the stored runs to check whether the misses depend on the selection rule.
This post hoc descriptive analysis is part D of the follow-up plan, adopted in A36.
At that point, a file-format check had displayed only one typed \last{} value, for CVTGAD on GX F1, T1a and seed 0.
No other typed \last{} or \oracle{} result had been read at that time.
The plan interprets blindness under \oracle{} as blindness at every candidate checkpoint, and blindness under \valnormal{} that disappears under another rule as dependent on checkpoint selection.

Table~\ref{tab:typed-rules} counts blind formal-test detectors for each type and variant.
The counts are identical under \last{} and \valnormal{} for every type and variant.
\oracle{} changes counts on only three types.
MS2 falls from six to two because the four two-stage detectors rise from 57.1 to 59.1 under \valnormal{} to 60.0 to 61.8 under \oracle{}.
T3a and T3b each fall by one because CVTGAD rises from 53.2 and 56.3 to 63.5 and 65.5, respectively.
The other formal-test detectors retain the same blind or non-blind classification under all three rules.

\begin{table}[!htb]
  \centering\footnotesize
  \caption{Blind calls under three checkpoint rules.}
  \label{tab:typed-rules}
  \begin{tabular}{l*{9}{c}}
\toprule
& \multicolumn{3}{c}{GX, size-preserving} & \multicolumn{3}{c}{MS-v2, size-preserving} & \multicolumn{3}{c}{MS-v2, size-changing} \\
\cmidrule(lr){2-4}\cmidrule(lr){5-7}\cmidrule(lr){8-10}
Checkpoint rule & T1a & T3a & T3b & MS1 & MS2 & MS3 & MS1 & MS2 & MS3 \\
\midrule
\valnormal{} & 6 & 8 & 7 & 6 & 6 & 6 & 1 & 1 & 1 \\
\last{} & 6 & 8 & 7 & 6 & 6 & 6 & 1 & 1 & 1 \\
\oracle{} & 6 & 7 & 6 & 6 & 2 & 6 & 1 & 1 & 1 \\
\bottomrule
\end{tabular}

\end{table}

For the two-stage detectors on MS2 and CVTGAD on proportion shifts, the blind classification depends on checkpoint selection.
However, selecting checkpoints with test labels can raise reported performance.
MS2 exceeds the threshold by at most 1.8 points, whereas selecting and reporting on the same UB9 graphs adds 3.64 points over disjoint labeled selection in Appendix~\ref{app:followup-split}.
CVTGAD's margins are comparable to its own 5.06-point optimism in that analysis.
On UB9, its gain from selection after label permutation also exceeds its gain with the original labels (Appendix~\ref{app:posthoc-null}).
Recent-method changes are larger, with GLADPro scoring 49.2 to 49.7 under \valnormal{} but 62.6 to 63.5 under \oracle{} on size-preserving MS-v2.

\subsection{Structural statistics that detect the edits}
\label{app:typed-lowctrl}

The graphlet control shows that the edits can be detected from graph structure.
It does not show whether detection depends on node degrees or ring composition.
We add three simpler references to check whether node degrees alone can detect the MS-v2 edits missed by the detectors.
They are fitted by seed on the same normal training graphs as the validity controls and use the same Ledoit--Wolf distance.
This is part H of the follow-up plan, specified in A42 before its scores were computed and supplementing Section~\ref{sec:f3} and Figure~\ref{fig:typed}a.

The \emph{degree histogram} counts nodes of degrees 0 to 9 and at least 10, while the \emph{labeled-degree histogram} counts node-label and degree pairs with degrees capped at 10.
\emph{Ring counts} count cycles of lengths 3 to 8 and at least 9 in a minimum cycle basis, and all counts use $\log(1+x)$.
A42 specified the same covariance regularization as the existing controls, 0.01 for log-count features and $10^{-6}$ for size features.
The code saved before execution treated all three references as log counts and used 0.01 throughout.
A43 approved this choice only after the results had been computed and read.
As an implementation check, the same code reproduced all 135 official control rows within $1.4\times10^{-14}$ points.

Table~\ref{tab:typed-lowctrl} reports each reference and validity control, with five-seed mean AUROC and the lowest and highest seed scores.
On size-preserving MS-v2, the degree histogram scores only 49.3 to 51.3 with no seed above 53.6, the labeled-degree histogram scores 52.1 to 58.2, and ring counts score 85.8 to 94.7.
These edits therefore change ring composition while leaving degrees nearly unchanged, so missing them includes a failure to detect ring-level changes.

\begin{table}[!htb]
  \centering\footnotesize
  \setlength{\tabcolsep}{4pt}
  \caption{Low-order references on typed anomalies.}
  \label{tab:typed-lowctrl}
  \begin{tabular}{l*{9}{c}}
\toprule
& \multicolumn{3}{c}{GX, size-preserving} & \multicolumn{3}{c}{MS-v2, size-preserving} & \multicolumn{3}{c}{MS-v2, size-changing} \\
\cmidrule(lr){2-4}\cmidrule(lr){5-7}\cmidrule(lr){8-10}
Reference & T1a & T3a & T3b & MS1 & MS2 & MS3 & MS1 & MS2 & MS3 \\
\midrule
Degree histogram + LW & 53.9 & 81.9 & 75.9 & 49.3 & 51.3 & 51.0 & 61.0 & 66.6 & 62.5 \\
\quad lowest seed & 48.4 & 80.6 & 69.3 & 47.9 & 48.6 & 50.4 & 59.6 & 63.2 & 61.3 \\
\quad highest seed & 56.3 & 83.0 & 79.0 & 52.2 & 53.6 & 51.6 & 62.7 & 68.7 & 64.1 \\
\addlinespace
Labeled-degree histogram + LW & 51.7 & 75.8 & 65.9 & 52.9 & 58.2 & 52.1 & 76.1 & 79.6 & 76.7 \\
\quad lowest seed & 50.0 & 72.9 & 63.3 & 51.3 & 56.8 & 50.5 & 74.3 & 77.5 & 74.5 \\
\quad highest seed & 54.2 & 78.4 & 68.7 & 54.3 & 58.9 & 53.4 & 78.0 & 81.7 & 78.3 \\
\addlinespace
Ring counts + LW & 88.9 & 91.7 & 95.3 & 94.7 & 86.0 & 85.8 & 95.7 & 87.9 & 88.4 \\
\quad lowest seed & 87.2 & 90.7 & 94.6 & 93.8 & 84.9 & 84.4 & 95.0 & 86.7 & 87.3 \\
\quad highest seed & 89.6 & 93.4 & 95.9 & 96.4 & 88.5 & 87.3 & 97.2 & 90.0 & 89.8 \\
\midrule
Graphlet count + LW & 100.0 & 90.1 & 91.2 & 95.1 & 86.4 & 94.2 & 95.5 & 88.3 & 95.1 \\
Size only & 51.0 & 49.1 & 51.0 & 48.8 & 51.7 & 49.3 & 67.4 & 72.2 & 68.0 \\
Edge-label pair + LW & 51.2 & 50.2 & 50.7 & 52.3 & 59.0 & 51.2 & 76.6 & 81.1 & 76.3 \\
\bottomrule
\end{tabular}

\end{table}

On GX, the degree histogram partly detects the T3a and T3b proportion shifts at 81.9 and 75.9, respectively, but scores only 53.9 on the new T1a motif.
A house and $K_{2,3}$ share a degree sequence, while a bowtie contains a degree-4 node.
T3a and T3b therefore test both degree and composition, while T1a tests composition changes.
Ring counts detect all three GX types.
These reference results are descriptive and do not change the validity controls or the original decision.

\subsection{Sensitivity to the blindness threshold}
\label{app:typed-threshold}
\label{app:typed-curve}

We use 60 to define blindness in this study.
It is not a universal cutoff for detection ability.
Table~\ref{tab:typed-threshold} varies the threshold from 50 to 70 in steps of 2.5 and reports blind shares among all 48 formal-test pairs and the 24 MS-v2 pairs.
It also reports the number of detectors blind on most MS-v2 types and whether all three decision conditions hold.
Figure~\ref{fig:typed}b plots both shares and marks the required proportions at 60.
Lines connect the reported thresholds for readability and do not give exact results at intermediate thresholds.

This sensitivity analysis is post hoc and descriptive.
The threshold-55 case was computed during a first-draft check, before A38 recorded the full table and A48 added the curve.
All three conditions hold at every threshold from 57.5 to 70 and fail at 55 and below because the MS-v2 blind share falls below 60\%.
Most formal-test detectors score near chance on many size-preserving types, and this pattern does not depend on one threshold.
Six detectors remain blind on most MS-v2 types from 55 to 67.5.
Table~\ref{tab:typed-rules} also shows identical per-type counts for \last{} and \valnormal{} at 60.

\begin{table}[!htb]
  \centering\footnotesize
  \setlength{\tabcolsep}{4pt}
  \caption{The typed decision at other thresholds.}
  \label{tab:typed-threshold}
  \begin{tabular}{ccccc}
\toprule
Blindness threshold & Blind pairs (of 48) & Blind MS-v2 pairs (of 24) & Detectors blind on most MS-v2 types & All three conditions \\
\midrule
50 & 27.1\% & 12.5\% & 1 & not met \\
52.5 & 54.2\% & 37.5\% & 4 & not met \\
55 & 66.7\% & 54.2\% & 6 & not met \\
57.5 & 77.1\% & 66.7\% & 6 & met \\
60 (fixed in advance) & 81.2\% & 75.0\% & 6 & met \\
62.5 & 83.3\% & 75.0\% & 6 & met \\
65 & 83.3\% & 75.0\% & 6 & met \\
67.5 & 83.3\% & 75.0\% & 6 & met \\
70 & 91.7\% & 87.5\% & 7 & met \\
\bottomrule
\end{tabular}

\end{table}

\subsection{The role of precomputed structural encodings}
\label{app:extra-ablation}

CVTGAD and GOOD-D are the only formal-test detectors to detect MS-v2 edits.
They are also the only UB-GOLD detectors to pass precomputed random-walk and degree encodings to a separate structural encoder.
We replace these encodings with a constant input of ones, $x_s=1$, during training and scoring to test whether they help detect the edits.
The structural encoder still passes messages along edges and can recover degrees and graph size.
This change removes the precomputed encodings but retains access to graph structure.
Both detectors are rerun with five seeds on GX F2 and MS-v2 and five folds on UB9.
This is a post hoc descriptive analysis.
A33 specified the ablation, and A34 changed the planned zero input to ones, both before any run.
UB-GOLD initializes biases to zero, so all-zero inputs cause division by zero in both losses.
It already uses ones for graphs without node features.

Table~\ref{tab:extra-ablation-typed} compares five-seed mean \valnormal{} AUROC with and without precomputed encodings, with values above 60 in bold.
With constant inputs, both detectors become blind on all six size-preserving types at 48.7 to 59.2, while retaining 78.8 to 84.0 on size-changing types.
MS-v2 scores fall by 16.6 to 27.1 points, and GOOD-D falls from 92.7 to 50.4 on T1a.
In the tested configurations, these results indicate that both detectors rely on the precomputed encodings to identify size-preserving changes.
This is consistent with random-walk return probabilities depending on short cycles.

\begin{table}[!htb]
  \centering\footnotesize
  \setlength{\tabcolsep}{4pt}
  \caption{Typed detection without precomputed encodings.}
  \label{tab:extra-ablation-typed}
  \begin{tabular}{llccc@{\hspace{8pt}}ccc@{\hspace{8pt}}ccc}
\toprule
 & & \multicolumn{3}{c}{GX, size kept} & \multicolumn{3}{c}{MS-v2, size kept} & \multicolumn{3}{c}{MS-v2, size changed} \\
\cmidrule(lr){3-5}\cmidrule(lr){6-8}\cmidrule(lr){9-11}
Detector & Input & T1a & T3a & T3b & MS1 & MS2 & MS3 & MS1 & MS2 & MS3 \\
\midrule
CVTGAD & encodings & \textbf{69.9} & 53.2 & 56.3 & \textbf{68.9} & \textbf{82.0} & \textbf{76.9} & \textbf{85.0} & \textbf{88.5} & \textbf{87.4} \\
 & $x_s = 1$ & 56.6 & 55.9 & 52.1 & 48.7 & 59.2 & 49.8 & \textbf{78.8} & \textbf{83.4} & \textbf{79.5} \\
\midrule
GOOD-D & encodings & \textbf{92.7} & 57.0 & \textbf{60.7} & \textbf{69.3} & \textbf{75.8} & \textbf{69.9} & \textbf{85.3} & \textbf{86.6} & \textbf{84.8} \\
 & $x_s = 1$ & 50.4 & 55.1 & 50.3 & 52.7 & 54.7 & 49.1 & \textbf{82.1} & \textbf{84.0} & \textbf{81.0} \\
\bottomrule
\end{tabular}

\end{table}

On UB9, \valnormal{} changes by $-5.43$ points for CVTGAD, with dataset-bootstrap interval $[-13.09,1.03]$, and by $-2.67$ for GOOD-D, with interval $[-5.03,-0.03]$.
Effects differ by dataset, with CVTGAD losing 30.5 points on NCI1 but gaining 10.9 on PROTEINS\_full.
Checkpoint leakage relative to \valnormal{} changes from 5.02 to 3.60 points for CVTGAD and from 4.65 to 5.64 for GOOD-D.
Replacing the encodings clearly changes the typed-test results, but its effect on benchmark AUROC differs across detectors and datasets.

\subsection{Approximate size matching on benchmark test sets}
\label{app:posthoc-size}

Typed anomalies test known size-preserving changes, while size matching asks how much detection performance remains on UB9 when normal and anomalous graphs have similar sizes.
This post hoc analysis, specified in A32, supplements Section~\ref{sec:c5} using stored per-graph detector scores without retraining or rescoring.
Within each fold, anomalies are processed in test-index order and paired with the nearest unmatched normal in $(\log n,\log(m+1))$.
Both features are standardized within the fold, and only pairs at distance at most 0.2 are accepted, giving approximate rather than exact node- and edge-count matches.

We retain the \valnormal{} and \oracle{} checkpoints used in the original full-test evaluation.
We then recompute AUROC on the matched subset from the stored scores, without selecting checkpoints again.
For detectors whose scores depend on the test batch, such as CVTGAD and GOOD-D (U23), the stored scores were computed with the full test set as one batch, so unmatched test graphs still enter the scores of matched graphs through the graph-level normalization, without using labels.
A dataset is reported only if at least 30\% of anomalies are matched in every fold.
As a matching check, size only is refitted on each fold's normal training graphs and should approach 50 when size no longer separates the classes.

Six datasets meet the 30\% rule, DHFR, ENZYMES, NCI1, PROTEINS\_full, IMDB-B and REDDIT-B.
AIDS has almost no class overlap in size, while BZR and COX2 have a minority of normal test graphs under UB-GOLD's anomaly definition, preventing all three from qualifying.
Across the six retained datasets, mean absolute standardized class differences fall from 0.61 to 0.01 for $\log n$ and from 0.65 to 0.02 for $\log(m+1)$.
Size only scores 47.4 to 51.8 on five datasets but 66.7 on REDDIT-B.
The distance limit leaves size differences on REDDIT-B that the reference can detect by scoring both unusually large and unusually small graphs highly.

Averaged across the six datasets, the eight formal-test UB-GOLD detectors with per-graph scores fall from 61.1 to 53.8 at the \valnormal{} checkpoint and from 65.5 to 56.2 at the \oracle{} checkpoint.
The eight recent methods with checkpoints fall from 58.3 to 54.7 and from 66.1 to 58.7, respectively.
At the \valnormal{} checkpoint, every detector except GLADPro loses AUROC, but the dataset effects differ.
The UB-GOLD mean falls from 80.0 to 58.9 on REDDIT-B and from 66.5 to 48.4 on PROTEINS\_full, but rises from 45.6 to 56.4 on NCI1.
CVTGAD, GOOD-D and GLASS retain 62.1, 59.2 and 61.4 on the matched subsets, compared with 53.2 for size only.

These results support a contribution of size to benchmark AUROC without establishing that detectors learn only size.
The matched subsets are not random test-set samples, so the changes are descriptive.
Because \oracle{} selects on the full test set, the smaller \oracle{}--\valnormal{} gap on matched subsets is not an independent checkpoint-leakage estimate.

\section{Statistical references and kernel baselines}
\label{app:kernels}

This appendix defines the statistical references and evaluates UB-GOLD's kernel baselines on the same folds.
The references show how well simple statistics predict benchmark labels.
The typed-anomaly experiments separately test whether detectors identify the constructed structural changes.

\subsection{Simple statistical references}
\label{app:kernels-refs}

Each reference fits only the fold's normal training graphs, without anomaly labels or tuning.
Writing $n$ for node count and $m$ for edge count, size only uses $\log n$, $\log(m+1)$ and density.
Graphlet count uses $\log(1+n)$, $\log(1+m)$ and log-transformed connected-graphlet counts \citep{graphlet}.
Both use Mahalanobis distance from the normal training feature mean, with the Ledoit--Wolf covariance estimate~\citep{ledoitwolf}.
Both unusually large and unusually small feature values can therefore receive high anomaly scores.
All counts are exact, using enumeration of connected induced subgraphs with three to five nodes for molecules and proteins and closed-form three- and four-node counts for social datasets.
REDDIT-B hubs make five-node enumeration too slow, so these limits were fixed by dataset family without labels.

Table~\ref{tab:extra-leaderboard} also includes isolation-forest scoring of the same counts and WL histogram + IF.
The latter uses a normalized Weisfeiler--Lehman subtree kernel with three iterations to compute features against at most 1000 training graphs, followed by isolation-forest scoring.
High scores from a simple reference show that its features help distinguish benchmark anomalies.
They do not by themselves show that deep detectors cannot detect structural changes.

\subsection{The fixed reference Combined statistics}
\label{app:grail0}

Combined statistics has the highest mean UB9 AUROC among the references that require no gradient training, at 62.1, compared with 60.4 for graphlet count and 57.3 for size only.
It was developed in an earlier project that ended after failing its criteria for further study, including the requirement to clearly outperform simple count references.
We use it as a fixed reference, without labels, gradient training or per-dataset tuning, and do not claim a new method.
Its runs on UB-GOLD's folds preceded the initial written plan for this study.
That plan already listed it as a reference.
The supplementary study record and the released code call this reference GRAIL-0.

Its score is the sum of four components fitted on normal training graphs.
The size component is identical to size only.
The composition component regresses each of 29 log graphlet counts on $\log n$ and $\log(m+1)$, then measures residual Mahalanobis distance with a Ledoit--Wolf covariance and fixed ridge $0.01I$.
For the node-role component, graphlet orbits distinguish a node's position within a graphlet.
We fit an eight-component diagonal Gaussian mixture to each node's log counts over the 73 orbits.
Its graph score is the mean node negative log-likelihood plus the mean over the highest-scoring 10\% of nodes.
Where node labels exist, the label component uses a categorical distribution with a Dirichlet prior to model label frequencies within each role.
It combines node scores in the same way.

For each component, we fit on half the normal training graphs and calibrate its scores on the other half using the median and 1.4826 times the median absolute deviation.
The halves are exchanged, the two z-scores averaged, and the four calibrated components summed.
All constants are identical across datasets and were fixed before any test result.
Five-fold smoke tests before the main run changed a WL control and added two combination rules from the earlier project, without changing the score, constants or summation rule of Combined statistics.

The reference has two known limitations.
First, the mixture's $k$-means initialization depends on training-node order.
An audit in the earlier project found fold-AUROC changes of about 2--4 points and dataset-mean changes of about 0.5 at most.
Second, a component with similar calibration scores can dominate the sum if it gives a few graphs very high scores, as the label component does on MUTAG and BZR.

\subsection{Kernel fitting and tuning}

UB-GOLD's four kernel detectors combine WL~\citep{wlkernel} or propagation kernels~\citep{pkernel} with IF~\citep{iforest} or OCSVM~\citep{ocsvm}, using normalized kernel matrices as outlier-detector inputs.
We use KernelGLAD at c940457 with GraKeL 0.1.10 on UB9 and four secondary datasets under default and tuned settings.
UB-GOLD releases no tuned kernel configurations, and its search script fails on import.
We therefore follow its declared search space and select the configuration with the highest five-fold mean test AUROC.
This is a test-selected upper bound rather than label-free tuning.
Kernel detectors have no checkpoints, so the tuned value fills all three checkpoint columns and the default value fills \hpfree{}.

The released implementation constructs the kernel and fits IF or OCSVM only on the test loader, including anomalies, and ignores the training loader.
To use the same training data as the other detectors, our leaderboard instead fits the kernel and outlier detector on normal training graphs and scores test graphs against them.
Test-set fitting lowers the UB9 mean for every kernel baseline under both settings.
Even with test-selected tuning, the strongest kernel baseline, WL-IF at 58.6, remains below graphlet count and Combined statistics.

AUROC below 50 does not automatically imply a reversed score, as illustrated by WL-OCSVM on AIDS at 8.4 with tuning and 4.2 with defaults.
A direction check using no labels confirms that the released scores have the intended anomaly-score direction.
Under these normalized kernels, anomalous test graphs resemble training normals more than normal test graphs do, consistent with the data property discussed by \citet{zhao2023ocgin}.
Constant node features also produce constant propagation-kernel matrices and exactly 50 for both PK-IF and PK-OCSVM.

\subsection{Direct checks of score--label pairing}
\label{app:kernels-pairing}

To isolate score--label misalignment, we run unchanged WL-IF with the released script on 13 datasets and five folds.
We reproduce each fold's UB-GOLD kernel input matrix exactly.
Keeping the scores fixed, we evaluate them once with labels from the second shuffled pass and once with correctly aligned labels.
The four two-stage deep detectors share this pairing pattern.

Table~\ref{tab:kernels-verbatim} reports five-fold means and the fraction of positions with different labels between passes.
ENZ., PROT. and RDT-B abbreviate ENZYMES, PROTEINS\_full and REDDIT-B, while P53 and MMP denote Tox21 tasks.
Labels differ at 19\% to 51\% of positions, and released pairing averages 49.83 over 65 folds, with dataset means of 45.89 to 54.64.
With correctly aligned labels, the same scores give AUROC from 30.90 on DD to 74.81 on REDDIT-B.
Wrong score--label pairing therefore pushes AUROC toward chance even when aligned results differ substantially.
All metrics reported in this paper use aligned labels.

The published-value distribution is consistent with this explanation, with 21 of 26 IF values and 24 of 26 OCSVM values within three chance standard deviations of 50.
A few much higher values, including PK-IF at 75.29 on DD, could not be recovered from the released code, and we retain them without inferring how they were produced.

\begin{table}[!htb]
  \centering\footnotesize
  \caption{Released and aligned pairing in unchanged WL-IF.}
  \label{tab:kernels-verbatim}
  \begin{tabular}{lcccc}
\toprule
Dataset & Folds & Released pairing & Aligned pairing & Labels out of place \\
\midrule
AIDS & 5 & 50.63 & 71.50 & 0.33 \\
BZR & 5 & 46.31 & 45.68 & 0.33 \\
COX2 & 5 & 54.64 & 53.75 & 0.36 \\
DHFR & 5 & 49.69 & 43.64 & 0.43 \\
ENZ. & 5 & 53.15 & 58.19 & 0.26 \\
NCI1 & 5 & 50.41 & 45.46 & 0.50 \\
PROT. & 5 & 49.48 & 32.41 & 0.48 \\
IMDB-B & 5 & 51.45 & 57.46 & 0.47 \\
RDT-B & 5 & 45.89 & 74.81 & 0.51 \\
DD & 5 & 48.20 & 30.90 & 0.48 \\
COLLAB & 5 & 49.55 & 50.49 & 0.50 \\
p53 & 5 & 48.33 & 47.47 & 0.19 \\
MMP & 5 & 50.00 & 44.46 & 0.27 \\
\midrule
All folds & 65 & 49.83 & 50.48 & -- \\
\bottomrule
\end{tabular}

\end{table}

\section{Compute and released resources}
\label{app:compute}

This appendix gives the environment, resource requirements and known limitations needed to reproduce the experiments.
We record estimated costs, measured runtimes and scheduler allocations separately because they cannot be added into one total.

\subsection{Environment and reproducibility}
\label{app:compute-hw}

All runs used a shared academic cluster with one node per job, AMD EPYC CPUs and NVIDIA A100 GPUs with 40\,GB per GPU.
Some CPU runs used GPU nodes for their larger host memory.
Job time limits required long runs to save their state every five minutes and resume in later jobs.
The main environment uses Python 3.12, PyTorch 2.5.1 with CUDA 12.4, PyTorch Geometric 2.6.1 and scikit-learn 1.9.1.
Method-specific dependencies are installed in separate package overlays.
Run records include the device, threads, software versions and code hash, with package lists for each environment in the release.

Follow-up experiments use the same environment and harness, with all Type III runs on CPU.
Other follow-ups retain the corresponding main-grid device type, except that the four pilot detectors use CPU.
Of 600 Type III runs, 106 resumed in later jobs, as did 26 of 295 GPU extra-seed runs.
GraphDE on AIDS folds 0 and 4 started on GPU and finished on CPU, preventing bit-for-bit reproduction on one device.

CPU reruns, including runs resumed in later jobs, reproduce results bit for bit except for GLASS's own randomness.
Among the GPU detectors checked, only GLADC is bit-identical both from scratch and after resuming.
Some methods also produce different scores on CPU and GPU at the same seed.
GPU nondeterminism adds to differences between GPU runs with different seeds.
It also affects one GPU cell in the randomness-isolation and all-normal experiments, as detailed in Appendix~\ref{app:ports-devices}.
Repeating a run with the same seed and comparing runs with different seeds therefore answer different questions.

\subsection{Compute by run group}
\label{app:compute-cost}

Table~\ref{tab:compute} lists the main and follow-up experiments separately, using the study-record part labels to match the logs.
Est. denotes estimates from pre-grid timing, meas. denotes measurements of completed runs, and acct. denotes scheduler allocations.
Measured core-hours equal threads times wall time and include shared-node contention.
Allocated core-hours and A100-hours equal allocated cores or GPUs times elapsed time, including idle resources at the end of jobs that run multiple experiments.
GPU-run-hours sum individual run times on GPUs shared by up to four runs.
Compute was not recorded for the four-detector pilot or the references that require no gradient training.
It was also not recorded for typed-checkpoint analysis D, which reused stored outputs without a separate scheduler job.

\begin{table}[!htb]
  \centering\footnotesize\setlength{\tabcolsep}{4pt}\appRaggedCells
  \caption{Compute by run group.}
  \label{tab:compute}
  \begin{tabular}{@{}p{3.2cm}p{1.45cm}p{10.0cm}l@{}}
    \toprule
    Group & Device & Recorded cost & Kind \\
    \midrule
    UB-GOLD main grid & CPU, GPU & 1601 core-hours on CPU nodes. 1263 on GPU nodes used for memory. 4.5 GPU-run-hours. 1121 core-hours for secondary-dataset defaults & est. \\
    MUSE & CPU & 14.6 core-hours & meas. \\
    LGKDE & CPU, GPU & About 20 core-hours for eight datasets. About 0.7 GPU-hours per REDDIT-B run & est., meas. \\
    GLADPro & CPU & About 495 core-hours, plus about 20 for diagnostics & est. \\
    GLADMamba & CPU & About 307 core-hours (55 runs, 10 diagnostics). 45 runs in one short job & est. \\
    SEGO & CPU & About 6.5 hours per REDDIT-B run, two per node. 90 runs in two short jobs & est., meas. \\
    GLASS & CPU, GPU & About 90 core-hours. Text embeddings in one short GPU job & meas., est. \\
    ProtoGLAD & CPU & 8.7 core-hours, 13.2 including the feature diagnostic & meas. \\
    HC-GLAD & GPU, CPU & 95 of 100 runs on GPU. Three NCI1 CPU runs at about 6 hours each with 40 threads & meas. \\
    SUL/GUL & CPU & 37.4 core-hours & meas. \\
    Kernel baselines & CPU & One CPU node for 23.5 minutes and a 10-core job for 26 minutes & meas. \\
    Typed tests, UB-GOLD & CPU, GPU & 8.2 A100-hours for 30 runs. 77 core-hours for 150 runs & est. \\
    Typed tests, recent & CPU, GPU & About 1900 core-hours for HC-GLAD and about 300 for the other eight & est. \\
    Earlier commit (A17, a96137b) & CPU, GPU & About 14 GPU-hours for 60 runs. About 14 CPU-node-hours for 30 runs & est. \\
    \midrule
    Type III (A) & CPU & 312 core-hours for 600 runs in four pooled jobs & acct. \\
    Extra seeds (C, C2) & CPU, GPU & 698 core-hours for 845 CPU runs. 8.6 A100-hours and 137 core-hours for 295 GPU runs & acct. \\
    Randomness isolation and all normals (F, G) & CPU, GPU & 274 core-hours for 170 CPU runs. 0.08 A100-hours and 1.3 core-hours for ten GPU runs & acct. \\
    Labeled split (E) & CPU & 1.0 core-hour on stored scores & acct. \\
    Low-order references (H) & CPU & 1.4 core-hours & acct. \\
    Follow-up total & & 1425 core-hours, including 139 on GPU nodes. 8.7 A100-hours & acct. \\
    \bottomrule
  \end{tabular}
\end{table}

We do not sum rows above the rule because they mix estimates and measurements.
Each follow-up part requires at most 5.5 node-hours on a 128-core CPU node, the allocation used by the 845 CPU runs of C and C2.
The table excludes smoke runs, which add 47 allocated core-hours and 0.6 A100-hours.
Collecting outputs and computing statistics for Type III and extra seeds (A, C, C2) add 1.3 core-hours.
The randomness-isolation and all-normal analyses (F, G) take less than two minutes.

Host memory often determines the node type for the main grid.
At c940457, GLADC pads each graph to a dense adjacency matrix of the largest graph size and preprocesses all batches before training.
This requires about 230\,GB of host memory on REDDIT-B, so the runs use CPU.
GLocalKD on DD needs about 187\,GB.
Six GLADC COLLAB runs diverge, three folds under each setting, and are reported as DIVERGED under A12 without imputation.

\subsection{Resource datasheet}
\label{app:compute-datasheet}

Table~\ref{tab:datasheet} lists the released materials, their uses and known limitations.
The repository datasheet gives the location of each file.
Retaining UB-GOLD's folds and anomaly definitions allows direct comparison under our protocols.
\oracle{} is a diagnostic upper bound.
The typed data contain controlled motif and ring changes, not a random sample of real anomalies.
\ifanon
\else
The release is at \url{https://github.com/Lu-Yujie/Label-Free-GLAD}.
\fi

\begin{table}[!htb]
  \centering\footnotesize\setlength{\tabcolsep}{4pt}\appRaggedCells
  \caption{Datasheet for the released resources.}
  \label{tab:datasheet}
  \begin{tabular}{@{}p{2.3cm}p{14.4cm}@{}}
    \toprule
    Item & Description \\
    \midrule
    Purpose & Evaluate graph-level anomaly detectors without test-label selection on the same folds, held-out normals, protocols and statistics as this paper. \\
    Splits and diagnostics & Fold indices matched graph by graph to UB-GOLD, including held-out normals. GX and MS-v2, both variants and five seeds, frozen by sha256, with control scores and the retired generator. \\
    Code and outputs & The 12-detector harness and nine recent-method adapters. Run seeds, device, versions, configuration, code hash, third-party commits, evaluation records during training and per-graph scores, except that 180 UB9 pilot runs have evaluation records only. \\
    Reporting & Frozen statistics and leaderboard tables with published values marked as upper bounds. \\
    Follow-up contents & Independently checked, sha256-frozen Type III indices. Per-graph scores for 600 Type III, 1140 extra-seed and 180 isolation/all-normal runs (F/G). Labeled-split tables (E) and low-order reference AUROCs (H). \\
    Source data & Public TU datasets~\citep{tudataset} and Tox21~\citep{tox21}, loaded by UB-GOLD. Type III also uses PTC\_MR, IMDB-MULTI and PROTEINS, distinct from PROTEINS\_full. MS-v2 edits normal NCI1 molecules. No new raw data were collected. \\
    Labels and features & UB-GOLD features and anomaly classes are retained, including majority-class anomalies. Type III treats the second dataset as anomalous. Minority-anomaly folds are included for BZR, COX2 and PROTEINS\_full. Corrected Tox21 HSE and PPAR-gamma folds are also released for future use but are not evaluated in this paper. \\
    People & IMDB-B, IMDB-MULTI, REDDIT-B and COLLAB are already-public social or collaboration networks with graph structure only. No data about people were added. \\
    Intended use & Label-free selection and evaluation of graph-level anomaly detectors, and tests of response to specific structural changes. \oracle{} is a diagnostic upper bound only. \\
    Uses to avoid & Selecting models, thresholds or anomaly classes with test labels and presenting the results as deployable, or interpreting the typed data as a sample of real anomalies. \\
    Known limitations & Type III covers five TU pairs and 12 UB-GOLD detectors. REDDIT-B has no additional training seeds in the checkpoint follow-up. Indices match under PyTorch 2.4.0, 2.5.1 and 2.14.1 but were not checked under UB-GOLD's pinned version. GPU runs are not bit-reproducible except for checked GLADC runs. GLASS is stochastic under current SciPy. MUSE, GLADMamba, HC-GLAD, ProtoGLAD and SUL/GUL are supplied as pinned download scripts and patches rather than redistributed source. \\
    Metadata and maintenance & Machine-readable Croissant metadata accompanies the release. Releases are versioned, corrections appear in new versions, and reports go to the issue tracker. \\
    \bottomrule
  \end{tabular}
\end{table}

\section{Additional tables, figures, and reporting guidance}
\label{app:additional}
This appendix contains the tables and figures cited in Sections~\ref{sec:results} and~\ref{sec:recommendations}.
Appendices~\ref{app:extra} through~\ref{app:typed} give the experimental designs and complete results.
Below we explain the additional tables and figures, note detector-specific exceptions, and list reporting recommendations.

\subsection{Selection rules in the audited code}
\label{app:additional-audit}
Table~\ref{tab:audit} compares released-code behavior with paper descriptions for UB-GOLD and the nine re-run methods.
A filled circle indicates test-label use, an open circle means none was found, L denotes a separate labeled validation set, and n/a indicates no checkpoints.
Checkpoint selection includes early stopping, and other choices include GLADPro's explanation metrics and GLASS's six scorers and 24 scoring rules.
Daggers identify papers that copy baseline rows.
Each row describes the listed code commit, which is not proof that the published results came from that code.
\begin{table}[!htb]
  \centering\footnotesize\appRaggedCells
  \caption{Test labels used for selection in released code.}
  \label{tab:audit}
  \setlength{\tabcolsep}{5pt}
  \begin{tabular}{@{}p{3.4cm}lcccp{6.5cm}@{}}
    \toprule
    Code base & Commit & Checkpoint & HP & Other & Selection rule stated in the paper \\
    \midrule
    UB-GOLD (12 detectors)\newline \citep{ubgold} & c940457 & \audY & \audY & \audN & HP search described as an upper bound, checkpoint rule not described \\
    \midrule
    MUSE\newline \citep{muse} & 07da392 & L & L & \audN & separate labeled validation split \\
    LGKDE$^\dagger$\newline \citep{lgkde} & 1bba5ef & \audY & \audN & \audN & early stopping on a validation set \\
    GLADPro\newline \citep{gladpro} & 1f811b9 & \audY & \audN & \audY & HP grid without a criterion, checkpoint rule not described \\
    GLADMamba$^\dagger$\newline \citep{gladmamba} & 2b2462c & \audY & \audN & \audN & not described \\
    SEGO\newline \citep{sego} & 45bdda7 & \audN & \audN & \audN & not described \\
    GLASS$^\dagger$\newline \citep{glass} & 3bbba39 & \audY & \audY & \audY & main-table rule not described, v2 adds a fixed-epoch re-run \\
    ProtoGLAD\newline \citep{protoglad} & f7d8ce9 & n/a & \audY & \audN & not described \\
    HC-GLAD$^\dagger$\newline \citep{hcglad} & 4238b20 & \audY & \audN & \audN & not described \\
    SUL/GUL\newline \citep{sulgul} & 1a1f2e4 & \audY & \audN & \audN & HP grid without a criterion, checkpoint rule not described \\
    \bottomrule
  \end{tabular}
\end{table}

\subsection{Complete comparison under four protocols}
\label{app:additional-leaderboard}
Table~\ref{tab:leaderboard} reports mean UB9 AUROC for 21 detectors and three italicized references, sorted by \valnormal{}, with ranks among the 21 detectors in parentheses.
Daggers mark recent methods, unlike Table~\ref{tab:audit}, where they mark papers that copy baseline rows.
\oracle{} selects checkpoints using test labels and follows each recent method's own rule when available.
ProtoGLAD has no checkpoints, so its first three columns coincide.
Appendix~\ref{app:extra-leaderboard} also reports results under the common UB-GOLD rule used to average checkpoint gaps across methods.

\begin{table}[!htb]
  \centering\small
  \caption{Complete UB9 leaderboard under four selection protocols.}
  \label{tab:leaderboard}
  \setlength{\tabcolsep}{9pt}
  \begin{tabular}{@{}lcccc@{}}
\toprule
Detector & \oracle{} & \last{} & \valnormal{} & \hpfree{} \\
& {\scriptsize test-selected} & & & \\
\midrule
SIGNET & 74.7 (2) & 70.7 & 71.6 (1) & 69.3 (1) \\
CVTGAD & 76.1 (1) & 71.0 & 71.1 (2) & 66.5 (4) \\
OCGTL & 72.1 (7) & 68.7 & 68.8 (3) & 60.1 (15) \\
GOOD-D & 72.3 (6) & 67.7 & 67.6 (4) & 66.8 (2) \\
GLASS$^\dagger$ & 72.8 (4) & 63.8 & 64.9 (5) & 64.5 (5) \\
OCGIN & 70.2 (8) & 64.5 & 64.8 (6) & 63.9 (10) \\
InfoGraph-OCSVM & 67.7 (12) & 61.3 & 64.7 (7) & 64.1 (8) \\
SEGO$^\dagger$ & 72.7 (5) & 68.4 & 64.5 (8) & 66.7 (3) \\
HC-GLAD$^\dagger$ & 66.5 (14) & 63.1 & 64.2 (9) & 64.3 (6) \\
InfoGraph-IF & 69.5 (9) & 59.8 & 64.1 (10) & 63.4 (11) \\
GraphCL-IF & 67.1 (13) & 63.1 & 64.0 (11) & 63.9 (9) \\
GLADMamba$^\dagger$ & 73.2 (3) & 69.8 & 63.7 (12) & 63.3 (12) \\
ProtoGLAD$^\dagger$ & 63.4 (17) & 63.4 & 63.4 (13) & 59.6 (16) \\
GLADC & 68.2 (11) & 63.1 & 63.4 (14) & 64.2 (7) \\
GraphCL-OCSVM & 68.9 (10) & 62.5 & 62.9 (15) & 62.7 (13) \\
GLocalKD & 66.4 (15) & 62.5 & 62.5 (16) & 61.1 (14) \\
\textit{Combined statistics} & 62.1 & 62.1 & 62.1 & 62.1 \\
\textit{Graphlet count} & 60.4 & 60.4 & 60.4 & 60.4 \\
MUSE$^\dagger$ & 65.6 (16) & 57.9 & 59.0 (17) & 54.8 (17) \\
\textit{Size only} & 57.3 & 57.3 & 57.3 & 57.3 \\
LGKDE$^\dagger$ & 52.8 (21) & 54.5 & 54.1 (18) & 51.5 (19) \\
GraphDE & 54.4 (20) & 51.0 & 52.3 (19) & 53.4 (18) \\
SUL/GUL$^\dagger$ & 58.3 (19) & 48.3 & 49.7 (20) & 49.7 (20) \\
GLADPro$^\dagger$ & 59.0 (18) & 46.8 & 46.3 (21) & 48.9 (21) \\
\bottomrule
\end{tabular}

\end{table}

SIGNET scores 71.6 under \valnormal{} and CVTGAD 71.1, both above every reference, including Combined statistics at 62.1.
OCGTL instead leads AUPRC and FPR95 under both label-free checkpoint rules, so the AUROC ordering does not apply to every metric.
Among the 21 detectors, GLADMamba moves from third under \last{} to twelfth under \valnormal{}, while UB-GOLD's top three change from CVTGAD, SIGNET, and GOOD-D under \oracle{} to SIGNET, CVTGAD, and OCGTL under \valnormal{}.
However, for the 12 UB-GOLD detectors, $\tau_b$ is only 0.70 between the two label-free rules and 0.52 [0.39, 0.76] between \valnormal{} and \hpfree{}, so reordering cannot be attributed entirely to test-label selection.
With the rule fixed, rank correlations range from 0.85 to 0.94 across seeds.
Within one seed, correlations between \oracle{} and \valnormal{} range from 0.73 to 0.82.
Reporting the best of three seeds instead of their mean adds 2.57 points [2.06, 3.03] to \last{}.
No detector pair is significant after Holm correction over nine datasets, so individual ranks remain descriptive.

Tuning gains 8.76 points for OCGTL and 4.53 for CVTGAD.
GLADC, GraphDE, SEGO, GLADPro, and HC-GLAD instead score higher with defaults.
Appendix~\ref{app:posthoc-selectors} compares seven label-free checkpoint rules.
For the six rules with broader coverage, UB-GOLD's mean gaps remain 4.13 to 5.22 points.
The seventh rule applies only to CVTGAD within UB-GOLD and leaves a gap of 10.81 points.
One of the broader rules covers only eight detectors.
Recent-method mean gaps range from 6.69 to 7.97 points, and every interval for an overall gap lies above zero.
Better label-free rules may still exist.

\subsection{Checkpoint gaps and their sources}
\begin{figure}[!htb]
  \centering
  \includegraphics[width=0.48\textwidth]{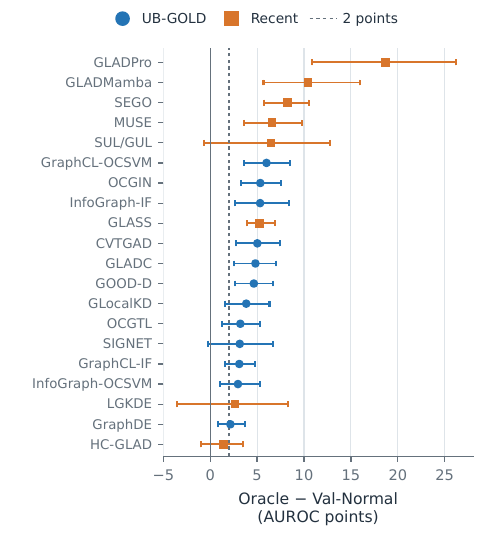}
  \caption{Detector-level gaps relative to \valnormal{}.}
  \label{fig:inflation}
\end{figure}

Figure~\ref{fig:inflation} shows \oracle{} minus \valnormal{} for every detector with checkpoints, using the common UB-GOLD rule and dataset-bootstrap intervals throughout.
Blue circles denote UB-GOLD detectors, orange squares recent methods, and the dotted line the 2-point threshold of G1, with intervals including zero for SIGNET and three recent methods, HC-GLAD, LGKDE and SUL/GUL.
Table~\ref{tab:mechanism} summarizes label-permutation, seed-transfer, and the \emph{disjoint labeled split}.
The checks use different seed configurations and evaluation graphs, so their results cannot be added to explain the main-experiment 5.12-point gap exactly.

After permuting test labels, the highest AUROC over all recorded checkpoints exceeds \last{} by 3.80 points [2.62, 4.98].
This is 71\% of the corresponding observed gap of 5.32 points [4.06, 6.49].
The maximum ignores the original stopping restriction, so it bounds the gain possible from selecting on noise rather than estimating the actual noise share.
In the earlier main-comparison analysis, the ratios are about 60\% for the eight formal-test UB-GOLD detectors and 80\% for recent methods.
A checkpoint evaluated at another seed's selected epoch gains 2.55 points [1.16, 4.02] over \last{} for the evaluated seed.
The remaining 2.77 points [1.97, 3.57] are specific to one training run and are not a pure noise estimate.
For six UB-GOLD detectors, including both InfoGraph variants, more than half the gain is retained when the epoch is chosen using another seed.
For SIGNET and GOOD-D, those checkpoints score below \last{}.
Separate labeled selection gains 1.86 points [0.36, 3.38] over \valnormal{} on the evaluation half, while \oracle{} selected on that same half exceeds \last{} by 6.50 points [5.11, 7.65].
Over \last{}, separate labeled selection gains 8.44 points for InfoGraph-IF, nearly zero for GOOD-D, and a negative amount for SIGNET.
The mean gains of roughly 2.5 to 2.9 points in the cross-seed and separate-selection analyses therefore do not imply that every method benefits.

\subsection{Seeds, scenarios, and protocol checks}
Table~\ref{tab:robust} summarizes checkpoint gaps, using datasets or Type III pairs as units and counting detectors whose mean gaps reach 2 points.
Brackets give dataset-bootstrap intervals, except that the daggered three-dataset comparison resamples folds.
The six-seed row instead gives standard deviations and minima, and n/a means \valnormal{} is undefined because no graphs are held out.
\begin{table}[!htb]
  \centering\small
  \caption{Checkpoint gaps across seeds, scenarios, and protocols.}
  \label{tab:robust}
  \begin{tabular}{@{}p{50pt}@{\hspace{3pt}}c@{\hspace{3pt}}c@{\hspace{5pt}}c@{\hspace{2pt}}c@{\hspace{5pt}}c@{\hspace{2pt}}c@{}}
\toprule
 & & & \multicolumn{2}{c}{\oracle{} $-$} & \multicolumn{2}{c}{\oracle{} $-$} \\
 & & & \multicolumn{2}{c}{\last{}} & \multicolumn{2}{c}{\valnormal{}} \\
\cmidrule(lr){4-5}\cmidrule(l){6-7}
Setting & Det. & Units & Mean & $\geq$2 & Mean & $\geq$2 \\
\midrule
\raggedright UB9, seed 0 & 12 & 9 & 5.12 & 12 & 4.13 & 12 \\
 & & & [3.97, 6.24] & & [3.32, 4.79] & \\
\midrule
\raggedright Seed 0 & 12 & 8 & 5.48 & 12 & 4.49 & 12 \\
 & & & [4.22, 6.63] & & [3.82, 5.16] & \\
\raggedright Seed 1 & 12 & 8 & 5.16 & 12 & 4.31 & 12 \\
 & & & [3.71, 6.50] & & [3.58, 5.17] & \\
\raggedright Seed 2 & 12 & 8 & 5.33 & 12 & 4.37 & 12 \\
 & & & [4.18, 6.37] & & [3.66, 5.00] & \\
\raggedright Six seeds & 4 & 3 & 5.55 & -- & 5.53 & -- \\
 & & & SD 0.42 & & SD 0.52 & \\
 & & & min 4.83 & & min 4.99 & \\
\midrule
\raggedright Type III & 12 & 5 pairs & 9.84 & 10 & 8.11 & 12 \\
 & & & [6.73, 13.07] & & [5.36, 11.00] & \\
\raggedright Minority & 12 & 3 & 2.96 & 8 & 5.00 & 9 \\
class & & & [2.41, 3.52]$^\dagger$ & & [4.24, 5.82]$^\dagger$ & \\
\midrule
\multicolumn{7}{@{}l}{\emph{Subset, seed 0}} \\
\raggedright Original runs & 4 & 3 & 6.05 & -- & 6.31 & -- \\
\raggedright Isolated test evaluations & 4 & 3 & 6.01 & -- & 6.14 & -- \\
\raggedright All normal graphs & 4 & 3 & 5.72 & -- & n/a & -- \\
\bottomrule
\end{tabular}

\end{table}
On eight datasets, \oracle{} minus \last{} is 5.48, 5.16, and 5.33 points across three seeds, while the six-seed subset averages 5.55 with an across-seed standard deviation of 0.42.
Individual detector--dataset--fold results vary more, with mean across-seed standard deviations of 0.8 to 4.1 depending on method and rule, and a maximum of 17.7.
Type III meets criterion R1, but uses only five dataset pairs and fixed anomaly blocks.
Its rank correlation is 0.58 [0.27, 0.76].
The label-permutation upper bound is below half the observed gain for most detectors in that scenario, but CVTGAD and GOOD-D remain within what selection on noise can produce.
Saving and restoring random-number states around test evaluation changes protocol means by 0.28 to 0.46 points and gaps by at most 0.17 points.
Without the only detector--dataset combination run on GPU, isolated test evaluations lower the gap against \valnormal{} by 0.50 points, from 6.64 to 6.14.
Isolation changes \hpfree{} by 0.42 points on average, and removing the normal hold-out changes \last{} by only 0.01 points, leaving a gap of 5.72 points.
After removing flagged runs, using defaults, or omitting one dataset at a time, gaps against \valnormal{} remain 3.96 to 4.43 points.
Resampling detectors as well as datasets widens the interval to [2.92, 5.36], whose lower end is below G1's 3-point threshold.

Re-running GLocalKD and GLADC at a96137b, before the external pull request, changes mean AUROC by at most 0.22 points and checkpoint gaps by at most 0.27 points.
The mean gap persists in these checks, but the results do not cover untested datasets or devices.

\subsection{Complete structural diagnostics and controls}
\label{app:additional-structural}
\begin{figure}[!htb]
  \centering
  \includegraphics[width=\textwidth]{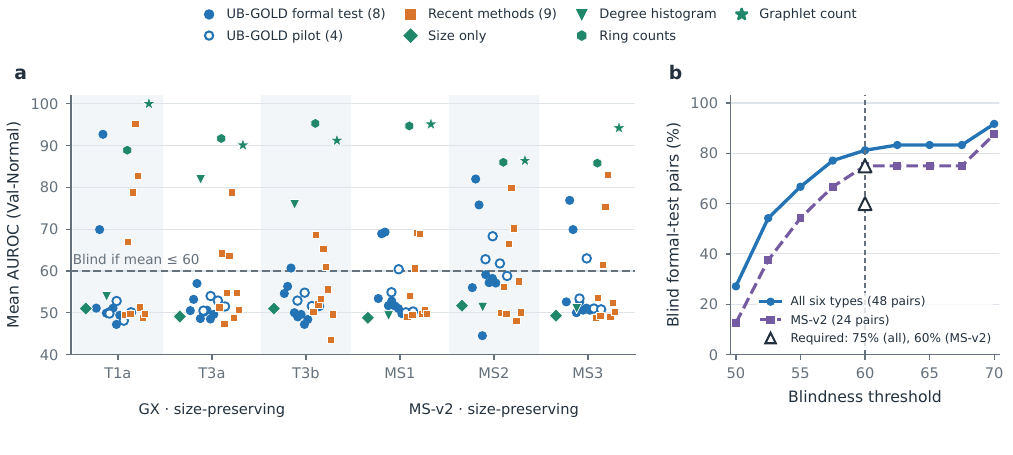}
  \caption{Complete structural responses and threshold sensitivity.}
  \label{fig:typed}
\end{figure}

Figure~\ref{fig:typed}a extends the main-text figure with recent methods and degree-histogram controls, and panel b shows sensitivity at the reported thresholds.
Filled blue circles, open blue circles, and orange squares denote the eight formal-test detectors, four pilot detectors, and nine recent methods, respectively, and green symbols denote references.
In panel b, the solid blue line represents all 48 pairs, the dashed purple line the 24 molecular pairs, and open triangles the required shares at threshold 60.
In a post hoc sensitivity analysis, the G2(c) criteria hold at the reported thresholds from 57.5 to 70 and fail at 55 and below, so the threshold still affects the binary conclusion.
CVTGAD and GOOD-D score at most 60.7 on GX proportion shifts.
After their structural encodings are replaced with constants and the models are retrained, both miss all six size-preserving types but still detect size-changing variants.
The size-changing control attaches a donor molecule, increasing node count by 20 to 100\%.
On size-changing variants, the detectors in the formal test score 74.7 to 88.5 with their fixed configurations, except GraphDE at 43 to 48.
On molecules, degree histograms score 49.3 to 51.3, labeled-degree histograms 52.1 to 58.2, and ring counts 85.8 to 94.7.
On GX, degree histograms score 81.9 and 75.9 on proportion shifts but only 53.9 on the new motif.
The synthetic and molecular controls therefore require separate interpretations.
No molecular proportion-shift generator passed the validity controls, so proportion shifts are tested only on GX.
In Type III, the size reference reaches 89.1 on AIDS+DHFR, near the detectors' \valnormal{} mean of 90.8, and 66.7 on ENZYMES+PROTEINS, above their \oracle{} mean of 61.1.
Size matching retains at least 30\% of anomalies in every fold of six datasets.
The size reference scores 47.4 to 51.8 on five datasets but still 66.7 on REDDIT-B.
The mean for the eight recent methods with checkpoints also falls from 58.3 to 54.7.
The drop after size matching supports a contribution from graph size to benchmark performance.
It does not show that detectors rely only on size, and the leading detectors still outperform simple references.

\subsection{Reporting guidance and remaining scope limits}
\label{app:additional-scope}
Table~\ref{tab:checklist} lists what future evaluations should report, based on our findings.

\begin{table}[!htb]
  \centering\small
  \caption{Reporting items supported by the findings.}
  \label{tab:checklist}
  \begin{tabular}{@{}lp{9cm}l@{}}
    \toprule
    Item & Information to report & Evidence \\
    \midrule
    Training & Training and anomaly classes & \S\ref{sec:f1} \\
    Tuning & Tuning data and default results & \S\ref{sec:f2} \\
    Checkpoint & Rule and data used for selection & \S\ref{sec:f2} \\
    Labels & Whether labeled anomalies are available, and separate selection and evaluation graphs when they are & \S\ref{sec:f2-mech} \\
    Seeds & All seeds or their mean & \S\ref{sec:c2} \\
    Direction & Score direction fixed before results & App.~\ref{app:defects} \\
    Baselines & Source, commit, selection rule, and anomaly class & \S\ref{sec:f1} \\
    Structure & Controlled structural tests, and size and count references & \S\ref{sec:f3} \\
    Test labels & Final metrics, with test-selected bounds marked & \S\ref{sec:f2} \\
    Outputs & Trajectories, scores, and configurations & \S\ref{sec:f2-mech} \\
    \bottomrule
  \end{tabular}
\end{table}

Unsupervised model selection also depends on which labels are available outside graph learning.
Repeated use of a public test set across studies differs from using test scores to choose a checkpoint within one run \citep{ma2023uoms,efficientad,ericsson2023better,recht2019imagenet,roelofs2019meta}.
Further tests are needed to determine whether label-free selectors from other domains can recover the gains retained across seeds here, and whether other graph benchmarks show similar gaps \citep{metaod,autouad}.
Our minority-class check covers only BZR, COX2, and PROTEINS\_full and retains settings tuned for the opposite orientation, so it is not a retuned minority-class benchmark.
Types I and II retain 13 of the 15 original datasets and add MUTAG, with secondary coverage including DD, COLLAB, Tox21 p53, and MMP, and six GLADC runs diverge on COLLAB.
The additional training-seed analysis in Appendix~\ref{app:followup-seeds} covers only UB-GOLD detectors and excludes REDDIT-BINARY because of its computational and memory costs.
Recent methods also have five-seed results in the controlled structural tests (Appendix~\ref{app:typed-results}).
Type III excludes OGB pairs, and the released Type IV loader is commented out and was not run.
LGKDE misses its reproduction check, GLADPro's check is contested, and SUL/GUL lacks a comparable setting, keeping recent-method comparisons descriptive.
A broader hyperparameter search using only normal graphs could improve performance without anomaly labels.
Defaults previously tuned on these datasets could also make \hpfree{} optimistic.
Our experiments rule out neither possibility.

\end{document}